# Advancing an Interdisciplinary Science of Conversation: Insights from a Large Multimodal Corpus of Human Speech

Andrew Reece[1]†*, Gus Cooney[2]†*, Peter Bull[3], Christine Chung[3], Bryn Dawson[1], Casey Fitzpatrick[3], Tamara Glazer[3], Dean Knox[2], Alex Liebscher[1], and Sebastian Marin[1]

Authors' Note:

†Andrew Reece and Gus Cooney share first authorship.

[1] BetterUp Inc. San Francisco, CA 94103, USA

[2] University of Pennsylvania, PA 19104, USA

[3] DrivenData Inc. Berkeley, CA, 94709, USA

*Correspondence should be addressed to Andrew Reece, BetterUp, Inc., andrew.reece@betterup.com, and Gus Cooney, University of Pennsylvania, gusco@wharton.upenn.edu.

Aside from AR and GC, the remaining authors are listed alphabetically; see Author Contributions.

We extend particular thanks to all the participants in the corpus who agreed to make their conversations public.

Data are available for access via free registration here: https://betterup-data-requests.herokuapp.com/

Materials, code, and important links (e.g., analysis scripts, Data Dictionary, Qualitative Review, etc.) are here: https://osf.io/fbsgh/



**Table of Contents**





**ABSTRACT**

People spend a substantial portion of their lives engaged in conversation—and yet our scientific understanding of conversation is still in its infancy. In this report we advance an interdisciplinary science of conversation, with findings from a large, novel, multimodal corpus of 1,656 recorded conversations in spoken English. This 7+ million word, 850-hour corpus totals over 1TB of audio, video, and transcripts, with moment-to-moment measures of vocal, facial, and semantic expression, along with an extensive survey of speakers' post-conversation reflections. We leverage the considerable scope of the corpus to: (1) extend key findings from the literature, such as the cooperativeness of human turn-taking; (2) define novel algorithmic procedures for the segmentation of speech into conversational turns; (3) apply machine learning insights across various textual, auditory, and visual features to analyze what makes conversations succeed or fail; and (4) explore how conversations are related to people's well-being across the lifespan. We also report (5) a comprehensive mixed-method report, based on quantitative analysis and qualitative review of each recording, that showcases how individuals from diverse backgrounds alter their communication patterns and find ways to connect. We conclude with a discussion of how this large-scale public dataset may offer new directions for future research, especially across disciplinary boundaries, as scholars from a variety of fields appear increasingly interested in the study of conversation.





Conversation hardly needs introduction. It is a uniquely human act of cooperation, requiring coordination across many levels of cognition (Clark, 1992; Enfield, 2017; Pickering & Garrod, 2004; 2021; Sacks et al., 1978; Stolk et al., 2016); it is the seat of language acquisition (Bateson, 1975; Tomasello, 2003); its system of turn-taking shows phylogenetic parallels in all clades of the primate lineage (Levinson & Holler, 2014; Pika et al., 2018); it is how people absorb and transmit culture (Henrich, 2015; Herrmann et al., 2007); it is the primary tool that people use to form and maintain their social relationships (Dunbar, 1998; 2004); it is has a significant impact on people's mental and physical health (Mehl et al., Holleran, 2010; Milek et al., 2018); and, more recently, generative models of conversation have emerged as a major milestone in artificial intelligence (Dinan et al., 2020; Ram et al., 2018; Roller et al., 2020).

Despite its centrality, the empirical study of conversation has been hampered by its complexity: Conversation is characterized by a high degree of interdependence between speaking partners, with one's words and behavior being adjusted on the fly in response to what one's partner is doing; conversation is massively multimodal, involving information transmission across linguistic, paralinguistic, and visual channels simultaneously; and conversation is highly contextualized, with people acting out social roles, pursuing specific goals, and negotiating status and power hierarchies. In turn, this complexity presents numerous scientific challenges—from operationalization, to measurement, to statistical modeling—but in this paper, we demonstrate that recent technological advances have begun to offer solutions to these challenges, placing previously inaccessible research questions within reach and offering new avenues for interdisciplinary collaboration.



Historically, progress on conversation research has been catalyzed by large public datasets, such as the Maptask Corpus (Anderson et al., 1991), the Switchboard Corpus (Godfrey et al., 1992), or newer multimodal datasets, such as the MELD (Poria et al, 2018; see also, Chen et al., 2018) and OMG-Empathy datasets (Barros et al., 2018) (for a review see, Serban et al., 2015). While these datasets have advanced conversation science, none include a large sample of naturalistic conversations, with full audio and video recordings, along with speakers' detailed post-conversation reports.

We have collected such a dataset of 1,656 unscripted conversations over video chat, comprising over 7 million words and 850+ hours of audio and video. All told, our corpus includes more than 1TB of raw and processed recordings, transcripts, high-frequency behavioral measures, and post-conversation impressions. The corpus draws on a large and diverse sample of participants, aged 19-66, from all around the United States. Participants were paired using an automatic matching algorithm of our own design, and simply instructed to have a conversation with one another for at least 25 minutes—although many talked for much longer. Conversations occurred over the year 2020, offering a unique lens into one of the most tumultuous years in recent history, including the start of a global pandemic and a hotly contested presidential election. All told, the corpus is among the largest multimodal datasets of naturalistic conversation, which we collectively refer to as the CANDOR corpus (*Conversation: A Naturalistic Dataset of Online Recordings*).

Large quantities of raw data alone are not enough to advance the study of conversation. In other domains, growth in computational power, the use of crowdsourcing platforms, and technological advances in machine learning—e.g., language and signal-processing algorithms like Word2Vec, BERT, and ResNet —have proven to be yet another catalyst for scientific



advance, enabling discovery and inference at scale (Bowman et al., 2015; Callison-Burch & Dredze, 2010; Devlin et al., 2018; He et al., 2016; Mikolov et al., 2013; Munroe et al., 2010). In this spirit, we applied an elaborate computational pipeline to quantify mechanical features of conversation such as: overlaps and pauses; second-by-second variation in facial features such as nods and emotion expressions; and full transcripts with accompanying prosodic characteristics of speech. Finally, we directly collected a battery of psychological measures from participants, including trait-level measures such as personality, as well as people's opinions about their conversation partner and their feelings about the conversation overall.

We report our results in five sections. Section 1 uses the corpus to replicate key findings from the literature on conversation, such as cooperativeness of turn-taking. We then develop novel algorithmic procedures for segmenting speech into conversational turns—a pre-processing step necessary for the study of conversation at scale—and demonstrate how analytic results hinge critically on the choice of appropriate segmentation algorithms. In Section 2, we explore the relationship between conversation and psychological well-being, demonstrating how people underestimate how much their partners enjoy the conversation. Furthermore, the scope of the corpus enables further exploration of unknown patterns of heterogeneity, such as how pessimistic self-evaluations decline markedly among older participants while other related phenomena remain stable over the lifespan. Section 3 demonstrates the strong and nonlinear relationship between the structure of conversation and its psychology, highlighting how these phenomena should not be studied in isolation. Next, in Section 4, we apply a series of computational models to transcripts, audio data, and visual data, extracting detailed measures of turn-by-turn behavior that we use to answer an open question in the literature: What distinguishes a good conversationalist? Finally, Section 5 conducts a rich, mixed-methods



analysis of the topical, relational, and demographic diversity of our corpus: we analyze how the national discourse shifted over a tumultuous year, qualitatively review the entire corpus to identify new concepts like depth of rapport; and quantitatively examine how people alter their speech, listening patterns, and facial expressions when talking to partners from diverse backgrounds and identity groups.

To structure these findings, we articulate a framework for the study of conversation based on a more "vertically integrated" approach. Our hierarchy of conversation spans: (1) "low-level" mechanical features of conversation, such as turn-taking, which delineate the structure of interaction; (2) "mid-level" information streams such as semantic exchange, psycholinguistic markers, and facial emotion, representing the subjective content of turn-by-turn conversation; and ultimately, (3)" high-level" impressions and judgments such as psychological well-being, personality, and perceptions of social status. Many of our results demonstrate that these levels cannot be meaningfully studied in isolation.

The findings we present from the corpus are far from exhaustive. Rather, they are intended as a launching point for future research and collaboration. In other contexts, the emergence of grassroots consortiums (e.g., ManyBabies Consortium, 2020; Many Primates et al., 2019) has allowed scientists to pool ideas and resources, allowing "big" science to tackle big questions (Coles et al., 2022). And indeed, many of our results demonstrate the considerable advantages that come with studying conversation through a multidisciplinary lens, but also the need for larger collaborative efforts in order to make empirical progress. Our aim for the corpus is to serve as both a model and touchstone for a new interdisciplinary science of conversation.

Many disciplines have been drawn to the study of human conversation—some for decades (discourse analysis, psycholinguistics, conversation analysis, communications,



pragmatics), and some more recently (cognitive and social psychology, neuroscience, organizational behavior, political science, computational linguistics, natural language processing, and artificial intelligence). In all cases, it seems clear that progress has been catalyzed by rich datasets, new frameworks, and empirical findings that beckon further collaboration, all of which we have aimed to provide. Together, may these offerings advance the study of that most fundamental of all human social activities: the spoken conversation.

## CORPUS CONSTRUCTION

Between January and November of 2020, six rounds of data collection yielded a total of 1,656 dyadic conversations (See Table 1). These conversations were recorded over video chat. In what follows, we explain our recruitment method and the construction of the final dataset.

### Methods

#### *Recruitment*

**Initial Survey.** Our target population consisted of people 18+ years of age who were based in the United States. We recruited participants using Prolific, an online crowdwork platform. Before entering the study, candidate participants were asked to read a consent form that explained the following: (i) they would have a conversation with another individual that would last at least 25 minutes; (ii) their audio and video would be recorded; (iii) they would complete a series of surveys before and after the conversation; (iv) they would be paid $0.85 for completing the initial survey and an additional $14.15 upon full completion of the recorded conversation and post-conversation survey; (v) their data, including the video and audio recording, would be shared with other researchers and could be made publicly available; and (vi) participation carried a risk of personal identification, due to the audio and video recordings. Due to the sensitive nature of releasing personally identifiable recordings, as well as the study design that required



participants meet up for a video call after the initial informed consent document, we then asked participants to verify, for a second time, that they were comfortable having a recorded conversation with a stranger. Only candidates who both indicated and reaffirmed their consent were permitted to continue as study participants.

Participants were then asked a series of questions to determine their availability over the next week. After doing so, participants received a follow-up email within 24 hours.[1]

In some data collection rounds, participants filled out a small number of additional psychological measures (see the round variable in the Data Dictionary for details). Finally, participants were given instructions for submitting a request for compensation.

**Matching.** A matching procedure was carried out once per day, based on participants' stated availability over the next week. Unmatched participants with overlapping availability were paired. No demographic information was used in the matching process. Once matched, participants were notified by email of the time and date of their conversation. A second email was sent one hour prior to the scheduled conversation that contained a link to a survey, which guided participants through the next phase of the study.

*Participants*

Of the participants who completed the intake survey (approximate N = 15,000), roughly 3,500 were matched with another participant, returned to have a conversation, and produced audio and video that was able to be automatically processed by our pipeline. Naturally, given the difficulties inherent in scheduling strangers on the internet to meet up at a specific date and time in a video chat room, over the course of data collection we experienced cancellations, no-shows, technologically confused participants, and other obstacles. For instance, just over 3,000 participants reported at least one instance of their partner simply not showing up for the scheduled video call. We addressed this contingency by compensating participants who experienced no-shows with a $1.50 honorarium, and by offering participants the opportunity to rejoin our matching pool on the subsequent day.

---

[1] A small number of participants were told that there were two ways that they could be matched with a conversation partner: (1) "try and find a chat partner now;" or (2) "set up something later this week." Participants who chose to find a chat partner now entered into an automatic matching service, which automatically found and paired participants who were online at the same time. If matched, participants immediately entered the second phase of the study, which included the actual conversation.

All told, we recorded just under 2,000 completed conversations by the end of the data collection period, totaling roughly 1,000 hours of footage. An additional human review of all our conversations flagged approximately 300 conversations for removal (see Section 5). Conversations were removed for two main reasons: another individual appeared on camera who did not consent to be filmed (e.g., a participant wanted their conversation partner to say hello to one of their children) or technical issues made the audio or video recording unusable.

Our final dataset included 1,656 conversations and 1,456 unique participants who spanned a broad range of gender, educational, ethnic, and generational identities (see Table 2). "Unique participants" refer to the number of participants who had *at least* 1 conversation, as more than 50% of our sample had 2+ conversations and 33% of the sample had 3+ conversations. Participants who had multiple conversations did not do so back-to-back, and in most cases, had several conversations spread out across the data collection period (see Table 1).

---------------------------------

| Recruitment Round | Recruitment Date | *N* Conversations | Percent of Conversations | *N* Unique Speakers |
|---|---|---|---|---|
| Round 1 | 01/07 - 01/14 | 46 | 2.78 | 92 |
| Round 2 | 05/01 - 05/29 | 183 | 11.05 | 366 |
| Round 3 | 06/26 - 07/14 | 196 | 11.84 | 391 |
| Round 4 | 07/29 - 08/31 | 403 | 24.34 | 475 |
| Round 5 | 10/08 - 11/07 | 423 | 25.54 | 480 |



| Recruitment Round | Recruitment Date | *N* Conversations | Percent of Conversations | *N* Unique Speakers |
|---|---|---|---|---|
| Round 6 | 11/09 - 11/25 | 405 | 24.46 | 493 |

*Table 1.* Dates and sample size of each collection round. Unique speakers reflect the unique speakers *per round*.

----------------------------------

----------------------------------

| Demographics | | *N* | Percentage |
|---|---|---|---|
| **Age** | 18-25 | 425 | 29.19 |
| | 25-35 | 499 | 34.27 |
| | 35-45 | 286 | 19.64 |
| | 45-55 | 129 | 8.86 |
| | 55+ | 83 | 5.70 |
| | Not Reported | 34 | 2.34 |
| **Gender** | Female | 782 | 53.71 |
| | Male | 610 | 41.90 |
| | Other or Prefer not to Answer | 30 | 2.06 |
| | Not Reported | 34 | 2.34 |
| **Race/Ethnicity** | American Indian or Alaska Native | 7 | 0.48 |
| | Asian | 200 | 13.74 |
| | Black or African American | 117 | 8.04 |
| | Hispanic or Latino | 108 | 7.42 |
| | Mixed | 53 | 3.64 |
| | Native Hawaiian or Pacific Islander | 2 | 0.14 |
| | Other | 13 | 0.89 |
| | Prefer not to Say | 2 | 0.14 |
| | White | 920 | 63.19 |
| | Not Reported | 34 | 2.34 |
| **Education** | Associate Degree | 97 | 6.66 |
| | Bachelor's Degree | 567 | 38.94 |
| | Completed High School | 81 | 5.56 |
| | Doctoral Degree | 32 | 2.20 |



| | | |
|---|---|---|
| Master's Degree | 247 | 16.96 |
| Professional Degree | 36 | 2.47 |
| Some College | 354 | 24.31 |
| Some High School | 8 | 0.55 |
| Not Reported | 34 | 2.34 |

*Table 2.* Demographic information for participants in the corpus.

### Pre-Conversation Survey

A pre-conversation survey briefly measured participants' current mood (i.e., valence and arousal; see Data Dictionary). The survey then reminded participants to enable their webcam and microphone, to make sure their conversation lasted at least 25 minutes, and to return to the survey tab in their browser once the conversation was over in order to complete the post-conversation survey. Finally, a link was provided to the video chat room.

### The Conversation

Clicking the "Join Conversation" link opened a new video chat window. Recording began as soon as the first conversation partner joined. Participants were asked to wait at least 5 minutes for their partner to arrive. Sessions for which only one participant joined were discarded.

Regarding conversation content, participants were not given specific instructions—they were simply told to "talk about whatever you like, just imagine you have met someone at a social event and you're getting to know each other." Participants were then instructed to have a conversation for at least 25 minutes, although conversation lengths varied considerably (mean length = 31.3 min, *SD* = 7.96, minimum = 20 min).

Conversations were digitally recorded using a web application based on the TokBox OpenTok Video API and conducted via camera-connected displays and microphones. Most conversations were conducted computer-to-computer, but occasionally a mobile device was



used. Upon completion, participants ended the recording session, and returned to complete the post-conversation survey.

*Post-Conversation Survey*

Participants were first asked whether significant issues prevented them from completing the conversation. If so, we offered the opportunity to reschedule with a new partner by responding with their updated availability. Otherwise, participants went on to complete a post-conversation survey, in which they reported their overall feelings about the conversation, their perceptions of their conversation partners, their personality, and so forth. For details, see the Data Dictionary (link in Data Availability Section), which details all measures. Finally, participants were thanked for their participation and provided instructions on how to submit a request for payment.

**Data Processing & Feature Extraction**

The two primary outputs from the conversation collection process were survey responses and the video archive, containing videos and metadata for each conversation. Each participant's video stream was saved as an independent video in .mkv format. If a participant's connection dropped and then rejoined the conversation session when back online, a new video file was created in addition to the existing one. Here, we describe how the survey responses and the video archives were processed into unified, user-friendly formats. This data processing pipeline extracted various feature sets and transformed the raw conversation data into the analysis-ready, structured data described in the Data Dictionary.

*Conversational Alignment*

Processing of recorded conversations started with the creation of a coherent, single-file representation of the conversation from each partner's respective video files. Programmatic



alignment consisted of four primary steps using the video processing software, FFMPEG. First, input media were reencoded to correct possibly corrupted timestamps. Second, the TokBox metadata, which provided a timeline of when participants joined, left, and possibly rejoined the conversation, was verified and corrected by measuring the duration of the media.[2] Third, after metadata correction, an individual participant's videos were combined into a single video, adding padding and blank filler segments where appropriate.[3] Finally, participants' aligned videos were combined into a unified representation. Separate audio channels for each participant were created for downstream automated transcription.

Overall, joining the conversation videos posed a non-trivial challenge, and required a number of subjective, albeit carefully reasoned, decisions. The corpus therefore includes both the raw video files along with their merged versions, so that other researchers may apply their own alternate methods for alignment and synchronization.

### Feature Extraction

We describe the processes used to generate and extract analyzable features according to the source of the information: transcripts, audio, video, and survey responses. The outputs of these processes are analysis-ready in the sense that they are structured into common file formats and indexed by a shared timeline or by conversational turns where appropriate.

---

[2] For some video segments the TokBox software does not accurately record the start and stop time of the video stream correctly relative to the overall timeline of the conversation. Video durations (and therefore offsets) can be verified using the FFMPEG tool FFPROBE to measure the audio and video stream durations and compare them to those reported in the metadata, adding or subtracting appropriate offsets to the `stopTimeOffset` where necessary. Correcting the `startTimeOffset` is more difficult and requires heuristics since there is no trusted reference point in time. We chose the heuristic of minimizing audio signal overlaps during playback as a proxy for proper alignment. Such a heuristic is imperfect which impacts a small number of conversations in the dataset leading to slight misalignment in the unified audio signals; the independent signals are also included. Finally, note that videos where the alignment was problematic in human review were excluded.

[3] For example, if a participant dropped and rejoined 10 seconds later, two videos would have been created in the archive. Directly joining these videos was not helpful, because of that 10 second gap; the resulting merged video would be misaligned. However, the gap in time was also reflected in the metadata, and so a 10 second "blank" video was inserted to maintain alignment with respect to the overall conversational timeline.



**Textual.** Processing of textual information involved transcription, turn identification, and extracting speaking statistics.

*Transcription.* To produce a transcript of the conversation, we processed the aligned conversation files using the Amazon Web Services (AWS) *Transcribe* automated transcription service. The raw transcript and tokens returned by the Transcribe API are included in this data release.

Note that while the transcripts are very usable, the quality of automated transcription is far from perfect. Throughout the development of this dataset, we tested numerous automated transcription services, and each of them left much to be desired. An important direction for future work on this dataset is the development of "gold standard" transcripts, either via improved automated transcription or human labeling.

*Turn identification.* The sequential representation of text in alternating turns is essential to many conversation analyses. The definition of a turn, however, can vary, depending on how pauses, overlaps, back-channels, and other complications are preprocessed. The simplest way to construct a conversational turn is to assign each word token to a participant's turn until a token from another participant occurs, at which point that participant's turn begins, and so on. This is the default method used to construct turns from the raw transcript. Although limited in many respects, this approach provides a useful reference point for improved algorithms that we develop below (for a discussion, see Section 1).

*Turn-based and time-based feature aggregation.* Once equipped with conversational turns, we considered two possible approaches to aggregating the remaining acoustic, visual, and textual features: time-based (finer-grained) and turn-based (coarser). In the release, we include time-based aggregations at a one-second resolution. Researchers can use the turn timestamps



noted in the transcript files to aggregate turn-based features as desired. For a comprehensive list of corpus features and how they were computed, see the Data Dictionary.[4]

**Acoustic.** Processing of acoustic information involved spectral characterizations, phonation, and prosody features.

***Spectral characterizations.*** A number of calculations were performed on the audio of the conversations. For example, the fundamental frequency (F0) of people's speech was computed over 0.01 s intervals using the Parselmouth package (Jadoul et al., 2018), which implements Praat (Boersma et al., 2022) functionality in Python. Further, the Python library Librosa (McFee et al,. 2015) was used to compute the first 13 Mel-frequency cepstrum coefficients, as well as various additional spectral features such as the spectral centroid, contrasts, zero-crossing rate, and others (Mermelstein, 1976). These features were aggregated using the mean value over 1 s intervals.

***Phonation and prosody.*** Baseline vocal pitch (F0) was used along with signal energy to compute other prosodic features; for example, jitter and shimmer, which measure the variance of pitch and volume respectively (Farrús et al., 2007). We also computed a measure of vocal "intensity" (sometimes referred to as "activation"), which is a measure of emotion and momentary affect (e.g., Frijda, 1988; 2017; See Section 4 of the Results). To do so, we trained a model on the Ryerson Audio-Visual Database of Emotional Speech and Song (RAVDESS) (Livingstone & Russo, 2018), and then applied this model to our corpus. These features were aggregated using the mean value over 1 s intervals.

---

[4] Note that while the entire TokBox session is included for each conversation, the conversation is said to have begun the moment both participants have joined the session. In all of the turn-based indexing included in this release, this moment is specified as `turn_id`=0, with all prior data for the session being indexed as `turn_id`=-1. So, if you are watching a conversation video and observe a particular moment of interest, you could locate the turn number by searching the utterance field of the turn-based aggregation of the conversation. Using the conversation ID and turn number, you could then index into any of the other features discussed in this section; for example, the probability that each participant is displaying a happy facial expression during that turn.

**Visual.** Processing of visual information involved smile, nod, and emotion detection. Visual features were computed at 1 s intervals (frames). In some frames, a face was not detected, and for these frames the visual features were recorded as null values.

*Smile and nod.* We used OpenCV, a computer vision software library, to extract a set of facial landmark locations within each visual frame. We then applied a set of heuristics to estimate whether or not a participant was smiling or nodding for any given frame. We attempted to measure gaze as well but were not satisfied with the consistency of results.

*Emotion detection.* An emotion recognition model was trained using the AffectNet dataset (Mollahosseini et al., 2019). The model, a convolutional neural network, assigned a probability distribution across eight emotional classes (happy, sad, angry, etc.) to each frame of facial expression, per speaker.

**Survey.** The survey data consists of: (1) the initial survey administered at the screening stage, (2) the pre-conversation survey, and (3) the post-conversation survey.

Survey responses were processed via the Qualtrics API into a flat file of comma-separated values. Participant responses were recorded at the level of the conversation. For details about survey items, please refer to the Data Dictionary.

Across all these modalities—textual, acoustic, visual, and survey—our goal was to streamline and extract as much information as time and technology would permit, producing a user-friendly corpus for researchers to use and improve upon. With future advances in preprocessing and machine-learning algorithms, human review, and additional grit, we anticipate that current and future scholars will unlock considerable additional value from this corpus. We



encourage researchers who develop improvements to this corpus to make their improvements

publicly available to the broader scientific community.

## A FRAMEWORK TO EXPLORE THE CORPUS

To guide our exploration of the CANDOR corpus, we embrace three principles: (1)

Structurally, conversation is built around a highly cooperative system of turn-taking,

conversation occurs across multiple modalities (e.g., text, audio, visual), and it involves linkages

across multiple "levels" of analysis (see Framework, below); (2) Understanding the full

complexity of conversation requires insights from a variety of disciplines that, despite examining

the same phenomenon, often remain siloed in their research questions and analytic tools; (3)

Examining conversation computationally and at scale is an enduring challenge, but new

technologies, especially advances in machine learning, promise to unlock many aspects of

conversation that were previously inaccessible to empirical research.

### A Framework

Our organizing framework classifies conversational features as *low-level*, *mid-level*, and

*high-level* (See Figure 1). Low-level features are closest to the raw signals in the audio, video,

and text of a conversation recording and often vary on a nearly continuous timescale. Although

even these features often require some degree of inference to generate, such as vocal markers

from processed audio signals or the linguistic inferences made by an automated transcription

service, the outputs of these processes are sufficiently concrete and specific (e.g., pitch, turn

duration, eye gaze, etc.) to constitute the objective properties of conversation upon which higher-

order inferences are derived.

Next, high-level features are individuals' subjective judgments about their conversations,

formed on a coarse timescale and reflected in the post-conversation survey responses. Survey



items included measures of liking, enjoyment, and conversational flow, as well as evaluations of one's partner's social status, intelligence, and personality. The value of these post-conversation ratings is considerable, as they allow for the linkage between *in*-conversation behaviors and *post*-conversation impressions.

Between these levels, we identify numerous features relating to subjective perceptions of interaction that typically vary at an intermediate timescale. These mid-level features capture intra-conversational psychology and are usually computed using a suite of algorithmic tools which were trained to attend to specific aspects of speech, sound, and movement to infer psychological content. A happy facial expression, a growing intensity in one's voice, a timely change of subject; noticing these conversational moments requires a mix of sense and sense-making—whether by human or machine—analytically distinct from low- and high-level phenomena.

We refer interested readers to the Discussion section for theoretical implications of this tiered framework for studying conversation. A simple example demonstrates the nature of low-, mid-, and high-level distinction: The contraction of a person's zygomaticus major muscle is an observable (low-level) feature. Most people recognize this contraction pattern as a smile (mid-level inference), a momentary expression of happiness. Finally, individuals who frequently smile during a conversation may also report having had an enjoyable experience (a high-level, subjective conclusion). We use this framework as a heuristic for organizing a diverse array of findings across a rich dataset.[5]

---

[5] Note that this example, while convenient as an illustration of our leveling framework, is not strictly prescriptive in the way it creates dependencies across all three levels; for example, there may be low-level features that directly trigger high-level impressions without requiring any sort of mid-level representation to influence judgment (see Discussion).



In what follows, we first present results related to conversational mechanics, on the lowest level of our framework, covering features such as turn-taking and back-channel feedback, and how studying these low-level mechanical features requires new algorithmic developments in transcript segmentation (Section 1). At the high level, we then examine how conversation influences an individual's well-being, how people make judgments about whether their partners enjoyed the conversation, and how this varies across the lifespan (Section 2). We then demonstrate the interplay between levels, showing, for example, how an individual's low-level speed of turn exchange relates to their partner's high-level enjoyment of the conversation (Section 3). Next, we explore the middle-layer of the corpus by extracting psychologically rich features with an array of computational models. These fine-grained measures of turn-by-turn interaction are used to link the mid- and the high-level, answering a basic unanswered question in conversation research: What distinguishes someone as a good conversationalist? (Section 4). Finally, we end with a mixed-method report exploring the topical, relational, and demographic diversity of our corpus (Section 5).

----------------------------------

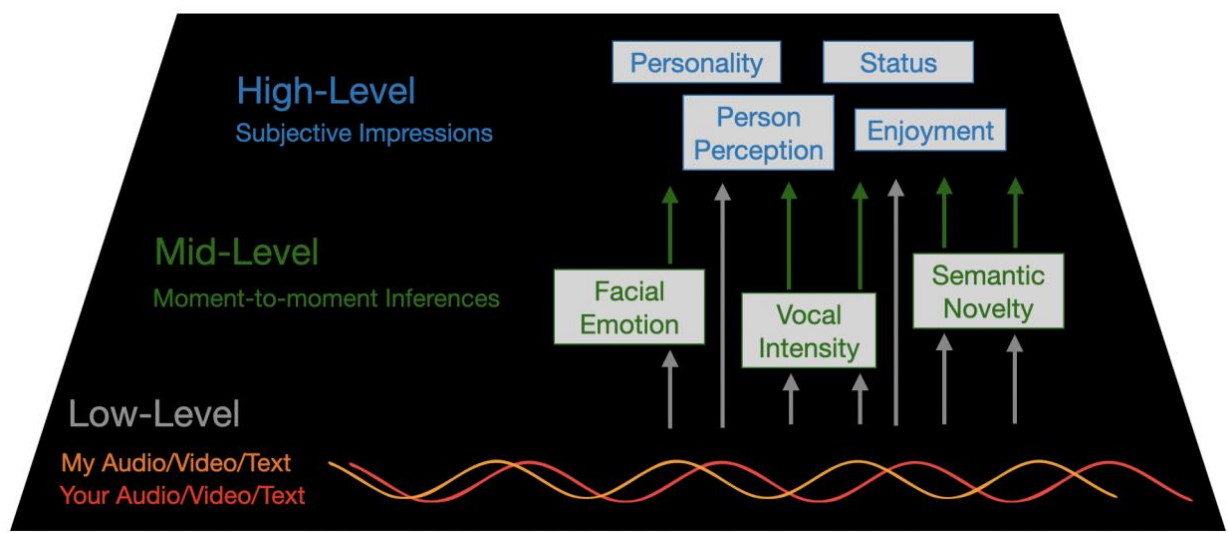

*Figure 1.* **A framework for studying conversation.** Results are organized according to an analytic framework that distinguishes between three related levels of conversation. *Low-level* features are directly



observable, vary over short time periods, and often relate to conversational structure, e.g., a pause that delineates the end of one speaker's turn and the start of the next. *Mid-level* features are generally indirectly inferred by listeners or algorithms, vary on a medium-frequency, or turn-by-turn basis, and capture linguistic or paralinguistic conversational content, e.g., psychologically meaningful quantities such as a happy facial expression or emotional intensity in a voice. *High-level* features relate to people's subjective judgments about a conversation, e.g., personal enjoyment or evaluations of their partner. Subsequent sections present empirical results at each level of the hierarchy, as well as analyses that demonstrate the interplay across levels.

---------------------------------

## RESULTS

### 1. Basic Building Blocks of Conversation – The Turn-taking Machine (Low-level)

What distinguishes conversation from other, more structured forms of interaction is that there is no predetermined order of who should speak, about what, and for how long. Given this precarious starting point, it is something of a marvel that conversation should work so well. This coordinated exchange between speakers rests on a complex system of turn-taking.

The turn-taking system has many parts, but three basic components are (1) turn exchange, how people manage to pass the floor back and forth in an orderly and efficient manner; (2) turn duration, how long speakers talk for before they turn over the floor; and (3) back-channel feedback, the active engagement displayed by listeners while speakers are talking, such as the use of nods or short utterances—"mhm," "yeah," "exactly"—to convey understanding and encouragement. The scope of our data permits close investigation of these basic features of conversation.

### 1.1. Turn Exchange

One important finding related to turn exchange has been that the average interval between turns is a relatively brief 200 milliseconds, a figure that appears consistent across languages and cultures (Stivers et al, 2009; see also, Levinson, 2016). Here, we replicate this prior work in a large corpus of video-mediated conversation. Following Heldner and Edlund



(2010), we applied a procedure for classifying communication states to our entire corpus to obtain a time series in which the presence or absence of speech from each speaker is recorded at 10 ms intervals. This allowed us to identify within-speaker and between-speaker intervals with high temporal precision. The resulting taxonomy included: *gaps* (between-speaker silences), *pauses* (within-speaker silences), *overlap* (between-speaker overlap), and *within-speaker overlap* (WSO), (when one speaker starts speaking in the middle of another speaker's turn, such as in the case of an attempted interruption or back-channel). Here, we focus on gaps and overlaps.

To deal with outliers, we removed between- and within-speaker intervals more than three standard deviations from the mean.[6] Upon visual inspection, we observed that these outliers were almost always due to technical issues, such as moments of poor internet connectivity, rather than genuine conversational anomalies.

Figure 2 shows that gaps and overlaps followed an approximately normal distribution centered on zero (equivalent to a perfectly timed, no-gap, no-overlap speaker transition). Gaps and overlaps were overwhelmingly less than 1 s long, and often much shorter, with a median of 380 ms and -410 ms, respectively. Gaps represented just over half (52.1%) of all speaker transitions, while just under half of speaker transitions were overlaps (47.9%). Overall, the median between-speaker interval was a fleeting 80 ms.

The brief interval between turns is particularly striking because the length of these intervals is much shorter than the time it takes a person to react and produce an utterance from scratch. This means that listeners must be *projecting* the end of a speaker's turn before it comes to an end. Indeed, prior research has identified mechanisms by which people accomplish this feat, for example by using various "turn-yielding" cues such as syntax and prosody (De Ruiter et al., 2006; Riest et al., 2015; see also, Bögels & Torreira, 2015).

---

[6] Pauses and WSOs cannot be lower than 0s. As such we effectively removed outliers >3 SD above the mean in the case of these two measures.

Overall, these figures closely match prior literature (see Heldner & Edlund, 2010; Levinson & Torreira, 2015), replicating earlier findings in a dataset containing nearly half a million speaker transitions. We also extend this literature to the increasingly important domain of video-mediated communication. At least at a basic mechanical level, certain conversational dynamics online closely resembled those seen in face-to-face interaction (see also, ten Bosch et al., 2005)

--------------------------------

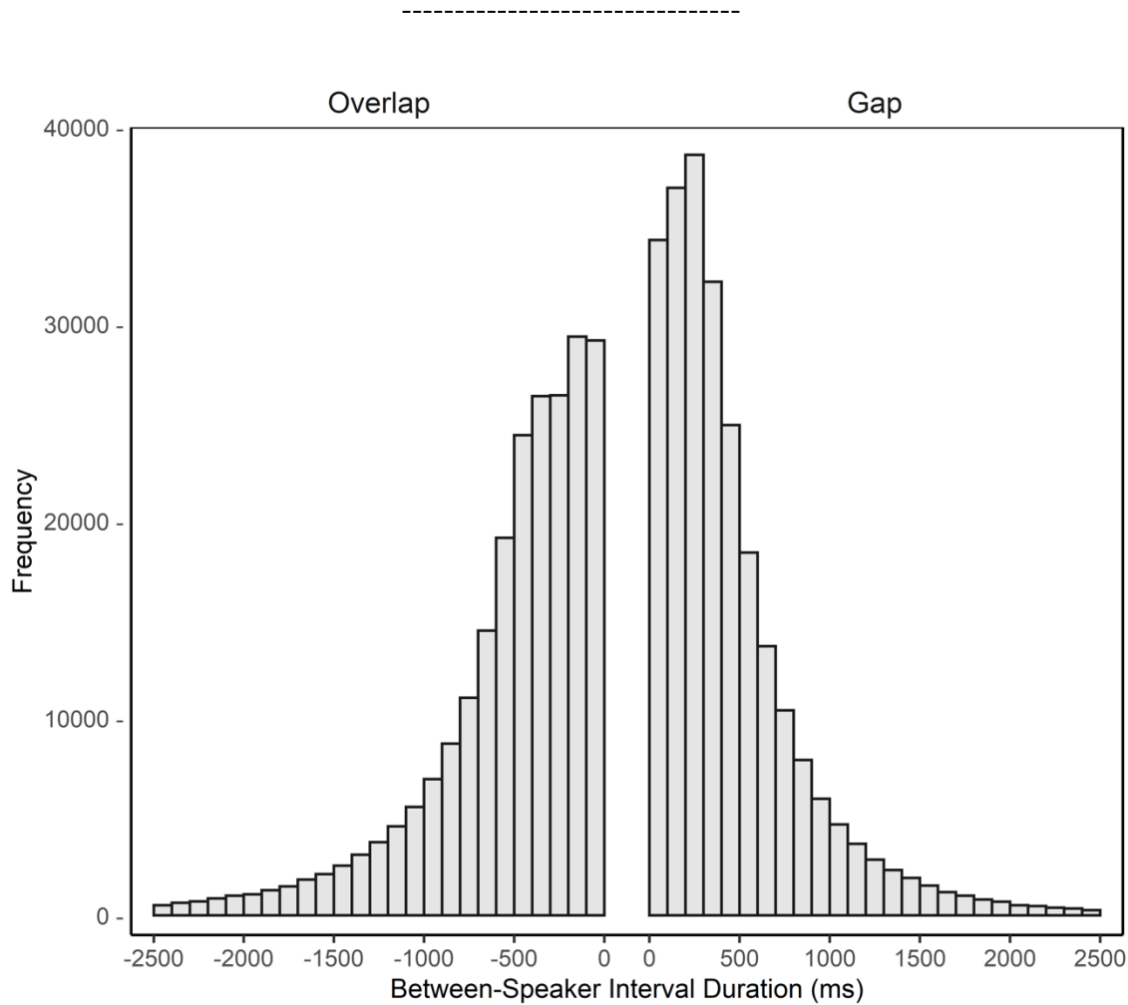

*Figure 2*. **Distribution of gaps and overlaps across speaker transitions.** Negative intervals are classified as "overlaps," indicating the presence of simultaneous speech by two successive conversation



participants. Positive values are "gaps" of silence. Despite differences in modality, results indicated that the median between-speaker turn interval was +80 ms and approximately normally distributed. These results are similar to those previously observed during in-person conversations across many cultures (Stivers et al, 2009).

----------------------------------

We return to the dynamics of turn exchange in Section 3, in which we examine the relationship between such low-level conversational features and higher-level psychological outcomes.

### 1.2. Turn Duration

Turn exchange, or passing the floor, is only one of many aspects of conversational mechanics. Turn *duration*, or how long people hold the floor, is another basic feature. Compared to the sizable literature on turn exchange, turn duration remains relatively understudied—due, in part, to a lack of available datasets of suitable quality and scale. To fill this gap, we present novel results on turn duration, based on a series of algorithms that we developed for segmenting transcripts into turns.

**Talk ratio.** Speaker turn duration can be conceptualized in two ways. The first way is to consider the overall time that a given speaker holds the floor. This, by extension, captures the floor ratio between conversation partners (e.g., Speaker A talked for 20 out of 30 minutes, resulting in a Speaker A:Speaker B "floor ratio" of 2:1). The second, more fine-grained approach is to measure the duration of *individual* turns.

The individual turn approach offers a richer portrait of conversation. For example, floor ratios necessarily fail to distinguish extended monologues from engaging, high-speed back-and-forth conversation. Finer-grained observation at the turn level has the potential to reshape active literatures that currently rely on coarser floor-ratio measurement, including on perceptions of leadership, likability, and gender dominance (Maclaren et al., 2020; Mast, 2001; 2002). Progress



in this direction, however, first requires grappling with a difficult definitional dilemma at the heart of large-scale conversation research: What do we mean by a turn?

**Individual Turn Duration.** To date, the most rigorous definition of conversational turns is likely found in the Conversation Analysis literature (and that of related disciplines), where researchers have developed rubrics for hand-transcribing conversations (e.g., Hepburn & Bolden, 2017). But large corpora such as ours, which includes hundreds of thousands of potential turn boundaries, makes conventional hand-coding impractical. Scaling these efforts computationally, however, would require the construction of a precisely defined series of steps—that is, an algorithm—for organizing two speakers' streams of speech into psychologically meaningful turns. Below, we demonstrate three candidate algorithms, or "turn models," and describe the design rationale and limitations of each. We regard these algorithms as useful, if early, starting points for continued research on the subject of conversational turns.

Our most elementary turn model, *Audiophile*, formalizes a simple assumption: a turn is what one speaker says until their partner speaks, at which point the partner's turn begins, and so on. This is essentially how the AWS Transcribe API parses recorded speech into turns. While useful as an easy-to-implement benchmark, Audiophile had significant drawbacks. For example, due to brief sounds (like laughter) or short bits of cross-talk, Audiophile often broke up turns too aggressively, creating several small turns that virtually any human observer would have regarded, syntactically and psychologically, as a single turn. However, as we demonstrate below, approximating "psychologically real" turns is a difficult task that has stymied research on turn duration—with some notable (and often hand-coded) exceptions, such as the convergence of partners' turn durations over time (e.g., Cappella & Planap, 1981).



In this section and the next, we describe the development of two competing computational models that better approximated human perceptions of turns. We begin with *Cliffhanger*, which attempts to more accurately capture the duration of people's turns by segmenting turns based on terminal punctuation marks (periods, question marks, and exclamation points), as generated by automated transcription. From Cliffhanger's "perspective," once the transcript indicates that Speaker A has begun speaking, if Speaker B interjects during A's sentence (i.e., before A reaches a terminal punctuation mark), then B's utterance is shifted into a new turn *after* A's sentence concludes.[7] Subsequent sentences by A are then assigned to a new turn, succeeding B's interruption. In essence, Cliffhanger disallows turn exchanges until after the primary speaker has finished their current sentence.

The simple example in Figure 3 makes the Cliffhanger procedure clear. As depicted, the mean and median Cliffhanger turn duration was 4x and 5x times greater, respectively, compared to Audiophile. The clear implication is that studies of turn duration will rely heavily on the researcher's choice of turn model. How might we empirically determine if Cliffhanger is a more suitable model for studying turn duration?

--------------------------------

[7] To be precise, Cliffhanger applied the same rule for questions and exclamations; any sort of terminal punctuation was treated as the same kind of boundary marker by the Cliffhanger algorithm.



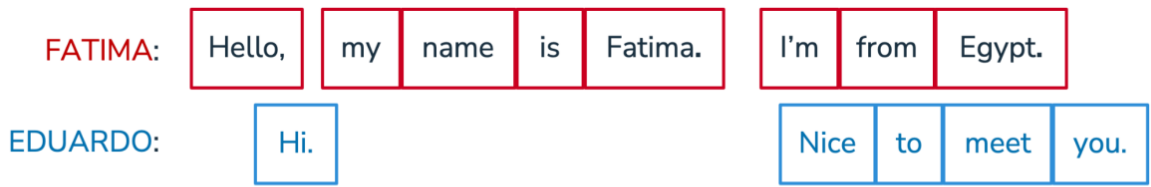

|  | Audiophile | |
|---|---|---|
| Hello, | | |
| | Hi. | |
| my name is Fatima. I'm | | |
| | Nice | |
| from | | |
| | to | |
| Egypt. | | |
| | meet you. | |

|  | Cliffhanger | |
|---|---|---|
| Hello, my name is Fatima. | | |
| | Hi. | |
| I'm from Egypt. | | |
| | Nice to meet you. | |

*Figure 3.* **A depiction of turn segmentation by Audiophile and Cliffhanger turn models.** The baseline Audiophile model regarded each interjection as initiating a new turn, disrupting the flow of the self-introduction by Fatima (red). In contrast, our improved Cliffhanger model presented an alternative organization of the same information into a more intuitive format in which Fatima and Eduardo (blue) alternate pleasantries.

-----------------------------------

-----------------------------------

| Model | Turn Duration | | Number of Words | | Average Turns |
|---|---|---|---|---|---|
| | Mean | Median | Mean | Median | |
| Audiophile | 2.22 | 0.92 | 6.40 | 2 | 440.70 |
| Cliffhanger | 8.52 | 5.81 | 17.81 | 9 | 159.41 |

*Table 3.* **Basic comparison of Audiophile and Cliffhanger turn models.** Speakers' mean and median turn durations were 4-5x greater for the Cliffhanger turn model compared to that of Audiophile. This hints at the possibility that analytic decisions about transcript segmentation will play a key role in the empirical investigation of speaking duration, a historically understudied topic.

-----------------------------------



One indirect test of a turn model's performance is its face validity, in terms of producing turns that align with human intuition. For example, anecdotal evidence suggests that turn length is correlated with the quality of a conversation: People who enjoy a conversation will likely have more to say, and thus on average, may be expected to take longer turns. In fact, we did observe this relationship between a speaker's turn duration and their overall enjoyment of the conversation—but only when using turns generated by Cliffhanger, not with Audiophile. For Audiophile turns, the relationship between people's median turn duration and their enjoyment was not significant ($b = 0.07$, 95% CI = [-0.04, 0.18], $t(3,255) = 1.32$, $p = 0.19$). By contrast, for Cliffhanger turns, the relationship between median turn duration and people's enjoyment was significant ($b = 0.06$, 95% CI = [0.03, 0.09], $t(3,255) = 4.12$, $p < .001$). At the same time, this analysis illustrates how results can depend critically on turn model selection, and in doing so introduces an important "researcher degree of freedom" (Simmons et al., 2011). This suggests that important directions for future work include improving turn models, developing guidelines for their use that are linked to specific research objectives, and demonstrating robustness of identified patterns across multiple turn models.

Over the course of this project, we developed and tested a wide array of algorithms for parsing naturalistic conversation into turns. No single model on its own was capable of handling all turn-related edge cases that occurred across the entire corpus. Rather, some models appeared to capture certain psychologically intuitive aspects while missing others. This suggests that either specific research questions will call for the use of specific turn models, or further development of turn models will eventually yield something close to an automated gold standard that can be reliably deployed across naturalistic conversation no matter the research question.

### 1.3. Back-channels



Back-channels—the short words and utterances that listeners use to signal speakers without taking over the floor (e.g., "yeah," "mhm," "exactly")—represent a third basic mechanical feature of conversation and a key component of the overall turn-taking system.

As illustrated in Table 3 above, the hyper-granular Audiophile approach to turn segmentation has limited utility for many research questions. For example, imagine Speaker A tells a two-minute story, during which Speaker B nods along and contributes simple back-channels such as "yeah" and "mhm" to demonstrate that they are engaged and paying attention. In a maximally granular formulation of turn exchange, such as Audiophile, each "yeah" is recorded as a distinct partner turn. Consider how, for certain research questions, this may be inappropriate because it permits the interruption of a continuous narrative based on small, routine interjections. To accommodate such research questions, we developed *Backbiter*, a turn model which allows the listener's encouraging "yeahs" to simply exist in parallel, rather than interrupting the storyteller's two-minute monologue. Such a model quantifies back-channels as informative, but nonetheless peripheral, annotations to primary speaking turns.

Backbiter creates two transcript entries for each speaking turn: first, the words uttered by that turn's active speaker; second, the back-channel phrases, if any, uttered by that turn's listener. Backbiter employs three basic rules to identify and reclassify utterances as *back-channel turns*: (1) a back-channel turn must be three words or fewer; (2) a back-channel turn must contain >50% back-channel words (e.g., "yeah," mhm," "exactly"); and finally (3) a back-channel turn must not start with a prohibited word, such as "I'm…". (See SOM for a complete list of back-channel words and non-back-channel beginnings.) Using these rules, the Backbiter algorithm moves utterances deemed as back-channel turns from the main turn registry into the turn's



accompanying back-channel registry. As shown in Figure 4, Backbiter altered the transcript

considerably through the successful identification and removal of back-channel turns.

----------------------------------

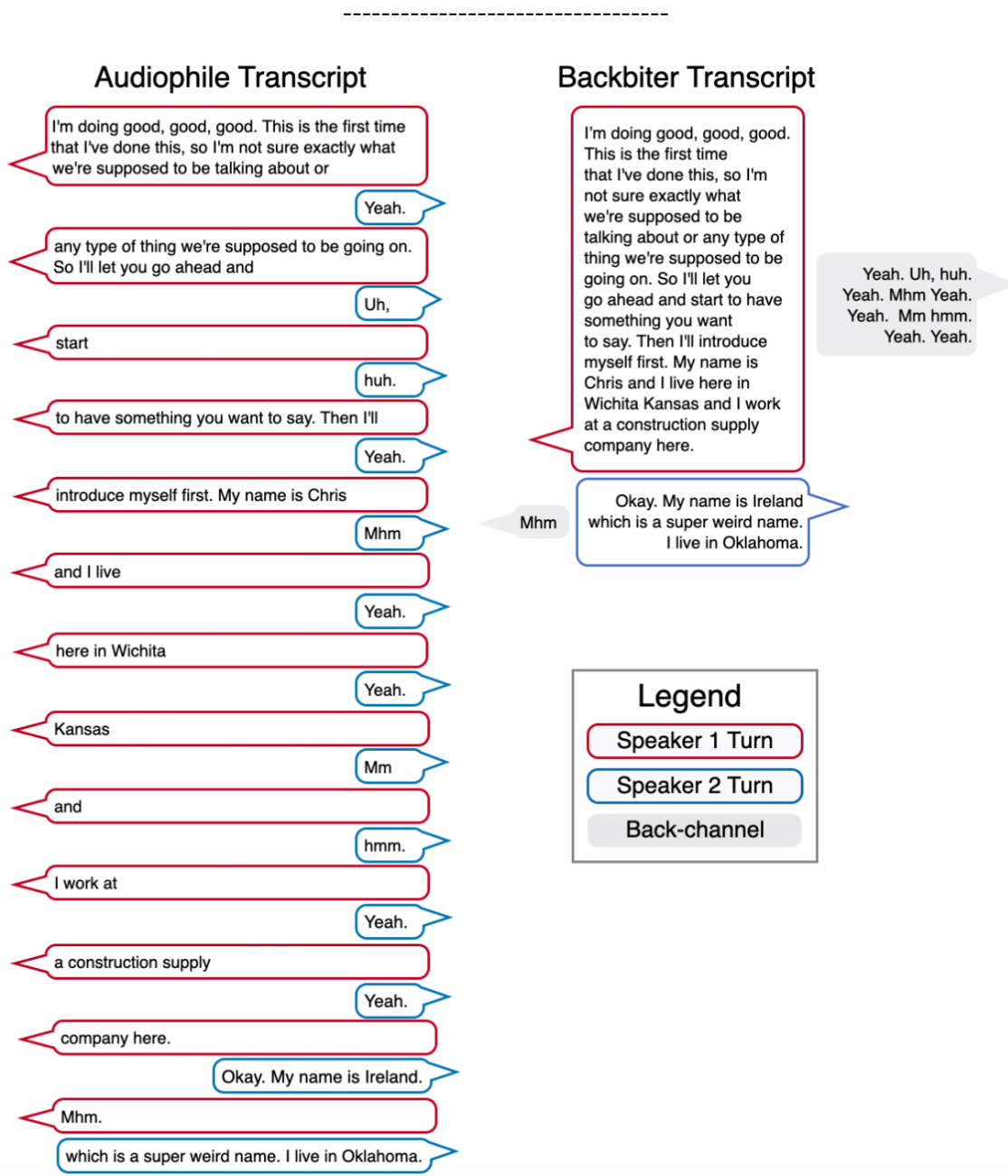

*Figure 4.* **Example transcripts from Audiophile and Backbiter turn models.** Audiophile regards each back-channel as initiating a new turn, disrupting the flow of the self-introduction by Speaker 1 (red). In contrast, our preferred Backbiter turn model presents an alternative organization of the same information into a more intuitive format in which Speaker 1 speaks a single introductory turn (while Speaker 2 is back-channeling). Speaker 1 then concludes that turn, yielding the floor to Speaker 2 (blue), at which



point Speaker 2 takes their first turn, also providing a self-introduction (while Speaker 1 occupies the back-channel).

----------------------------------

A single "mhm" may seem eminently forgettable, but when well-timed and especially in the aggregate, back-channels become essential conduits of understanding and affiliation (Bavelas & Gerwing, 2011; Bavelas et al., 2017). In our corpus, we observed that back-channels were universally deployed by listeners: 33.7% of speaker turns elicited at least one listener back-channel. This figure rose to 65.5% of speaker turns that were 5 words or longer; we estimate a rate of roughly 1,000 back-channel words per hour of spoken conversation.[8]  Notably, among many possible back-channel words, one reigned supreme: The word "yeah" alone accounted for nearly 40% of all back-channels, either in singular form ("yeah"), double, ("yeah, yeah"), triple ("yeah, yeah, yeah"), and beyond ("yeah, yeah…yeah, yeah, yeah"). (See Figure 5 for a distribution of back-channel frequencies.)

---------------------------------

---

[8] As we have noted, many participants in the corpus had more than one conversation. The back-channel rates were calculated by averaging at the level of the participant before averaging at the level of the corpus.



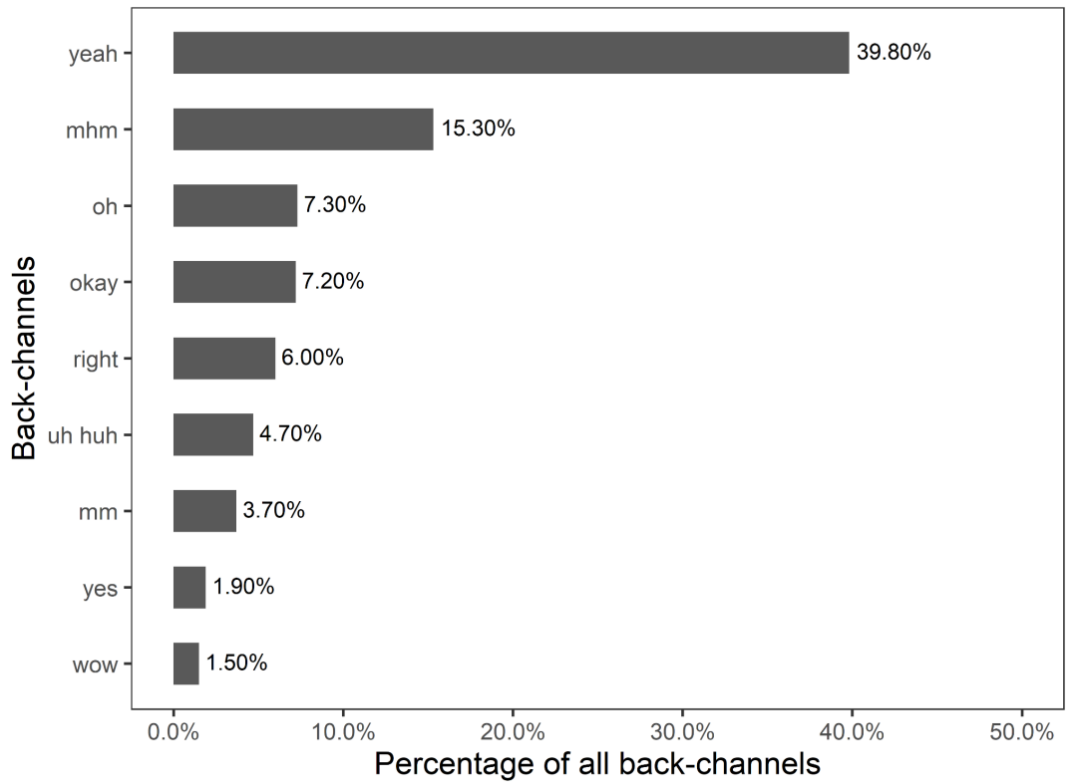

*Figure 5.* **Frequency of back-channel words across the corpus.** Back-channel words are a foundational element of conversation, occurring at roughly a rate of 1,000 such words per hour of speech; listeners deployed them in nearly two-thirds of speaker turns that were five words or longer. Generic continuers, such as "uh huh," help signal that a speaker may keep talking. In contrast, specific back-channel words, such as "wow," may convey context-appropriate content such as the mirroring of a speaker's emotion. Later analyses examine the frequency of back-channeling by different types of listeners and toward different types of speakers. This distribution reflects English spoken in the United States in the year 2020.

----------------------------------

It is worth noting that conversation researchers have identified various ways that not all back-channels are created equal. For example, one distinction is between "generic" and "specific" back-channels. Generic back-channels, sometimes referred to as "continuers," display understanding and function as a green light to the current speaker to continue talking (Enfield, 2017; Stivers, 2008). Specific back-channels, on the other hand, respond to the content of the current speaker's speech, often by mirroring an emotion of some sort, such as saying "yuck" at the climax of a disgusting story (Bavelas et al., 2000; Tolins & Tree, 2014). As such, while the



generic "yeah" may dominate our corpus by sheer frequency, the use of specific back-channels such as "wow" may actually be more important for building social connection or guiding the direction of a narrative. A second example is that while "mhm" may function to signal continued understanding, or "passive receptivity," as it has been called, researchers have also suggested that back-channels can signal "incipient speakership," or a listener's readiness to take the floor (e.g., Jefferson, 1993). Large-scale empirical investigation of the different functions of back-channels is lacking, but our corpus represents a resource that enables this line of research.

When considering all these different functions of back-channels, it is clear that simple continuers (e.g., "mhm") may be considered a low-level feature of conversation. But insofar as back-channels also serve more socially complex functions—fitting into a broader class of multimodal signals involving language, gesture, and intonation—they may warrant more focused attention as "mid-level" features in their own right, requiring more interpretation, layers of inference, and future research to improve detection and analysis

### *Summary – Basic Mechanics*

Using the corpus, we explored three basic mechanical features of conversation: turn exchange, turn duration, and back-channel feedback. In doing so, we replicated a key turn-interval finding and extended it to the domain of video chat. We also developed novel turn-segmentation algorithms for parsing speech streams into meaningful units of analysis. Finally, we explored back-channels—widely used listener signals of understanding and affiliation—and demonstrated how their automated extraction will likely facilitate further analysis and serve as a starting point for more sophisticated detection algorithms.

Going forward, researchers interested in the causes, correlates, and consequences of conversational turns will benefit from having explicitly defined turn-taking algorithms. Notably,



this move towards the use of automated algorithms in conversation research unlocks an exciting

benefit: In developing a common and explicit language for encoding conversational turns,

researchers can not only customize turn models to address specific questions but may share these

models with one another—facilitating replication and extension of research and allowing for

substantial efficiency gains. In what follows, we explore high-level features of the corpus,

focusing on the relationship between conversation and well-being.

## 2. The Primary Social Functions of Conversation (High-level)

We have established that turn-taking mechanics, a fundamental element of conversation

structure, can be observed and analyzed in our corpus. But why do humans take such pains to

closely coordinate their turn-exchange? Conversation has many goals, including the exchange of

information, but one primary objective is the formation and maintenance of social relationships

(Dunbar, 1998; 2004). This raises many questions: What is the relationship between conversation

and well-being? What social norms govern conversation? How do people understand themselves

as conversationalists? How do people form impressions of their conversation partners? What are

some of the biggest conversational mistakes people make and how might we fix them?

One core finding from past work is that social interaction, largely mediated through

conversation, plays a critical role in people's physical and mental health (e.g., Berry & Hansen,

1996; Clark &Watson, 1988; Hawkley & Cacioppo, 2010; Milek et al., 2018; Vittengl & Holt,

1998). Our corpus offers the opportunity to examine this idea at the earliest stages of relationship

development—people meeting for the first time—in well-preserved records and at a scale that

was hard to achieve before the ubiquity of video chat (and inexpensive online data storage). Our

corpus allowed us to not only determine how conversations with strangers increase well-being,

but also probe a set of puzzling recent findings about how people consistently underestimate how



much new conversation partners like them and enjoy their company (e.g., Boothby et al., 2018; Cooney et al., 2021; Epley & Schroeder, 2014; Mastroianni et al., 2021; Schroeder et al., 2021; Wolf et al., 2021). Finally, we exploit the breadth of our diverse sample to show how these effects vary across the lifespan.

## 2.1. Pre-Post Positive Affect

To explore the hedonic benefits of conversation, we had participants report their general mood immediately prior to their conversation, and then immediately after their conversation had ended, responding to "To what extent do you feel positive feelings (e.g., good, pleasant, happy) or negative feelings (e.g., bad, unpleasant, unhappy) right now?"

A mixed-effects model with random intercepts for participant and conversation revealed that post-conversation affect ($M = 7.32$) was significantly greater than their pre-conversation affect ($M = 6.12$, $b = 1.20$, 95% CI = [1.15, 1.25], $t(4085) = 44.43$, $p < .001$). This pre-post affective benefit held true across age groups (p's $< .001$, see Figure 6). In other words, at least for individuals that voluntarily opted into a conversation with a stranger, the act significantly boosted positive affect, consistently across the lifespan. Overall, the corpus provides a powerful estimate of the effect of conversation on well-being.

--------------------------------



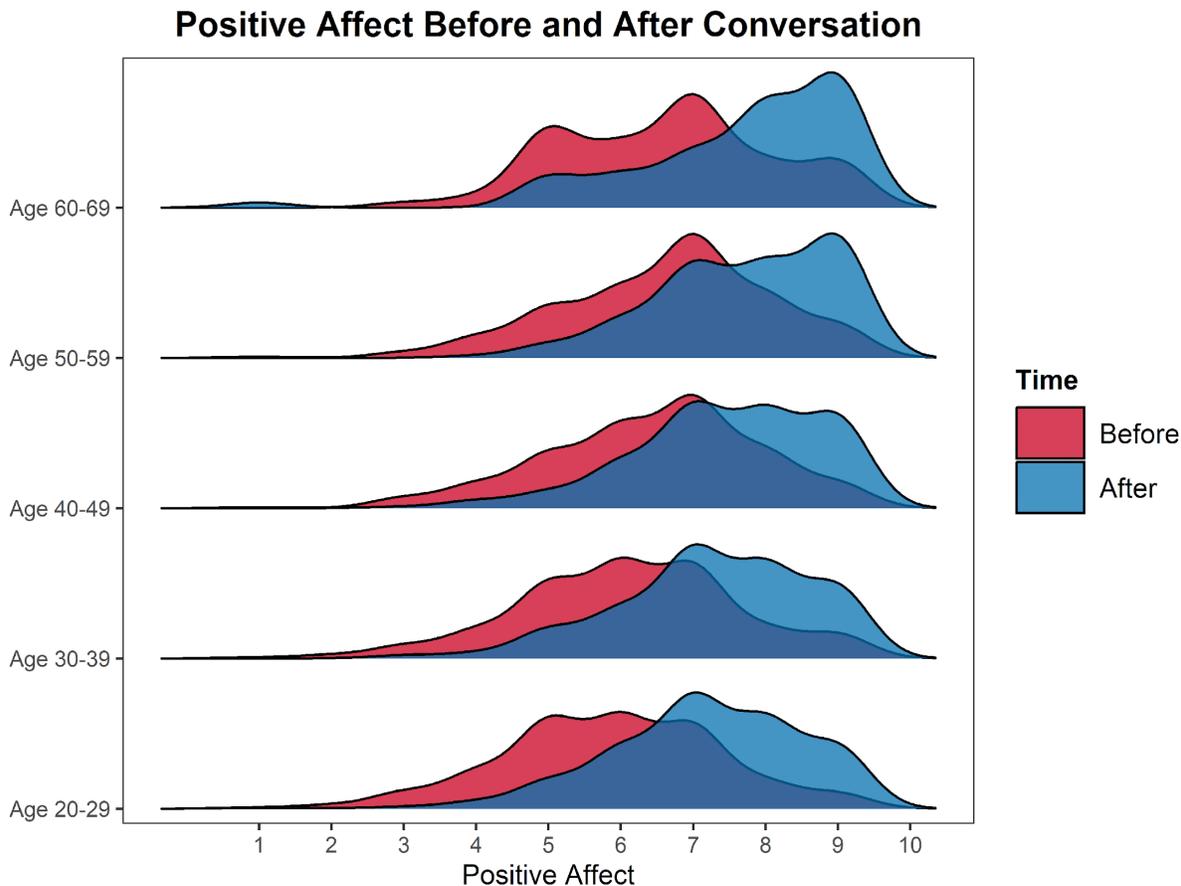

*Figure 6.* **Positive affect is significantly higher after a conversation, compared to before.** Each row of density plots corresponds to an age group. Respondents were asked to report their mood immediately before (red) and after (blue) the conversation. The impact of conversation on people's mood is positive, significant, and of considerable magnitude. This effect does not appear to be moderated by age, contrasting sharply with results (reported below) on highly age-dependent misperceptions of partner enjoyment.

--------------------------------

### 2.2. The Liking Gap

While unscripted conversations with a stranger seemed to noticeably brighten a person's mood—at least for those who opt into them—prior research has shown that this phenomenon both (1) contradicts participants' own intuitions and (2) is not fully recognized by participants as it occurs (for a review see, Epley, et al., in press). Indeed, when people were asked about future conversations with a stranger, such as the conversations studied here, they consistently (and



incorrectly) anticipated that their conversations would be less interesting, enjoyable, and valuable than they actually were (Epley & Schroeder, 2014; Schroeder et al., 2021).

This surprising pessimism, contradicted by participants' own experiences, manifests in people's post-conversation beliefs about their partners as well. For example, it has been shown that people systematically underestimate how much their conversation partners liked them and enjoyed their company after initial conversations, a failure of perception referred to as a "liking gap" (Boothby et al., 2018). This liking gap been documented in group conversations, cross-culturally, and in children as young as age 5, suggesting that people's concern for their own reputation may lead them to be critical of their conversational performance, and thus mistakenly projecting these self-critical thoughts onto their conversation partners (Boothby et al., 2018; Mastroianni et al., 2021; Wolf, et al., 2021). However, smaller sample sizes in prior work made it difficult to evaluate liking-gap heterogeneity in a consistent environment. Below, we use our corpus to both replicate prior work—providing still more evidence that well-established psychological phenomena extend to video-mediated interaction—and then exploit the scale of our corpus to explore how the liking gap varies across the lifespan.

Our post-conversation survey asked participants two questions, "How much did you like your conversation partner?" and "How much do you think your conversation partner liked you?" (endpoints: 1 - Not very much, 7 - Very much). A mixed-effects model with a random conversation intercept revealed that participants significantly underestimated how much they were liked by their conversation partners ($M = 5.31$), relative to how much they were actually liked ($M = 6.00$, $b = -0.64$, 95% CI = [-0.68, -0.60], $t(4862) = -27.83$, $p < .001$). A difference between these two estimates revealed the magnitude of a participant's liking gap.



The liking gap varied considerably by age. Adding age as a moderator revealed a significant interaction ($b = 0.01$, 95% CI = [0.005, 0.014], $t(4834) = 4.74$, $p < .001$). Figure 7 shows that as people get older, their judgments about how much their conversation partners liked them grew steadily more accurate. The causal factors behind this change with age present an opportunity for future research.

----------------------------------

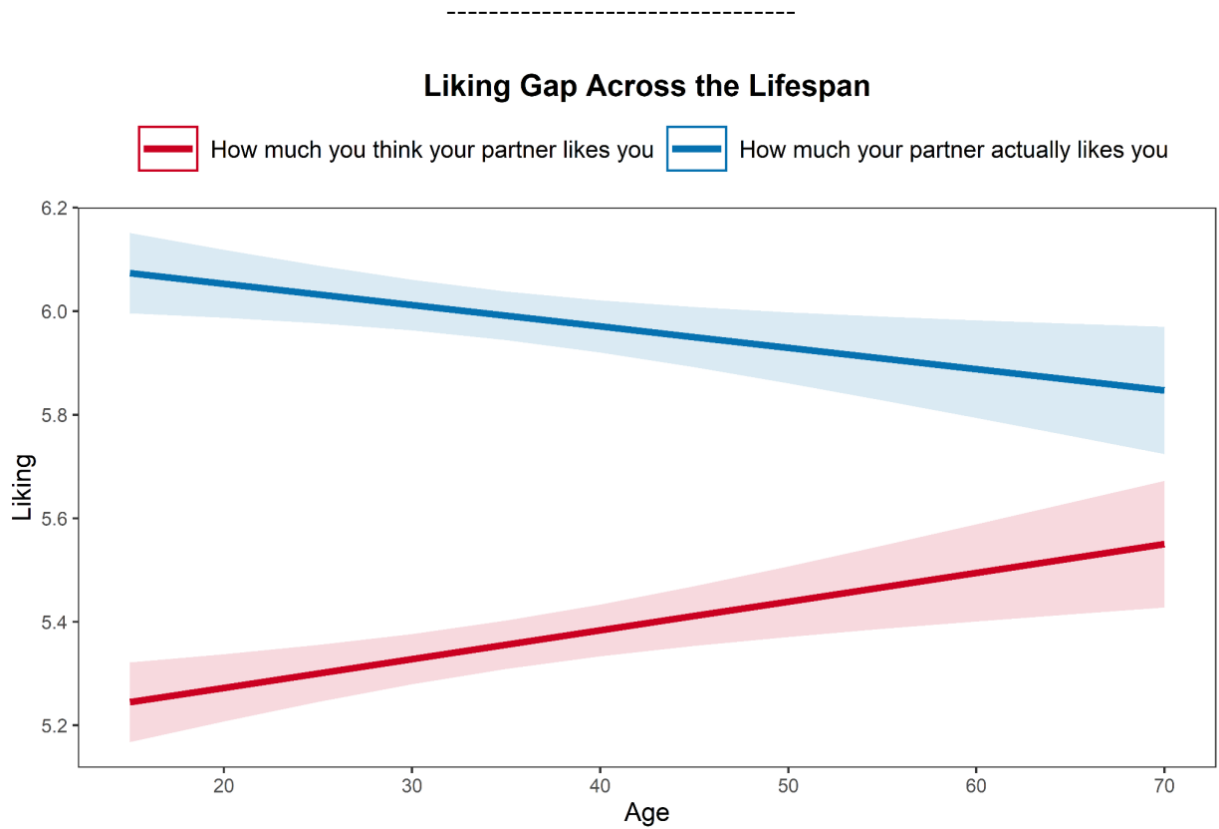

**Liking Gap Across the Lifespan**

*Figure 7.* **Conversation participants underestimate how much they are liked by their partners, but the magnitude of this pessimism declines sharply among older participants.** The *y* axis represents a judgment about the focal participant. The red line indicates how much the average focal participant thinks they are liked by their conversational partner, estimated as a linear function of the focal respondent's age (*x* axis). The blue line indicates how much the average focal participant is actually liked by their partner, which is substantially more. This "liking gap" indicates that participants are mistakenly pessimistic about how they are perceived. The magnitude of the liking gap shrinks among older participants but remains significant.

----------------------------------

Overall, our analyses revealed that even initial conversations with strangers can have a powerful effect on people's well-being. Our data also highlighted a puzzle: Despite reporting a



significant post-conversation boost in positive affect, people nevertheless walked away feeling relatively unliked by their conversation partners—an unfortunate pessimism belied by the fact that their partners actually liked them significantly more than believed. The positive effect of conversation on well-being appeared stable across the lifespan, in sharp contrast with the pessimism characterized by the liking gap, which decreased with age. This extension of prior work, revealing previously unknown patterns of heterogeneity, is one of many examples illustrating how a large-scale public dataset such as our corpus can shed new light on the unexplored social cognition that arises during the formation and maintenance of social connections.

## 3. Conversation Analysis Connecting Multiple Strata (Low-level & High-level)

The previous two sections demonstrated the range of the corpus, respectively covering low-level mechanical features of conversation, such as turn-taking, and higher-level outcomes, such as people's overall enjoyment and well-being. The breadth of the corpus uniquely enables analyses that bridge multiple levels, presented here and in Section 4.

A limited body of work has probed links across the conversational hierarchy, such as influential research on how listener back-channels determine the quality and trajectory of a speaker's story (e.g., Bavelas et al., 2000). But such work is the exception rather than the norm, and countless empirical questions remain unanswered. This has partly been due to the difficulty of collecting conversational data, inadequate computational methods for quantifying features at scale, and the lack of cross-disciplinary collaboration. Below, we demonstrate how the rapid disappearance of these obstacles has now placed this class of research within reach.

*Turn Exchange and Conversational Enjoyment*



In an attempt to link the lower-level mechanics of conversation with higher-level psychological outcomes, recent research has started to investigate how the interval between people's turns may function as an honest signal of whether two conversation partners "click" and enjoy each other's company (Templeton et al., 2022). Here, we replicate and extend these prior results before turning to a more comprehensive cross-level analysis in Section 4.

Following the measurement approach of Section 1.1 (Heldner & Edlund, 2010), we define a person's average turn interval based on the durations between their conversation partner's turn endings and their own turn beginnings. Participants reported how much they enjoyed the conversation on a 9-point Likert scale (end points: "1 - Not At All" and "9 - Extremely"). We then regressed how much people's partners enjoyed the conversation on people's mean turn interval. This analysis revealed that as the mean turn interval decreased, partner enjoyment increased ($b = -0.73$, 95% CI = [-0.92, -0.53], $t(3255) = -7.17$, $p < .001$). The same analysis using *median* turn interval revealed a similar result: ($b = -0.69$, 95% CI = [-0.91, -0.47], $t(3255) = -6.07$, $p < .001$). In other words, the faster people responded when taking over the floor, the more their partners enjoyed the conversation, consistent with Templeton et al. (2022). However, per our earlier discussion (see Section 1), turn intervals come in two different types—gaps and overlaps—which are conceptually, and likely psychologically, distinct. We can make use of this information to expand upon the relationship between turn intervals and conversational enjoyment.

By indicating in our model whether a person's average turn interval was a gap or an overlap, we were able to examine whether the relationship between turn interval and enjoyment is different for people who are, on average, "gappers" versus "overlappers." We found that the relationship between turn interval and partner enjoyment was moderated by a significant turn



interval x interval type (i.e., gap vs overlap) interaction ($b$ = 0.74, 95% CI = [0.13, 1.36], $t$(3253) = 2.38, $p$ = .017. Post-estimation analysis revealed that for people whose average interval was an overlap, there was no discernible relationship between the duration of their overlaps and their partners' enjoyment ($b$ = -0.26, 95% CI = [-0.71, 0.20], $t$(3253) = 1.11, $p$ = .27). On the other hand, for those whose average interval was a gap, there was a significant negative relationship between gap duration and partner enjoyment ($b$ = -1.00, 95% CI = [-1.41, -0.59], $t$(3253) = -4.76, $p$ < .001).[9]

In short, as shown in Figure 8, people's turn intervals were indeed related to how much their partners enjoyed the conversation. But it was not simply the case that the faster one responded, the more one's partner enjoyed the conversation; the effect appeared driven by the relationship between longer gaps and lower enjoyment.

---------------------------------

[9] Instead of aggregating people's turn intervals into an average for each conversation, we can model all of people's turn intervals across each conversation. To account for the structure of the data, we used a linear mixed effects model, with turn interval and interval type (e.g., gap or overlap) as fixed effects, and a random intercept and slope for participant ID. This analysis also revealed the same significant interval x interval type interaction (b = 0.016, 95% CI = [-0.03, 0.004], $t$(331854) = -2.58, $p$ < .01). This finding reinforces the main conclusion, although future may work to improve the modeling of within-person and within-conversation variance.



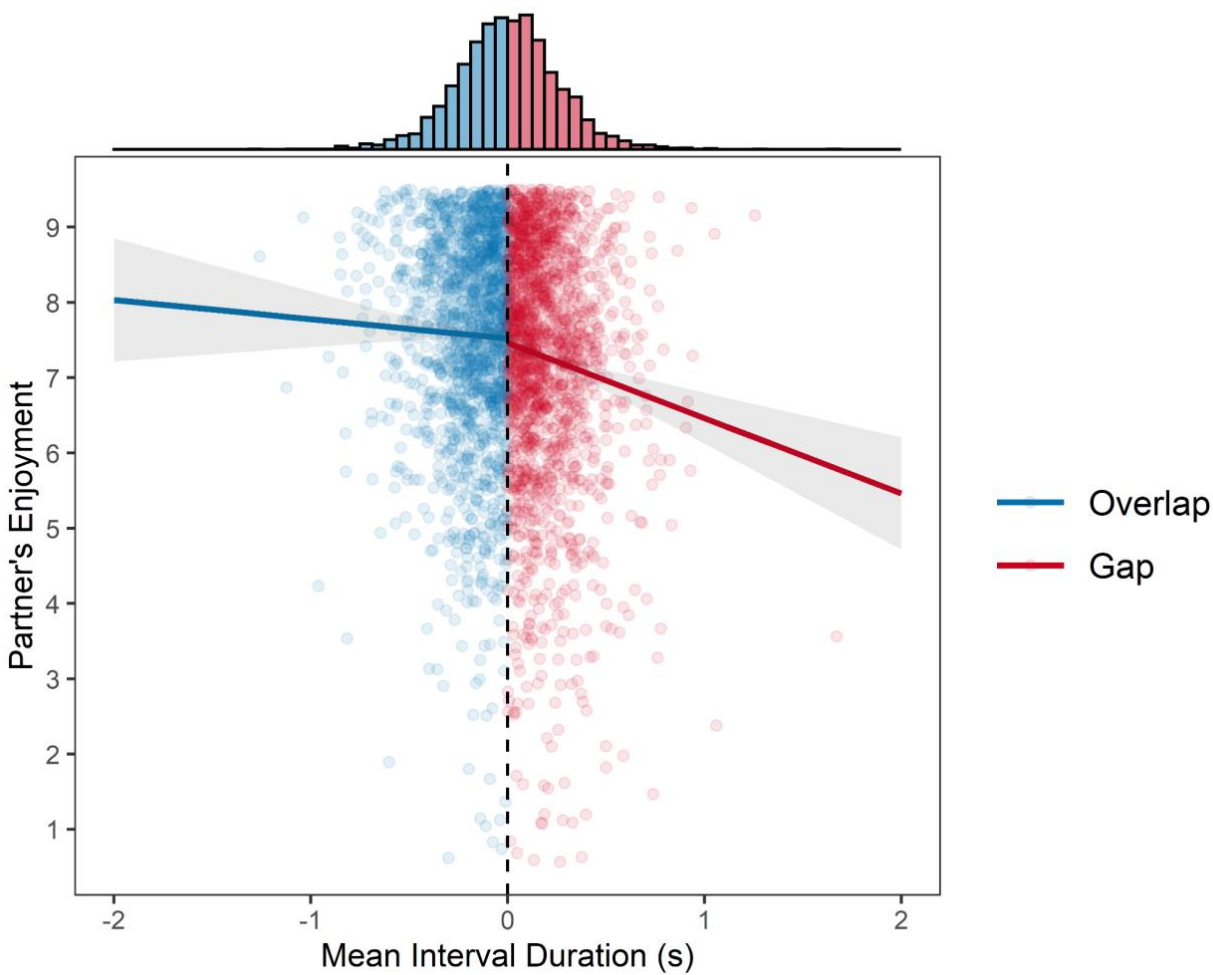

*Figure 8.* **Turn exchange is nonlinearly related to partner enjoyment.** The *x* axis indicates an individual's average interval between their conversation partner's turn ending and their turn beginning. Positive intervals indicate gaps between turns. Negative intervals indicate overlaps in speech near turn boundaries. We find that longer positive intervals (gaps) between turns are strongly negatively associated with overall partner enjoyment, but we find no relationship between enjoyment and negative intervals (overlaps).

---------------------------------

Many questions remain about the influence of low-level structural factors on people's high-level impressions of their conversations. One direction for future work is evaluating the stability of people's turn intervals, or the extent to which people's pattern of turn taking functions like a trait over time, by examining speakers who had multiple conversations. Future



research might also experimentally manipulate not only intervals, but also turn duration and other conversation mechanics, to assess their causal effects on enjoyment and other high-level judgments and impressions. Finally, while we used Heldner and Endlund's (2010) turn model for consistency with Section 1, an important new direction would be considering robustness to alternative segmentation algorithms (including but not limited to those we developed in Section 1).

## 4. What Distinguishes Good Conversationalists?

We now move from  the low-level structure of turns examined in Section 1 to the mid-level content of those turns. The breadth, scale, and detail of our corpus offers an unprecedentedly rich view of how conversation unfolds—moment to moment and turn by turn, through text, audio, and video modalities—across over 7 million words and 50,000 minutes of recordings. In this section, we introduce and analyze mid-level features like the semantic content of turns, dynamic vocal prosody, and facial expressions. These factors have historically been difficult to analyze due to the labor-intensiveness of annotation, the subjectivity of their perception, and the high-dimensional nature of textual and audiovisual data. However, recent advances in speech analysis and machine learning have allowed scholars to approximate these nuanced aspects of social interaction once deemed inaccessible to quantitative research.

In this section, we explore this rich "middle layer" of human interaction and link it to high-level impressions by answering an open question in conversation research: What distinguishes a good conversationalist?

We begin by using a suite of open-source, audio-processing and computer-vision models to extract detailed, high-frequency audiovisual information—features such as head pose, speech spectrum, and so forth—for each moment of our nearly 850-hour corpus. These fine-grained



measurements were then transformed into turn-level features, in many cases requiring an additional layer of inference. For example, our nod detection algorithm determined whether a particular pattern of up-and-down head movements occurred; our measures of prosody quantified a speaker's vocal pitch as flat or dynamic. More subjective mid-level features—such as whether a listener's face appeared "happy" or whether a speaker sounded "intense"—were computed with a series of machine-learning classifiers that we trained using human-annotated data from the RAVDESS (Livingstone & Russo, 2018) and AffectNet (Mollahosseini et al., 2019) datasets. Finally, we utilized pretrained text embedding models (e.g., Reimers & Gurevych, 2019; Song et al., 2020) to measure the semantic content of each turn and evaluate its similarity to the previous turn. All in all, our analysis characterized the linguistic, auditory, and visual content of 557,864 conversational turns (using the Backbiter turn model, see Section 1.3) along 19 dimensions (see SOM for full details).

Applying this workflow to our corpus allowed us to study variation in conversation dynamics at a depth and scale far larger than previous work. To illustrate, we conduct an analysis of how good and bad conversationalists—as evaluated by their partners—communicate. Our analyses revealed substantial differences in semantic novelty, vocal dynamism, and facial engagement. We also identified numerous avenues where additional human annotation, refinement of computational models, or application of domain-transfer techniques may help advance the study of conversation.

**Characteristics of Good and Bad Conversationalists**

After a conversation concluded, each participant was asked to rate their partner as follows: "Imagine you were to rank the last 100 people you had a conversation with according to how good of a conversationalist they are. '0' is the least good conversationalist you've talked to.



'50' is right in the middle. '100' is the best conversationalist. Where would you rank the person that you just talked to on this scale?" In general, participants reported their conversation partners as above-average conversationalists (*mean* = 73.0, *SD* = 20.1). For simplicity of exposition, we defined "good" and "bad" as the top and bottom quartiles of partner-rated conversationalist scores. Main-text results focus on the contrast between these groups; complete results on "middling" quartiles (25–50th and 50–75th percentiles) are provided in the SOM.

Next, we explored how the behaviors of good conversationalists differed from those of bad conversationalists across a suite of text, audio, and video characteristics (see SOM for full details). Here we present results for six turn-level features that illustrate the breadth of the analysis: (a) speech rate, i.e., words per second; (b) the semantic novelty of a speaker's current turn compared to their partner's previous turn; (c) loudness; (d) vocal intensity; (e) nodding "yes" and shaking "no" while listening; and (f) happy facial expressions while listening. Each of these features is depicted in one panel of Figure 9, with the top, middle, and bottom rows respectively representing text, audio, and video modalities. The left column—speech rate, loudness, and head movement—are features that are, at least in principle, directly observable. By contrast, features in the right column—semantic novelty, vocal intensity, and face happiness— are more complex, subjective "mid-level" concepts that require an additional layer of machine-learning inference to proxy.

People's speech and behavior varied considerably over the course of a half-hour conversation, and these complex patterns were difficult to capture with simple linear analysis. Instead, we examined whether good and bad conversationalists had different *distributions* across features. To facilitate presentation, we represent feature distributions as frequency plots, binned by deciles. For example, Figure 9(a) shows the frequency, indicated by the proportion of



conversational turns, at which good and bad conversationalists spoke very slowly (i.e., in the bottom 10% of speaking speeds), and so forth, up to the fastest speech rate (i.e., in the top 10% of speaking speeds). To quantify the general direction of trends, we also report differences in means; in doing so, we Winsorize all unbounded features at the 95% level to reduce the leverage of outliers that can arise due to issues in automated preprocessing. Turns were weighted to ensure that each conversation-participant contributed equally, so that results were not disproportionately driven by longer conversations. Standard errors were clustered by conversation to account for arbitrary dependence across turns and participants. We adjusted for multiple comparisons following Benjamini-Hochberg (1995), across all turn-level features analyzed; this adjustment also accounts for additional duration-adjusted and gender-specific robustness tests presented in the SOM.

This approach allowed us to examine how good and bad conversationalists varied in the proportion of turns they spent displaying more or less of any given feature. In what follows, our analysis, visualization, and interpretation of features follow this analytic approach across six selected features, revealing distinct patterns for how good and bad conversationalists engage with their partners.

### Speech Rate

We begin with speech rate, a low-level feature that is both straightforward to calculate and is linked to traits that are conceivably related to one's ability as a conversationalist, such as competence (Smith et al., 1975), persuasiveness (Miller et al., 1976), and intelligence (Murphy et al., 2003).

For each turn, we computed speech rate by dividing the count of words spoken in each turn by that turn's duration in seconds, yielding a rate of spoken words per second (WPS). For



analysis, WPS was binned into deciles, per the procedure outlined above. Figure 9(a) shows that good conversationalists spent more of their turns speaking quickly (i.e., in the upper five deciles). In contrast, bad conversationalists spent a greater proportion of turns speaking slowly (i.e., in the lower four deciles).

After adjusting for multiple comparisons, the speech-rate distributions of good and bad conversationalists differed significantly; the null of equal distributions was rejected at $p_{adj}<0.001$. In sum, Figure 9(a) shows good conversationalists spent more time speaking quickly, whereas bad conversationalists spent more time speaking slowly. On average, good conversationalists spoke at a rate of 0.1 WPS faster than bad conversationalists (95% CI [0.06, 0.14]). For comparison, the average speech rate across our corpus was only 3.3 WPS, indicating a 3% speedup.

--------------------------------



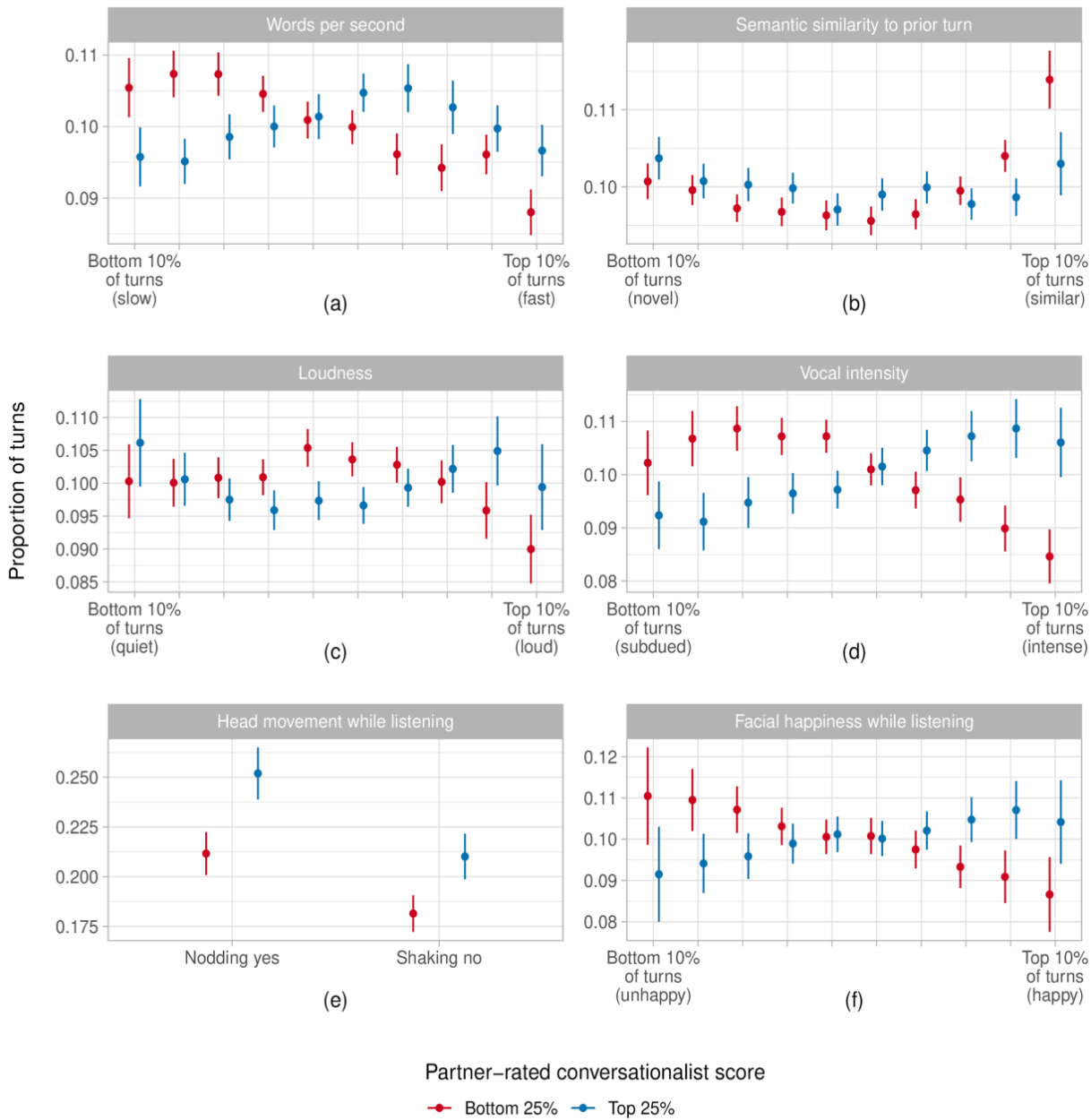

***Figure 9.* Behavior patterns of good and bad conversationalists.** Each panel depicts the engagement patterns of good conversationalists (top 25% of partner-rated conversationalist score, depicted in blue) and bad conversationalists (bottom 25%, depicted in red). Horizontal axes denote turn-level feature deciles. The y-axis indicates the average proportion of turns in a category for a good or bad conversationalist. Error bars represent 95% confidence intervals. Top, middle, and bottom rows respectively correspond to text, audio, and visual modalities; left and right columns include directly observable features and features that require an additional layer of machine learning to estimate, respectively.

----------------------------------



### *Semantic Similarity*

While speech rate is a straightforward property of a conversational turn (assuming a time-stamped transcript to start from), the semantic exchange between two speakers is a more nuanced and psychologically complex aspect of conversation.

Using machine-learning methods for dimension reduction, we computed a measure of semantic content based on text embeddings. Text embedding techniques numerically represent sentences or documents as vectors, based on co-occurrence patterns in a large training corpus; the resulting representations are widely used for measuring semantic distance between words, sentences, and documents (Kusner et al., 2015). This afforded a proxy of each speaker's novel "contribution" in any given turn, relative to the previous turn in a conversation.

Consider the following excerpt from a conversation in our corpus between two speakers that we will refer to as D and K. Following a relatively boring turn from D that threatens to stall the dialogue, K skillfully pivots into a fresh line of inquiry.

D: [talking blandly about the weather] "It's the same, I think it's the same here about 32 degrees."

K: [High semantic novelty response] "Yeah, exactly. Okay. It's going to seem like a totally random question, but being a Wisconsinite, how frequently do you attend fish fries?"

In this turn couplet, K responded to her partner's floundering statement about the weather with a novel question about attending Wisconsinite fish fries, resetting the conversation's momentum.

To study this computationally, we generated turn-level text embeddings from corpus transcripts with MPNet, a pre-trained language model which currently achieves top performance on a variety of linguistic tasks (Song et al., 2020; via the Sentence-Transformers Python module,



from Reimers & Gurevych, 2019). We then used the cosine similarity between the embedding

vectors of (1) the current turn and (2) the turn immediately prior, obtaining a proxy for the

degree of semantic novelty injected into the conversation.

Figure 9(b) shows that good and bad conversationalists differ significantly in the novel

semantic content of their turns (null hypothesis of equal distributions rejected at $p_{adj}$=0.001).

Results are robust to an alternative Euclidean distance metric, as well as to the widely used

RoBERTa embedding model (Liu et al., 2019); see SOM for details. However, as Figure 9(b)

makes clear, it is not the case that good conversationalists add more novelty to their turns across

the board; rather, they employ a mix of semantically novel and semantically similar turns.

This result warrants several caveats. The interpretation of high-dimensional text

embeddings is a notoriously complex subject, and results can vary significantly across

embedding models. For example, while the results we obtained from MPNet and RoBERTa

embeddings were broadly consistent in direction, they disagreed on specifics. MPNet results

suggested that good conversationalists differed most starkly from bad conversationalists

primarily in their reduced use of high-similarity turns, whereas RoBERTa results suggested that

the difference was primarily due to good conversationalists injecting more novel turns. We also

found that certain methods for measuring semantic distance appeared to depend heavily on the

number of words in current and prior turns, though importantly our primary specification and

results were robust to residualizing on word count. Finally, most embedding models are trained

to interpret the intricacies of language based on corpora of written documents, and not from

transcripts of spoken communication. As such, the rules that these models learn about language,

and the subsequent numerical representations they produce, may not be fully suitable for a

corpus such as ours, which is composed of entirely naturalistic conversations. Application of



domain-transfer techniques may help address this gap, though success may be difficult to

evaluate without extensive human annotation. For these reasons, we urge readers to exercise

caution in interpreting the association between semantic novelty and conversationalist quality.

We nevertheless find it encouraging that prior work finds pre-trained models can, in fact, achieve

near-human performance across a range of domains and hyperparameter choices (Rodriguez &

Spirling, 2002).

Despite these caveats, we report results on textual novelty due to their robustness across

multiple specifications and the apparent ability of unsupervised machine-learning techniques to

capture nuanced aspects of conversational skill that until recently could not have been studied

computationally.

### *Loudness*

While the exchange of semantic content plays a central role in conversation, a vast

literature has also established a similar importance in paralinguistic cues, such as vocal tone. The

low-level acoustic characteristics of speech can be quantified in many ways; we focus here on

loudness (as measured in decibels, or log-scaled vocal energy). It is worth noting that technically

"loudness" (the perceptual strength of sound) is what most people colloquially refer to as

"volume." Technically volume is the auditory sensation reflecting the *size* of sound, from small

to large, or less formally, the "bigness" or "spread" or "space-fillingness" of sound (Cabrera,

1999) (see Florentine 2011; and also, Marks & Florentine, 2010, for a loudness primer). Louder

speech is often effective in gaining the attention of a listener, though overuse can backfire (Page

& Balloun, 1978). Anecdotally, speakers who vary their volume for emphasis may be seen as

more dynamic speakers, whereas those with a uniformly loud or quiet voice may be perceived as

monotonous. To examine these patterns in our corpus, we computed per-turn loudness values



and used those to compute inter-turn variation in loudness (see Supplement for additional analyses of intra-turn modulation of loudness, as well as an analysis of pitch).

Our analysis, presented in Figure 9(c), revealed that good conversationalists and bad conversationalists differed significantly in the distributions of the loudness of their turns ($p_{adj}$=0.025). Differences in loudness distributions persisted when clipping the first and last seconds of a turn or when adjusting linearly for turn duration (see SOM for details). In additional analyses that were disaggregated by speaker gender, we found that these results were primarily driven by male speakers. Moreover, consistent with intuition, loudness patterns were *highly* nonlinear: in fact, we found no significant difference in the average turn loudness of good and bad conversationalists. Rather, bad conversationalists spend more time taking turns that are of middling loudness, whereas good conversationalists spend more time taking turns with either lower or higher average loudness—perhaps more adeptly matching the needs of the dialogue. In short, using one's voice to occupy a range of loudness values appears to be associated with conversational skill, but additional research is required.

### *Vocal Intensity*

Although loudness is appealing as an important and transparently measured, acoustic feature, humans often employ more complex combinations of vocal characteristics—including roughness, sibilance, or the contrast of lower and higher frequencies—to convey information such as emotion or happiness (e.g., Weidman et al, 2020). One such acoustic amalgam is emotional "intensity" (sometimes referred to as "activation"), a basic property of emotion and momentary affect (Diener et al., 1985; Frijda, 2017; Reisenzein, 1994). As with other subjectively perceived concepts, the precise definition of intensity is debated; prior work has related it to changes in one's body, how long the emotion lingers, how much it motivates action,



and whether it changes one's long-term beliefs (Sonnemans & Frijda, 1992). For our purposes, we sought to develop a measure of acoustic intensity, to test the hypothesis that spoken intensity differs among good and bad conversationalists.

To do so, we used the Ryerson Audio-Visual Database of Emotional Speech and Song (RAVDESS) (Livingstone & Russo, 2018) to train a vocal intensity classifier, and then applied this model to assign intensity scores for each speaking turn in our corpus. The RAVDESS dataset consists of recordings of trained actors who were prompted to read simple statements with either "normal" or "high" emotional intensity; we treated this intensity label for each recording (*normal* or *high*) as the response variable for training our classifier (a logistic regression model). Model predictors consisted of summary statistics for each RAVDESS recording, across a range of common prosodic features: mean, maximum, and standard deviation for fundamental frequency (F0) and volume (log-energy), as well as voiced and unvoiced duration (see SOM). We used our resulting trained classifier to predict vocal intensity for every 1 s interval in the corpus, then averaged these values within turns for a single intensity score per turn.

As revealed in Figure 9(d), people rated as good conversationalists spoke with more intensity than bad conversationalists; null hypothesis of equal distributions rejected at $p_{adj}<0.001$. These results were fairly linear (difference in means 1.3 percentage points in predicted probability of high intensity, 95% CI [0.007, 0.019]) and were also robust to an alternative specification that adjusted for turn duration. In analyses that disaggregate by speaker gender, differences persist among female speakers but lose significance among male speakers (see SOM). Whether vocal intensity registers for listeners as enlivening, emotional, or empathetic remains an open question.



Indeed, these results, like other model-based predictions for subjective mid-level features, are intended simply as a starting point for future work. We used a relatively simple model for the sake of exposition and computational efficiency, leaving room for considerable improvement through additional feature engineering, domain transfer techniques, and the use of more sophisticated classifiers. To enable continued refinement, we provide raw audio files for all conversations in the corpus, in addition to tabular records of the extracted features we selected for analysis. For replicability, our vocal intensity estimates are also provided. We further emphasize that concepts such as intensity are not merely unimodal. Beyond its manifestation in voice, intensity is also communicated by facial expression, word choice, and so forth—highlighting the need for continued research on the measurement of expressed emotion in conversation.

### Head Movement

In the cultural context of this American corpus, a commonplace nonverbal cue of assent or agreement is the head nod, an up-and-down "yes" movement of one's head. Similarly, shaking one's head laterally from side to side, "no," usually signals negation or disagreement. To capture highly informative head movement patterns from the corpus recordings, we developed an algorithmic "nod detector." Using facial recognition software in the Dlib C++ library (King, 2009), we computed a set of facial landmarks in order to identify the position of any human face detected on-screen. From there, we developed a rule-based schema to evaluate whether, over a 2s period, (a) at least 10% of a participant's face (b) crossed its starting position at least twice. When this occurred along the vertical axis, we recorded a "nod." If it occurred along the horizontal axis, it was recorded as a "shake." Binary summary features were computed



separately for nods and shakes, indicating the presence or absence of nodding and shaking at any point in the turn.

Results show that good conversationalists differed in both of these common nonverbal listening behaviors. Figure 9(e) depicts the rate of head nodding and head shaking among both groups. We find that good conversationalists were significantly more engaged not only in their rate of nodding "yes" (4.0-percentage-point increase; 95% CI [0.022, 0.059]; $p_{adj}$<0.001), but also in their rate of shaking "no" (3.0-percentage-point increase; 95% CI [0.013, 0.046]; $p_{adj}$=0.001). Notably, these results suggest that good conversationalists are not merely cheerful listeners, nodding supportively at each new contribution from their partners. Rather, they also make judicious use of nonverbal negations (head shakes) when deemed appropriate. This suggests that good conversationalists' head movement during conversation is best characterized as engagement, rather than just simple positivity.

As with our other results, we emphasize the need for continued research on the measurement of nonverbal engagement. Informal testing suggests that our nod detector demonstrates good precision (i.e., a low false positive rate), but weaker recall (i.e., more false negatives than desired). A more refined facial recognition algorithm may be able to detect more fine-grained head movements, as well as extract additional head-pose information (e.g., whether a listener's head is cocked to the side), and these and other improvements will undoubtedly improve performance on this and similar tasks of visual classification.

*Facial happiness.*

Finally, we turn to facial expressions, a more nuanced visual cue. Unlike nods, which consist of simple up-and-down movement, the space of possible facial expressions is large and can be difficult to interpret. To address this challenge, we used an emotion recognition model



pre-trained on the AffectNet corpus of facial expression images categorized into eight emotional

groups (Mollahosseini et al., 2019). For every second in the corpus, we provided speaker and

listener images to a convolutional neural network that assigned a probability to each emotional

label. It is important to note that perception of facial emotions is highly subjective, as evidenced

by AffectNet's low reported inter-coder agreement. This is both a technological and conceptual

issue, as people express the same emotion differently, often quickly transition between emotions,

and even combine aspects of different emotions in idiosyncratic ways (Barret, 2006; Hess et al.,

2016; Hamann & Canli, 2004). Moreover, the model was not adapted to our video conversation

context, where extreme facial contortion is rare and emotions such as contempt, disgust, fear,

sadness, and surprise appear to be virtually undetected. We sidestep these issues by examining

facial happiness alone, as happiness and neutrality are by far the most common expressions

detected in our corpus. In what follows, we present results on facial happiness while listening

(see SOM for an additional analysis of happiness while speaking).

 Figure 9(f) demonstrates that good conversationalists exhibited significantly more

(difference in mean predicted probability 3.5 percentage points, 95% CI [0.015, 0.056]) facial

happiness while listening, compared to bad conversationalists (null of equal distributions rejected

at level $p_{adj}$=0.045). These results were primarily driven by male speakers and lost statistical

significance in an analysis limited to female listeners (see SOM). As with results on head

movement, this finding highlights the role of engaged listening in differentiating good and bad

conversationalists.

 One important consideration in interpreting these results, and indeed in any analysis that

makes use of algorithmically generated inferences, is "algorithmic bias." Scholarship on

algorithmic fairness has identified numerous gaps in classification performance across racial,



gender, and other lines (Angwin et al., 2016; Buolamwini & Gebru, 2018), highlighting the need for context- and group-specific annotation to audit machine-generated labels. For example, in our corpus, model-based estimates of facial happiness were higher among female listeners (an estimated 39% of turns, compared to 33% for male listeners; $p<0.001$). One interpretation is that female participants did in fact visually express happiness at higher rates. At the same time, researchers must take seriously the possibility that (1) male participants expressed happiness in ways that were difficult for an image-based classifier to detect; or (2) the AffectNet training dataset disproportionately included images of men in serious contexts where happiness is less often visibly expressed. Gold-standard annotation of the corpus for expressed emotion and other related characteristics is an important direction for future work that will improve the interpretability of aggregate analyses, make it possible to audit the performance of machine-learning models, and allow corrections that enable principled cross-group comparisons.

### Summary

Across our corpus, we found that good conversationalists stood out on a number of directly measurable objective behaviors: they spoke more rapidly, showed greater variation in loudness across their speaking turns, and engaged in active listening through nonverbal cues (head nods and shakes). We also identified a number of related patterns among more nuanced and psychologically complex "mid-level" behaviors that required trained algorithms to detect: Good conversationalists injected more semantically novel content into their turns, exhibited greater vocal intensity while speaking, and increased facial happiness while listening. Together, these findings demonstrate the considerable potential of our corpus to explore conversation in new ways—especially across levels of analysis that past research has left largely unexamined.



**5. A Qualitative Glance at the Corpus - Topical, Relational, and Demographic Diversity**

While our report primarily focuses on empirical patterns, the corpus also offers a unique lens into American discourse in 2020. Consider that our corpus consists of conversations collected during a hotly contested Presidential election and a global pandemic. This makes the dataset a social repository of one of the most unusual stretches of time in recent memory.

As shown in Figure 10, using simple string matching, one can clearly see the rise of discourse about the election (keywords: "election", "biden", "trump", "republican", "democrat"), as well as the even more dramatic rise of covid as a topic of conversation as the pandemic gripped the globe (keywords: "covid", "pandemic", "vaccine", "mask"). Alongside these topics, one can also see the growing summertime focus on policing and police killings (keywords: "taylor", "floyd", "police"). Finally, note the near-universal inclination of people everywhere to want to talk about their family and kids (keywords: "my kids", "parents", "family"). In an effort to explore novel and understudied aspects of conversations and to articulate a larger structural framework, we left questions of topic choice relatively untouched, which we suspect will be a particularly fruitful direction for future research.

--------------------------------



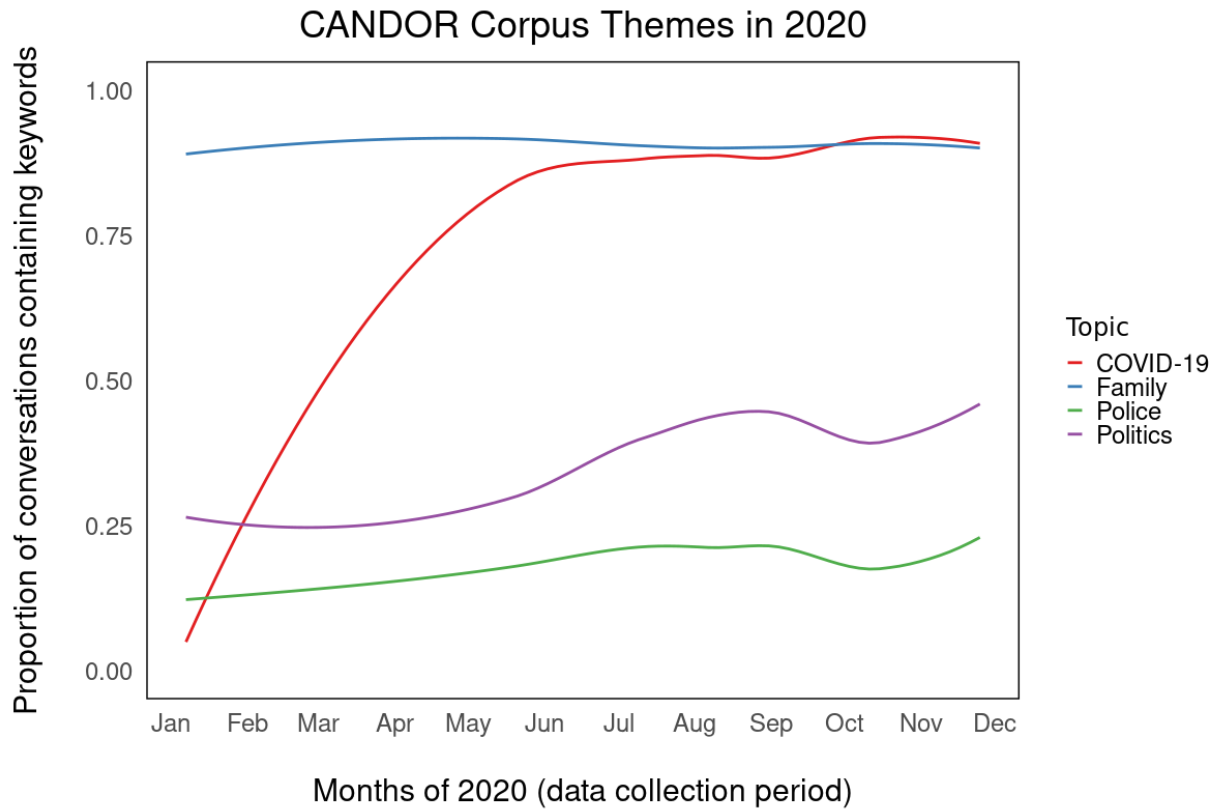

**Figure 10. Topic flow within CANDOR corpus.** Talking points in conversations between strangers, as measured in CANDOR transcripts by a simple keyword dictionary, trace the ebb and flow of societal issues in an unusually tumultuous year. COVID-19 (red) surged from unknown to the talk of the nation by mid-2020, matching or even exceeding family-related discussion (blue), a reliable staple of conversation. CANDOR frequencies for the presidential election (purple) and policing (green) highlight the trajectories of these nationally debated issues.

---------------------------------

No matter how much technology is applied, conversational discourse cannot be fully appreciated without actually *watching* people talk. Over the course of a year, a member of our research team engaged in a close qualitative examination of corpus's contents, watching every minute of the 850-hour corpus. Based on notes taken over the course of this monumental effort, we provide records describing, among other categories, conversations that were especially noteworthy, either for their awkwardness or for the astounding degree of rapport that emerged (see Qualitative Review). Such qualities, which are currently impossible to identify



computationally, immediately suggest new directions for research into the characteristics that distinguish these interactions.

Much qualitative research remains to be done. Several literatures, including Discourse Analysis and Conversation Analysis, have historically studied conversation though high-quality manual transcription, fine-grained annotation, and close reading of important conversations (e.g., Albert, & De Ruiter, 2018; Levinson, 2013; Sacks et al., 1978; Stivers, 2008; Stokoe, 2018; Tannen, 1991). This careful work has yielded many fundamental insights about conversation, many of which have had a deep impact on the analyses that appear in Section 1. In making the corpus public, our aim is not only to make further work in this vein feasible, but to benefit from the expertise of these literatures in refining turn-segmentation algorithms, establishing gold-standard measures for training of machine-learning models, and identifying new conversational phenomena.

An additional recommendation for future research is to explore the notable diversity of our participant pool, which represents a broad cross-section of the United States—particularly in cases where dissimilar participants were paired together. While our matching procedure that generated conversational pairs was not purposely randomized, it nevertheless generated a wide diversity of pairings. As such, our study facilitated countless inter-group conversations. For example, in one conversation our reviewer noted "[speakers are] a white man and a black woman, starts off very reticent and one-sided, rapport develops slowly with the help of a kind and talkative partner." In a different conversation, "[speakers are a] 40-year-old mother of 3 in Louisiana and 20-ish daughter of Polish immigrants located in Chicago. Mother is in recovery and was raised in the foster system. She is illiterate. Discussion about cycles of trauma and privilege and resilience." Watching such conversations, one is reminded that no matter how



advanced our computational techniques become, reducing conversations to tabular rows and columns of data will struggle to convey the full richness of their humanity.

The cross-demographic pairings in our corpus also offer a rare opportunity to study how people navigate a lengthy one-on-one interaction with an unfamiliar conversation partner, who is often someone that seems very different from them. Exploratory analyses revealed how people's conversational behaviors shifted when they were assigned to partners of differing age, gender, racial, educational, and political groups (see SOM for full results). These analyses revealed a number of notable behavioral differences, compared to conversations among demographically similar participants.

We found, for example, that older speakers—defined as belonging to the upper age tertile—tended to take significantly longer turns when talking to a younger conversation partner compared to when talking to another older speaker (1.5 s, 95% CI [1.1, 1.8], $p_{adj}<0.001$). This effect was large, amounting to a roughly 9% increase in floor time for older participants in age-mismatched conversations, compared to age-matched conversations. A second comparison revealed that among female participants, vocal expressiveness (measured as the standard deviation of vocal pitch) increased by 1.2 Hz when speaking with female partners, compared to a baseline standard deviation of 37.7 Hz when speaking to males (95% CI [0.6, 1.8], $p_{adj}=0.007$) (cf. Jurafsky et al., 2010). Finally, we observed that White participants used 15% fewer back-channels when paired with Black partners compared to when paired with White partners ($M=0.012$ fewer instances per second, 95% CI [-0.017, -0.008], $p_{adj}<0.001$). This aligns with prior research showing that White participants may often decrease nonverbal signals in interracial interactions (e.g., Dovidio et al., 2002).



We urge caution in interpreting these results, as apparent inter-group differences should not be attributed solely to the perceived out-group aspect of a conversation partner's identity. Social identities are complex and often contain components that are difficult to disentangle (see, Sen & Wasow, 2016; for a discussion of causal inferences about bundled identities). Nonetheless, our corpus does contain a wide diversity of dyadic pairings, and as such may lend itself to any number of interesting future lines of inquiry regarding inter-group communication.

## DISCUSSION

To guide our exploration of the CANDOR corpus, we divided conversational features into three levels: (1) lower-level objective features of conversation that are directly measurable at high frequency; (2) mid-level, psychologically rich features that can only be indirectly inferred thanks to advances in machine learning; and (3) higher-level subjective impressions from the speakers themselves as reflected in survey responses. We first explored these levels in isolation, examining low-level features such as turn-taking mechanics, followed by high-level features such as people's post-conversation well-being. We then used the corpus to draw connections across levels of analysis—an underutilized form of conversational research that opens numerous lines of inquiry, many of which require interdisciplinary collaboration. In doing so, we examined the relationship between people's speed of turn exchange and their partners' enjoyment; we further explored the rich "middle layer" of conversation, finding that various mid-level features— including facial emotions, semantic similarity, and acoustic intensity—were capable of distinguishing good conversationalists from bad ones.

In the most casual interpretation, these "levels" simply help to organize a vast, multi-featured dataset into convenient categories for reporting analyses that clearly belong to different



families of content. We propose, however, that this notion of a conversational hierarchy, in addition to its practical utility, may also prove fruitful in generating new theoretical insights. Below, we develop this framework more fully and offer it as a starting point for future discussion.

**Theoretical Considerations: Inference and information flow**

A lower-level conversational feature is a purely descriptive statement of record. The pitch of a speaker's voice, the presence or absence of eye contact, and time spent in silence are all examples of low-level features. Typically, these actions can be captured at a sub-second timescale. While it is true that the automated extraction of such features from a recording may require considerable feats of algorithmic inference (e.g., tracking on-screen gaze), these features nevertheless at least seek to capture an objective record of a conversation.

Moving upward through the hierarchy, different levels are characterized by the degree of indirect inference required and the breadth of contextual information used to make these inferences. The distinction we make between a mid-level versus a high-level inference therefore becomes a matter of scope. Humans, it seems, often employ a wide range of inputs to make their judgments (e.g., Whether Jill likes Jack is a highly indirect inference based on information across space, time, and textual/acoustic/visual modalities). Moreover, humans frequently, perhaps even rather helplessly, employ the full scope of their lived experiences to make sense of the present moment (e.g., "This person's voice reminds me of my dear Aunt Sally, whom I remember fondly"). Because subjective impressions and judgments about a conversation incorporate the broadest range of information and context, we distinguish them as *high-level* inferences.

In contrast, mid-level inferences are characterized by their use of a narrower scope of context and antecedent reference. Informally, they may dig deep, but not wide, to know what



they know. For example, language embeddings, which create numerical representations of the

semantic meaning of spoken words (See Section 4), are made possible because an underlying

statistical model was trained on a deep and extensive corpus of written language. While this kind

of inference is based on a wealth of previously encountered information—similar to human

judgment, in that regard—the context behind language embeddings is arguably deeper (i.e.,

billions of words of text training data) than it is wide. A hitch in the voice, a sad glance away; all

these signals, essential for shared understanding between humans, will go unnoticed by a

machine that knows only language. We thus refer to these inferences, which typically vary on the

timescale of a conversational turn, as mid-level inferences. This layer of the conversational

hierarchy thus operates somewhere between the objective immediacy of low-level events and the

subjective expanse of high-level human judgment.

Notably, humans, too, can make mid-level inferences in conversation, such as when some

aspect of the conversation carries a particular salience (e.g., a captivating facial expression), at

which point people often stop attending to the wider array of signals that normally influence their

ongoing impression formation. Similarly, but in the converse, the more algorithms are able to

account for context that once seemed solely in the domain of human capacity, the more they

seem eerily human, threatening to cross over the safety of the uncanny valley. Despite the

crossing of levels by human and machine, discretizing this continuum of inference into a middle

and higher level seems of theoretical and practical utility.

One unresolved question in conversation research relates to how information flows

*across* levels. It is clear that low-level, factual accounting of a conversation must necessarily

underlie higher-order sense-making about a conversation. From there, however, the trajectory of

inference remains an open area for future study.



One possible representation of the hierarchy is as a cascade of dependencies, with high-level judgments relying on mid-level inferences, which in turn draw on low-level behaviors to convey meaning. We invoked this model earlier, with an example of a smile (low-level), which may be a key component of what is perceived as a happy facial expression (mid-level), which in turn may serve as input into one's assessment of their conversation as enjoyable (high-level). But it seems equally plausible that low-level features may influence high-level impressions directly, without moving "through" a mid-level inference.

To complicate matters further, once formed, high-level impressions may percolate back down through the hierarchy. For example, after registering a perceived insult, a speaker's low-level behavior may change—elevating vocal pitch, increasing facial tension, or clipping speech. In turn, subsequent mid-level perceptions may be distorted, leading participants to draw differing conclusions about the same objective events. Scholars of conversation currently have little empirical basis for choosing among these intuitively plausible models for information flow, and their resistance to a simple accounting reflects the complexity of human conversation. Ultimately, information dynamics within these levels of conversation remains a subject in need of considerable future research.

As this discussion makes clear, the results we present only scratch the surface of the corpus. We regard our findings as an initial overture that will require additional efforts from many researchers across the social and computational science. Scholars from a variety of disciplines appear increasingly interested in the dynamics of conversation, and there are countless aspects of conversation that we did not begin to cover: common ground, information transmission, topic choice, dialog construction, and so forth. The raw material is there—and in



many cases, processed and ready to be analyzed. Below, we outline some key considerations for scholars pursuing this work.

**Practical considerations: Exploring the corpus**

During the construction of the CANDOR corpus, we encountered a number of interesting challenges, ranging from the technical (e.g., video alignment, feature engineering, emotion estimation), to the conceptual (e.g., segmentation of dialogue into psychologically sensible turns, representing conversations in terms of phenomenological "levels" of analysis). Throughout this report, we have sought to justify the decisions required to make analytic headway, while transparently reasoning through limitations, alternative approaches, and the potential downstream consequences. We invite readers to examine our choices (and omissions) and, ideally, improve upon them in future research.

*Politics and pandemics.* Our corpus consists of conversations collected during a contentious year in America, at the onset of a global pandemic. This makes the dataset a fascinating social repository of historical record. At the same time, as a reference for conversation science, it is important to consider the context, especially when asking certain questions, such as those related to politics, topic choice, feelings of social isolation, and so forth.

*W.E.I.R.D. sample.* Our participants were English speaking and resided in the United States, which represents the living conditions of only a fraction of the world's population (Henrich et al., 2010). Our sample also consists of people willing to talk to strangers online, which may be their "weirdest" feature of all. Ultimately, we will have to wait for other corpora to be released to make progress on important cross-cultural comparisons.

*Dyadic interaction.* If the science of dyadic conversation is incomplete, that is nothing compared to the lack of research on group conversation (Cooney et al., 2020; Moreland, 2010;



Stivers, 2021). Unfortunately, our dataset does not fill this gap, although it may serve as a starting point for future studies, for example, by establishing robust baseline values for dyadic conversation to contrast with future work on groups.

*Talking with strangers.* Our dataset consists of people who have never met before. At the very least, this means that certain conversational phenomena will be underrepresented. For example, gossip, which often functions to reinforce people's social bonds, occupies a considerable portion of talk time in everyday conversation (e.g., Dunbar, 2004; Jolly & Chang, 2021). Two strangers can certainly still "gossip" in the colloquial sense of talking about celebrities, but the gossip that truly dominates everyday conversation relates to mutual acquaintances who are not present. Similarly, our dataset might under-report a phenomenon like self-disclosure—which occurs more in intimate relationships, such as between close friends and significant others (e.g., Greene et al., 2006)—although our Qualitative Report documents a surprising number of deeply disclosing conversations.

*"Getting to know you" conversations*. Our participants were not given specific instructions; rather, they were simply told to have a conversation. This has the benefit of producing a corpus of naturalistic conversation that complements existing task-specific datasets. On the other hand, to investigate certain conversational phenomena, researchers may benefit from more structured forms of talk. For example, researchers interested in how people cooperate to create common ground and mutual understanding may be better served by conversations involving joint task completion (e.g., Wilkes-Gibbs & Clark, 1992).

*Video chat.* In 2020, the Covid-19 pandemic restricted in-person interaction, and catapulted video chat into a dominant communication medium. Much work remains to be done on the effects of video chat on perceived eye contact, facial expressions, turn-taking, impression



formation, and so forth. Questions remain about which behavioral patterns may be inherent to the digital medium in general, as opposed to dependent on specific aspects of the medium such as internet speed or camera resolution. While many phenomena may ultimately prove to be medium-independent (for example, turn-taking results in Section 1 were remarkably consistent with face-to-face results), scholars should be cautious in extrapolating to other contexts. Nevertheless, as society rethinks human communication—including a move toward more digital communication, remote work arrangements, and so forth—understanding video-mediated conversation is an increasingly important endeavor.

*Downtime.* At the beginning of each conversation, one person signed on before their partner. During this downtime, extraneous signals sometimes made it into the recording: background noise captured as speech, facial expressions, and even people talking to themselves. As such, when calculating aggregate statistics, we recommend using only the period when two speakers have already appeared (reasonably approximated by the start of the second transcript turn). Overall, when calculating statistics such as smiles per minute, caution should be exercised in the choice of denominator.

*Repeat speakers.* Our corpus contains numerous people who had more than one conversation: Of our 1,456 unique participants, more than half had multiple conversations; and roughly a third had three or more. This opens up many interesting questions, such as the variability of people's conversational behavior over time, how conversation partners adapt to one another, the stability of the impressions that people make, and so forth. We regard this as a particularly unique and exciting aspect of the corpus.

*Survey limitations.* In quantifying auditory and visual conversational behavior, we sought to capture all information that current technology would permit. In the future, better methods for



processing and analyzing audiovisual recordings may become available. In contrast, the post-conversation survey necessarily covers a fixed (though broad) set of questions.

   *Intra-conversational forecasting.* Across our findings, we focused on three types of analysis: (1) patterns of low-level conversational mechanics, such as turn dynamics; (2) high-level subjective outcomes, such as the link between conversation and well-being; and (3) relationships across the levels of the hierarchy, such as the use of vocal dynamism by good conversationalists. One untapped form of inquiry is what we refer to as *intra-conversational forecasting,* that is, how an intra-conversational feature such as a speaker's tone of voice might influence a listener's own reactions during that same turn or carried over into subsequent turns (consider how laughter can be infectious). The premise of intra-conversational forecasting—or the "flow of conversation" (Knox & Lucas, 2021)—is especially intriguing as a subject for multimodal analysis; for example, a sudden shift in a listener's facial expression (e.g., a frown or a fading smile) may prompt the speaker to change the topic or soften their tone. In an unpublished exploratory analysis, we found suggestive evidence that participants often fall into identifiable typologies of verbal and nonverbal "leading" and "following," such that one person consistently steered the semantic content or emotional tone of a conversation. The considerable range of possible questions, along with the complexity involved in parameterizing these kinds of temporal state inference models, makes intra-conversational forecasting a particularly rich idea for future research.

   *Additional labeling.* The more high-quality annotations (labels) that are added to the CANDOR corpus, the more useful its data will become. In its current state, the corpus already enables the development and training of sophisticated machine-learning models. With more comprehensive labels including better tracking of facial movements, along with enhanced



emotional and semantic inferences, the potential applications become endless: Consider, for example, how labeled datasets have recently been used in the computational study of linguistics to advance our understanding of decades old concepts such as politeness (Danescu-Niculescu-Mizil et al., 2013), or how multimodal deep learning architectures for emotion detection are increasingly used in conversational AI. We are excited to see what other characteristics might be quantifiable, and we encourage scholars to make their annotations and analysis available to the broader community.

*New Detectors.* While software libraries do exist that enable analysis of certain aspects of conversation, such as politeness (e.g., Yeomans et al., 2018), we remain in need of many more (e.g., models that detect self-disclosure, or high-accuracy laughter detectors). By releasing the entirety of the raw and processed corpus recordings, we anticipate the corpus will grow alongside advancing technologies.

*Diversity of Pairings.* One strength of the corpus that is worth reiterating is the diversity of pairings that exist, such as conversations between old and young, or conversations across gender, race, and political orientation (See Section 5). These pairings represent a significant opportunity for research on inter-group contact.

With the public release of the corpus, it is our hope that other teams of researchers will push it to new heights: reprocessing, labeling, and extracting more features; analyzing people's stability and variability across repeat conversations; examining how video mediated conversation differs from face-to-face; and of course, attending to the big questions that remain outstanding. Perhaps other researchers will release their own complementary datasets of groups, friends, or work colleagues, or those with social anxiety, or people talking across group divides. Ultimately



the expansion of this corpus, along with the release of more corpora, will allow accelerated progress toward a science of conversation.

## Conclusion

Over the course of the year 2020, nearly 1,500 people, ages 19-66, were paired to have recorded online video conversations. These recordings, which contain over 7 million words across 850+ hours, together make up the CANDOR corpus, which we have introduced in this report. The wealth of linguistic, acoustic, visual, behavioral, and textual data that comprise this corpus enable researchers across a number of scientific disciplines to open new lines of inquiry into the most fundamental of human social activities: the spoken conversation.

We strongly encourage people to take the time to actually *watch* these recordings. We imagine you will find, as we did, not only new ideas for future research, but also the striking power of conversation to connect people. Despite the awkward small talk, the differing politics, and the understandable reticence, at least initially, of strangers meeting for the first time—people nonetheless managed to come together, often with great kindness, grace, and understanding.

This orientation towards social connection is among the distinguishing features of our species, but it is also fundamental to the act of conversation itself. After all, conversation requires of its patrons a remarkable degree of cognitive and social interdependence, from the joint construction of dialogue to the inexorable search for common ground. This was a joy to watch—although it was clear to us that the real joy belonged to the speakers themselves, engaged as they were in the magic of building a shared experience through the spoken word. We should be grateful as scholars of human behavior that so much of this ancient ritual remains open to investigation.



## DATA & CODE AVAILABILITY

Data are available for access via registration here: https://betterup-data-requests.herokuapp.com/

Materials, code, and important links (e.g., analysis scripts, Data Dictionary, Qualitative Review, etc.) are here: https://osf.io/fbsgh/

## AUTHOR CONTRIBUTIONS

A. Reece and G. Cooney share first authorship. A. Reece, P. Bull, C. Fitzpatrick, C. Chung, and T. Glazer built the pipeline to recruit participants and process conversations, along with help from the rest of the DrivenData team. P. Bull, C. Fitzpatrick, T. Glazer and C. Chung developed turn-taking algorithms and feature extraction methods. A Reece and P. Bull designed and oversaw the final public release version of the dataset. G. Cooney and A. Reece designed the survey. G. Cooney and A. Reece provided overall direction for the analysis. A. Liebscher and S. Marin performed the analyses for Sections 1 and 3 with support from G. Cooney and A. Reece. G. Cooney and A. Liebscher performed the analyses for Section 2. D. Knox, G. Cooney, and A. Reece designed Section 4. D. Knox performed the analysis, created the figures, and wrote the Supplement for Section 4, with support from A. Reece, A. Liebscher, P. Bull, and G. Cooney. A. Liebscher and D. Knox created the pipeline to process the raw data into analysis-ready data, with support from A. Reece. S. Marin led the design and creation of the final figures with support from G. Cooney and A. Liebscher. B. Dawson performed the qualitative review. G. Cooney drafted the initial manuscript; G. Cooney, A. Reece, and D. Knox produced subsequent drafts, with support from the rest of the team.

## ACKNOWLEDGEMENTS

The authors wish to thank BetterUp, Inc.—Alexi Robichaux and Gabriella Kellerman in particular—for its sponsorship of this research and for BetterUp's willingness to share the data collected for research among the wider scientific community. We also thank the entire team at DrivenData for their considerable expertise in data science and machine intelligence, which made this entire undertaking possible.

*Funding statement:* This research was funded by BetterUp Inc.

## IRB APPROVAL

This study was approved by Ethical & Independent Review Services, protocol #19160-01.

## COMPETING INTERESTS

Authors with BetterUp affiliations were paid employees of BetterUp Inc., Authors with a DrivenData affiliation were employed by DrivenData Inc., DrivenData authors and G. Cooney were paid consultants at BetterUp Inc. at the time of this project.

## SUPPLEMENTAL ONLINE MATERIALS (SOM)

**Manuscript Section 1.1**

Conversations were transcribed with AWS Transcribe. Each transcription's most basic form was a list of individual tokens, accompanied by start and end timestamps, speaker IDs, and confidence estimates. The minimum temporal resolution was 10ms (0.01 s). Within each conversation and speaker, tokens were joined with adjacent tokens if 20ms or less of pause separated them. This output is considered the Heldner & Edlund transcript, where each row is considered a speaking turn.

Following this, the Heldner & Edlund (2010) communication state classification algorithm was applied to each transcript. This algorithm created, for each conversation, a time-series at 10ms increments of who, if anyone, was speaking at that moment in time. Using this time-series, a new dataset was created where each state transition was classified as either a Gap (between-speaker silence), Pause (within-speaker silence), Overlap (between-speaker overlap), or WSO (within-speaker overlap, an interruption). The units for each classification are durations in milliseconds. Overlaps are the only intervals which have values below 0ms.

**Manuscript Section 1.2**

```
TERMINAL_PUNC_CUES = [
    ".",
    "?",
    "!",
]
```

**Manuscript Section 1.3**

```
backchannel_CUES = [
    "a",
    "ah",
    "alright",
    "awesome",
```



```
    "cool",
    "dope",
    "e",
    "exactly",
    "god",
    "gotcha",
    "huh",
    "hmm",
    "mhm",
    "mm",
    "mmm",
    "nice",
    "oh",
    "okay",
    "really",
    "right",
    "sick",
    "sucks",
    "sure",
    "uh",
    "um",
    "wow",
    "yeah",
    "yep",
    "yes",
    "yup",
]

NOT_backchannel_CUES = [
    "and",
    "but",
    "i",
    "i'm",
    "it",
    "it's",
    "like",
    "so",
    "that",
    "that's",
    "we",
    "we're",
    "well",
    "you",
    "you're",
]
```



## Section 4: Mid-level Features & Distinguishing Good Conversationalists

Section A.1 begins by explaining how we computed turn-level audio, visual, and textual features. In A.2, we then describe our statistical procedure for assessing differences on these turn-level features among participants who varied in their partner-rated conversationalist scores; we also describe our procedure for multiple-testing adjustment of $p$ values. Section A.3 presents complete results for our analysis of partner-rated conversationalist scores, including (1) additional features not discussed in the main text; (2) patterns of results for the "middling" conversationalist groups, defined as those in the 25-50th and 50-75th percentiles of their partner-rated score; and (3) gender-specific and turn-duration-adjusted results.

### A.1. Obtaining Turn-level Features

In this section, we describe how continuous audio recordings and image frames were aggregated, based on start/stop times in a transcript segmented by speaking turns, into turn-level summary features describing speaker and listener behavior. We also describe how transcript-based summary features were extracted from a segmented transcript. We used the Backbiter turn segmentation model, although the same procedure could be employed with any speaker-attributed transcript with turn start/stop timestamps (for a discussion of turn segmentation, see Section 1 of the manuscript.)

### A.1.1. Transcript Features

Six transcript-based features were computed for each turn. Of these, four were straightforward summary statistics extracted from the segmented transcript.



- *Pause*: The difference between the end time of the prior turn and the start time of the current turn. This value is negative for turns that overlap with the previous speaker and positive for those that are preceded by a period of silence.

- *Duration*: The difference between the start and end times of the current turn.

- *Speech rate:* The number of words uttered during a turn, divided by the turn duration.

- *Back-channel rate*: The number of back-channeling events (See Section 1.3 of the Manuscript) by a listener during a turn, divided by the turn duration.

The remaining two textual features use pre-trained sentence embedding models to convert turn-level transcripts into a vector representation. Our main results are based on a sentence-level embeddings implementation (Reimers and Gurevych, 2019) of MPNet (Song et al., 2020), an embedding model which exhibits current state-of-the-art performance across a diverse set of language tasks. We used cosine similarity and Euclidean distance as two measures of semantic distance.

A common alternative measure of semantic distance, the dot product between two vectors, is identical to cosine similarity in our application because MPNet results are standardized to unit length. To evaluate the robustness of our results to different embedding models, Section A.3 reports comparative results using RoBERTa embeddings (Liu et al., 2019).

### A.1.1. Audio Features

Six audio features are reported for each turn, aggregating a variety of lower-level measures computed at various timescales: short-term (corresponding to 40-millisecond



intervals), medium-term (1 s intervals) and long-term (speaker turns of varying length). This aggregation proceeds in two steps.

First, short-term values for numerous low-level features were computed by summarizing the audio signal in rolling 40-millisecond windows. These low-level features included whether the window contained voiced speech, the fundamental vocal frequency of that speech (F0, measured in Hz), volume (log energy, proportional to decibels), and 14 Mel-frequency cepstral coefficients (MFCCs) that describe the shape of the power spectrum. Low-level feature extraction was conducted using the Python libraries librosa (McFee et al., 2015), Parselmouth (Jadoul, Thompson, & de Boer, 2018; Boersma & Weenink, 2021), pysptk (Kobayashi et al., 2017) and DisVoice (Dehak, Dumouchel, & Kenny, 2007; Vásquez-Correa et al., 2018).

These short-term auditory measures were aggregated at 1 s resolution. Average pitch and loudness were computed by taking the average non-missing values of all frames within a 1 s interval. This aggregation step helped address peculiarities in certain audio features, such as the fact that the fundamental frequency is undefined in windows of unvoiced speech (e.g., during the unvoiced sibilant /s/). In addition to these transparent averages of objective speech attributes, we also computed two model-based proxies of emotional expression—concepts which can be difficult to directly measure due to the subjective nature of their perception.

Because human annotation of speech is highly labor intensive, we utilized labeled data from the Ryerson Audio-Visual Database of Emotional Speech and Song (RAVDESS; Livingstone & Russo, 2018) to train a computational model that was subsequently applied to our corpus. The RAVDESS dataset contains recordings of 24 trained actors reading statements of varying emotional categories (e.g., calm, happy, sad, angry, fearful, surprise, disgust, or



neutrality), expressed with either normal or high emotional intensity. To estimate speech intensity, we computed a series of summary statistics—mean, maximum, and standard deviation for fundamental frequency, log energy, and voiced and unvoiced duration—for the short-term feature time-series within each 1 s interval in our corpus. Medium-term summary statistics were then input into a logistic regression trained on intensity labels and similarly featurized 1 s intervals from the RAVDESS corpus (described below). The resulting model predictions were used as a proxy for speech intensity in our corpus.

Finally, five medium-term measures were aggregated to long-term turn measures as follows.

- *Pitch*: Average of 1 s fundamental frequency values from speaker audio channel within each turn (includes all 1 s intervals from turn start to end)
- *Pitch variation*: Standard deviation of 1 s fundamental frequency values from speaker audio channel within turn (all 1 s intervals from turn start to end)
- *Loudness*: Average of speaker 1 s log energy values within turn
- *Loudness variation*: Standard deviation of 1 s log energy values within turn
- *Intensity*: Average of 1 s model-predicted intensity values within turn

### *A.1.1. Visual Features*

All visual features were captured at 1 s intervals. Three classes of visual features were extracted: head movement, gaze, and facial emotion. Each visual feature was computed for both listener and speaker. To capture the objective visual signals of head nodding/shaking and gaze, we developed our own algorithmic detectors. Using facial recognition software in the Dlib C++ library, we computed a set of facial landmarks (King, 2009). Our head nodding/shaking detector



then employed a manually tuned, rule-based approach that evaluated whether facial landmarks moved at least 10% of the total detected face size and crossed their starting position at least twice within two seconds. When this occurred along the vertical camera axis, we recorded a "nod," generally taken as a nonverbal signal of "yes." If it occurred along the horizontal axis, it was recorded as a "shake," typically indicating "no." Nods and shakes were aggregated to the turn level by computing the maximum across all 1 s intervals in the turn, indicating whether any nodding or shaking occurred. Second, to measure whether participants were gazing at the screen, we used eye landmarks to compute the proportion of pixels within the eye regions that were white. If this proportion fell between 0.22 to 0.45, we estimated gaze to be directed at the screen. We caution that this variable appears to be noisy and has not been tested for accuracy. Finally, to obtain a proxy for facial emotion, we used FastAI (Howard et al., 2018) to train an emotion recognition model on the AffectNet corpus of facial expression images (Mollahosseini, Hasani, & Mahoor, 2019). AffectNet categorizes facial expressions into eight emotional groups; given the low estimated incidence of facial emotions other than happiness (and neutrality), we extracted only the predicted probability of happiness.

- *Listener/speaker nodding yes*: Vertical movement of facial landmarks exceeding a manually tuned threshold

- *Listener/speaker shaking no*: Horizontal movement of facial landmarks exceeding a manually tuned threshold

- *Listener/speaker gazing on screen*: White pixel proportion within eye region exceeding a manually tuned threshold

- *Listener/speaker facial happiness*: Predicted probability obtained from AffectNet-trained neural network



*A.2. Statistical Methods*

In this section, we describe our procedure for assessing whether groups of participants diverge in their conversational behavior. The same procedures are employed both for analyzing speech patterns by (1) partner-rated conversationalist score and (2) partner identity (see Manuscript Section 5 and Supplement Section B).

Our primary analysis of conversationalist score compares the outermost quartiles; additional results are given in the Supplement for all four quartiles.

Below, we first describe how we tested the null hypothesis that a conversational feature Y, such as loudness, was distributed equally among the K groups—or that $f(y \mid X=x_k) = f(y \mid X=x_{k'})$ for all $k, k' \in \{1, \ldots, K\}$. In other words, we evaluate whether every group k uses highly intense speech at the same rate as every other group k'. This approach utilized the full distribution of each feature and as such was well suited to capturing nonlinearities often observed in conversational data. However, a key limitation was that it produced *p* values as a test statistic and must be interpreted primarily by visually comparing feature distributions. To report numerical differences, the following subsection describes how we analyzed differences in means, $E[Y \mid k] - E[Y \mid k']$, and produced confidence intervals. Finally, we describe our procedure for accounting for multiple significance testing.

*A.2.1. Assessing Differences in Distributions*

Two key statistical challenges arose in these analyses. First, conversational features exhibit clustered dependence within a participant-conversation unit and across participants within a conversation. For example, idiosyncrasies in microphone positioning might cause



speech from one participant to sound louder, or background noise might cause both participants to speak more loudly. Second, the number of observations (turns) within a cluster (conversations) can be influenced by the explanatory variables of interest (e.g., conversationalist score).

To test for differences in conversational patterns while accommodating these statistical issues, we first discretized each feature into deciles (i.e., quietest 10% of turns, turns between the $10^{th}$ and $20^{th}$ percentile on loudness, etc.) that captured much of the variation in how participants engaged with each other. That is, we represented each turn-level value, $Y_i$, with a one-hot encoding of the form $Y_i^* = [1, 0, 0, 0, 0, 0, 0, 0, 0, 0]$, here indicating a value of $Y_i$ for i=1, (i.e., in the lowest decile). We then conducted a test of equal category proportions using an asymptotic multivariate Gaussian approximation for the multinomial distribution. Specifically, we constructed a matrix of turn-level discretized feature values, $\mathbf{Y}^* = [Y_{1,\backslash 1}^{*T}, \ldots, Y_{N,\backslash 1}^{*T}]^T$, where $Y_{i,\backslash 1}^* = [Y_{i,2}^*, \ldots, Y_{i,K}^*]$ represented the one-hot encoding of $Y_i$ with the lowest reference category omitted (as category proportions sum to unity). Each turn was weighted to ensure that the total weight of each speaker in each conversation was equal, i.e., longer conversations received smaller weight on each turn. We then conducted a weighted multiple outcome regression of $\mathbf{Y}^*$ on group indicator variables $\mathbf{X}$, with $\mathbf{X} = [X_1, \ldots, X_N]$ and $X_i$ as a k-dimensional one-hot vector in which a positive entry in the $k^{th}$ position indicating the turn belonged to a participant with membership in group k. This produced a 9-dimensional vector of coefficient estimates for each group's categorical proportions, $\boldsymbol{\beta}_k = [\beta_{k,2}, \ldots, \beta_{k,10}]^T$, of turn features for that group (recalling that the omitted category proportion, $\beta_{k,1}$, sums to unity). Finally, we conducted an F test for the linear hypotheses of equality that coefficient vectors between each pair of groups—i.e., $\boldsymbol{\beta}_k = \boldsymbol{\beta}_{k'}$ for all k, k'—using an estimated variance-covariance



matrix clustered at the conversation level. This approach had the advantage of easily accommodating conversation-level clustering and turn-level weights; it carried the disadvantage of requiring discretization of continuous features, discarding information, and resulting in some loss of statistical power. Alternative approaches based on clustered rank-sum tests (e.g., Dutta & Datta, 2016), which do not discretize the data, offer greater statistical power but are computationally infeasible in large datasets like the one studied here.

*A.2.2. Assessing Differences in Winsorized Means*

While tests of distributional equality are well-suited for assessing nonlinear differences in speech patterns, the *p* values they produce do not provide insight about precisely where and how those differences arise. For this reason, we provide distributional plots that convey, for example, how bad conversationalists are more likely to speak in a moderately loud voice, whereas good conversationalists are more polarized between quiet and loud speech.

To aid interpretation, we also computed differences in central tendency, which were straightforward to summarize and facilitated the reporting of confidence regions. This was complicated by the fact that automated processing occasionally resulted in outliers that strongly distorted simple averages. For example, in some cases, slight errors in the start/stop timestamps of short turns produced outlying values speech-speed estimates due to division by near-zero values. Similarly, audio artifacts occasionally arose from non-speech events such as laptop movement, producing outlying values that commanded disproportionate leverage in subsequent analyses. To address these issues, we employed Winsorization, a technique commonly used in analyses of audio data, for all unbounded variables (Yale and Forsythe, 1976; Hilton et al., 2021). Winsorizing at the (arbitrarily determined) 95% level replaced extreme values outside the



2.5th and 97.5th percentiles with the values of the boundary percentiles themselves. Finally, we conducted linear regressions of the resulting trimmed features on group indicators, obtaining estimated differences in Winsorized means. As in our distributional tests, we clustered standard errors at the conversation level and weighted turns to ensure that each speaker-conversation contributed equally to our estimates.

### A.2.3. Multiple Testing Corrections

Our first set of mid-level analyses compared the best- and worst-rated conversationalists on 20 textual, auditory, and visual measures These tests resulted in 60 robustness analyses, as we repeated the same comparisons among subsets of female and male participants, as well as adjusting for turn duration. To control the false-discovery rate at conventional levels, we applied the multiple-testing correction of Benjamini and Hochberg (1995) within each study. All reported $p$ values were inflated by a corrective factor, ensuring they can be interpreted as usual (e.g., with reference to a 0.05 significance level) rather than utilizing a modified significance threshold.

### A.3. Results from Descriptive Analysis of Patterns by Partner-rated Conversationalist Score

In this section, we report complete results and robustness tests from our study of high- and low-skilled conversationalists. Section A.3.1 presents a comprehensive set of results from our main analysis, including additional transcript-based, auditory, and visual features not reported in the main text as well as additional subgroups of "middling" conversationalists. Section A.3.2 contains results on female and male participants alone, allowing an assessment of gender heterogeneity of results. Section A.3.3. reports results after controlling for turn duration, allowing for an assessment of whether differences in non-duration conversational patterns such



as speech speed or loudness may be in part driven by differences in turn duration. Finally, Section A.3.4. demonstrates that semantic similarity results are robust to the choice of a widely used alternative embedding model, RoBERTa, in place of the MPNet-based results presented in the main analyses.

### A.3.1. Complete Results from Main Analysis

In this section, we present comprehensive findings from our study of how highly skilled conversationalists (as rated by their partners) differ from their low-skilled counterparts. Results proceed as follows. Figures A.1, A.2, and A.3 respectively provide results on all transcript-based, auditory, and visual features; for completeness, we also report the estimated behavior of middling conversationalists (i.e., groups rated in the 25-50th percentile and 50-75th percentile) in addition to results on bad and good conversationalists (0-25th percentile and 75-100th percentile). Table A.1 provides a summary table assessing the statistical significance of differences in distributions between bad and good conversationalists, using $p$ values corrected for multiple testing.



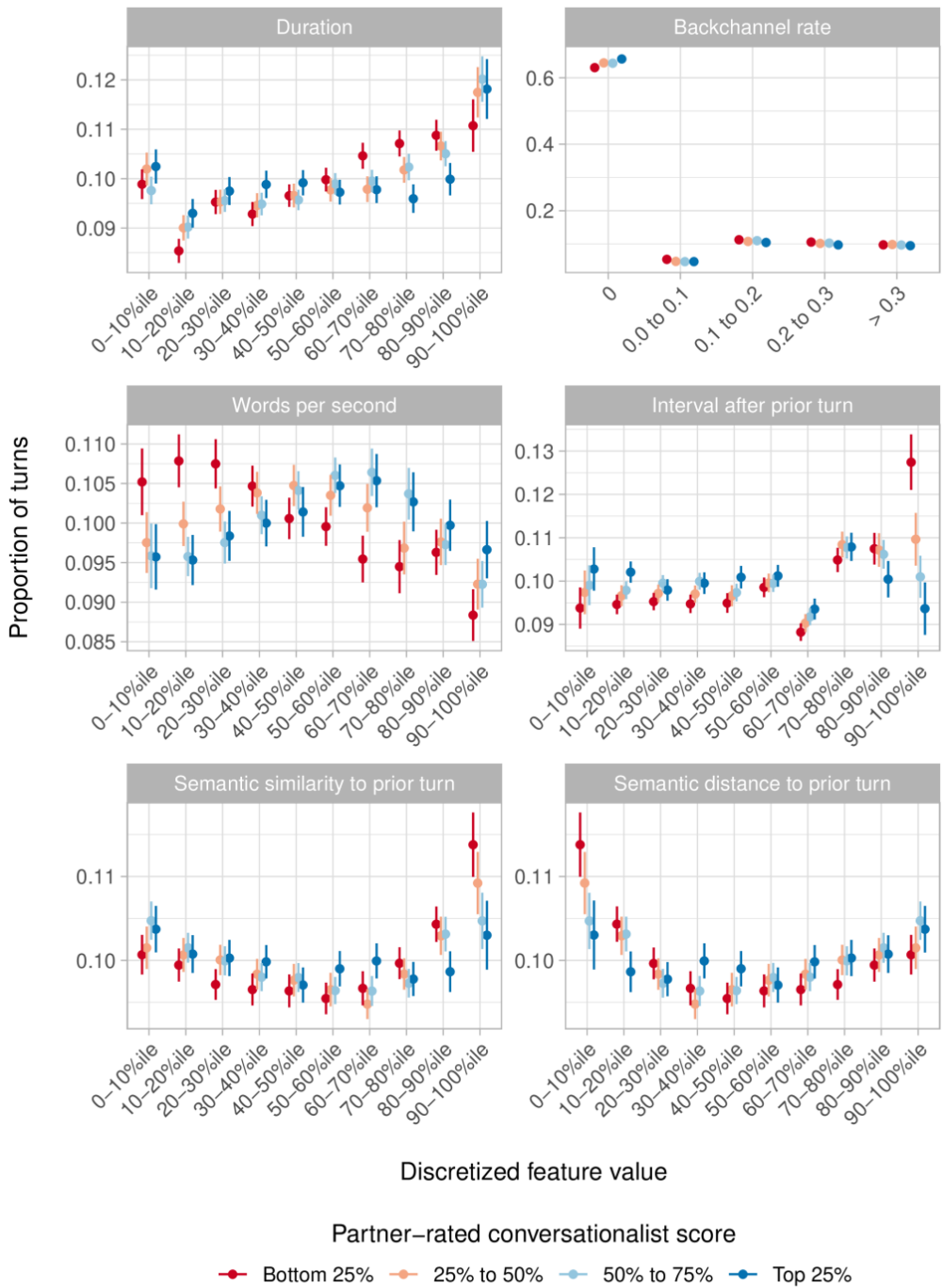

**Figure A.1. Behavior of good, middling, and bad conversationalists on transcript-based features.** Each panel depicts the engagement patterns of good conversationalists (top 25% of partner-rated conversationalist score, depicted in blue) and bad conversationalists (bottom 25%, red) on a turn-level characteristic, expanding upon Figure 9 in the main text with additional panels. For completeness, the plot also depicts middling conversationalists who are above the median (50–75th percentile, light blue) and below the median (25–50th percentile, light red). Horizontal axes denote categories of turn-



level characteristics, defined in terms of feature deciles. The vertical position of each point indicates the average proportion of turns in a category for each group of conversationalists.

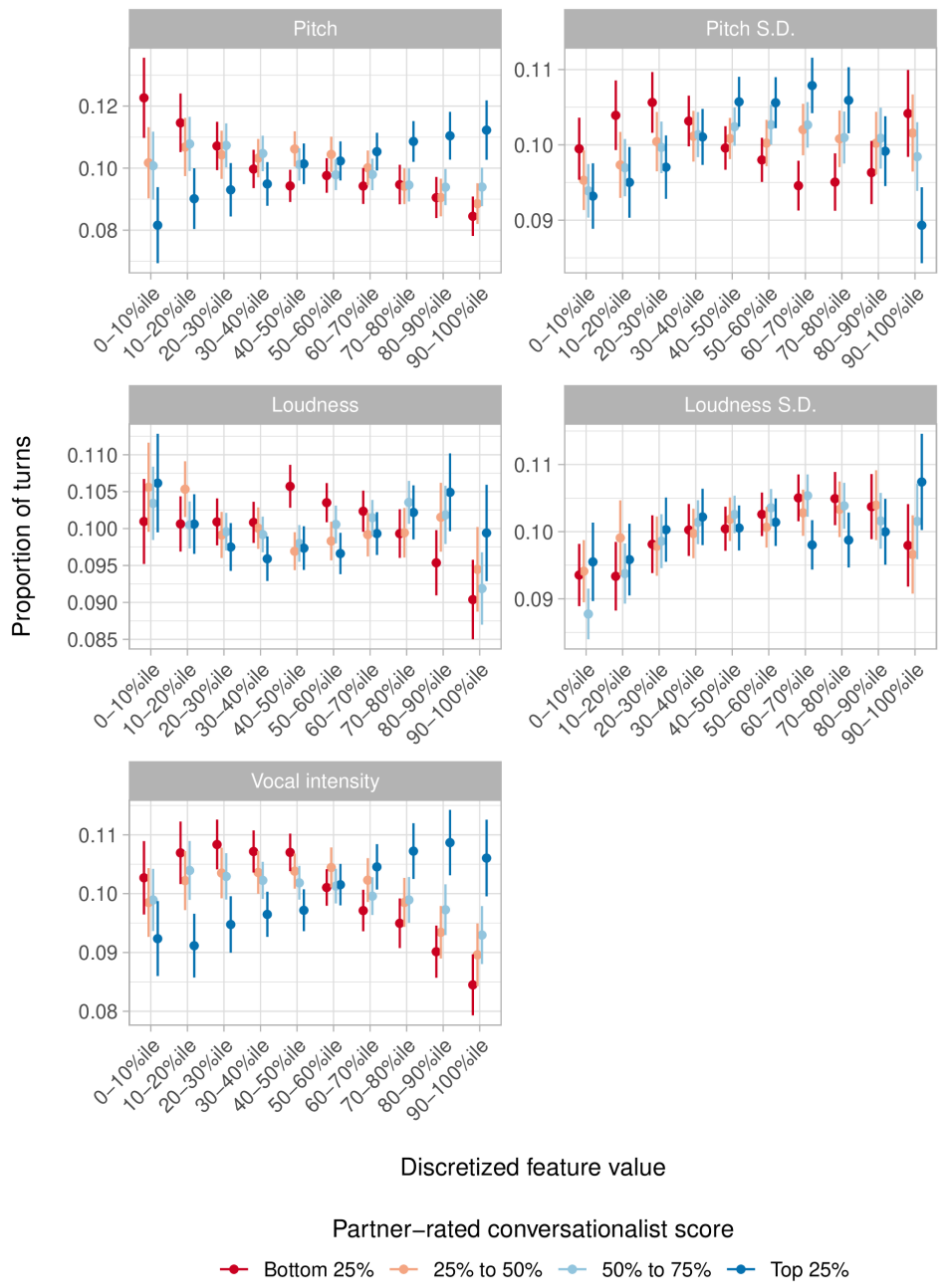

**Figure A.2. Behavior of good, middling, and bad conversationalists on auditory-based features.** Each panel depicts the engagement patterns of good conversationalists (top 25% of partner-rated conversationalist score, depicted in blue) and bad conversationalists (bottom 25%, red) on a turn-level characteristic, expanding upon Figure 9 in the main text with additional panels. For completeness, the plot also depicts middling conversationalists who are above the median (50–75th percentile, light blue) and below the median (25–50th percentile, light red). Horizontal axes denote categories of turn-



level characteristics, defined in terms of feature deciles. The vertical position of each point indicates the average proportion of turns in a category for each group of conversationalists.

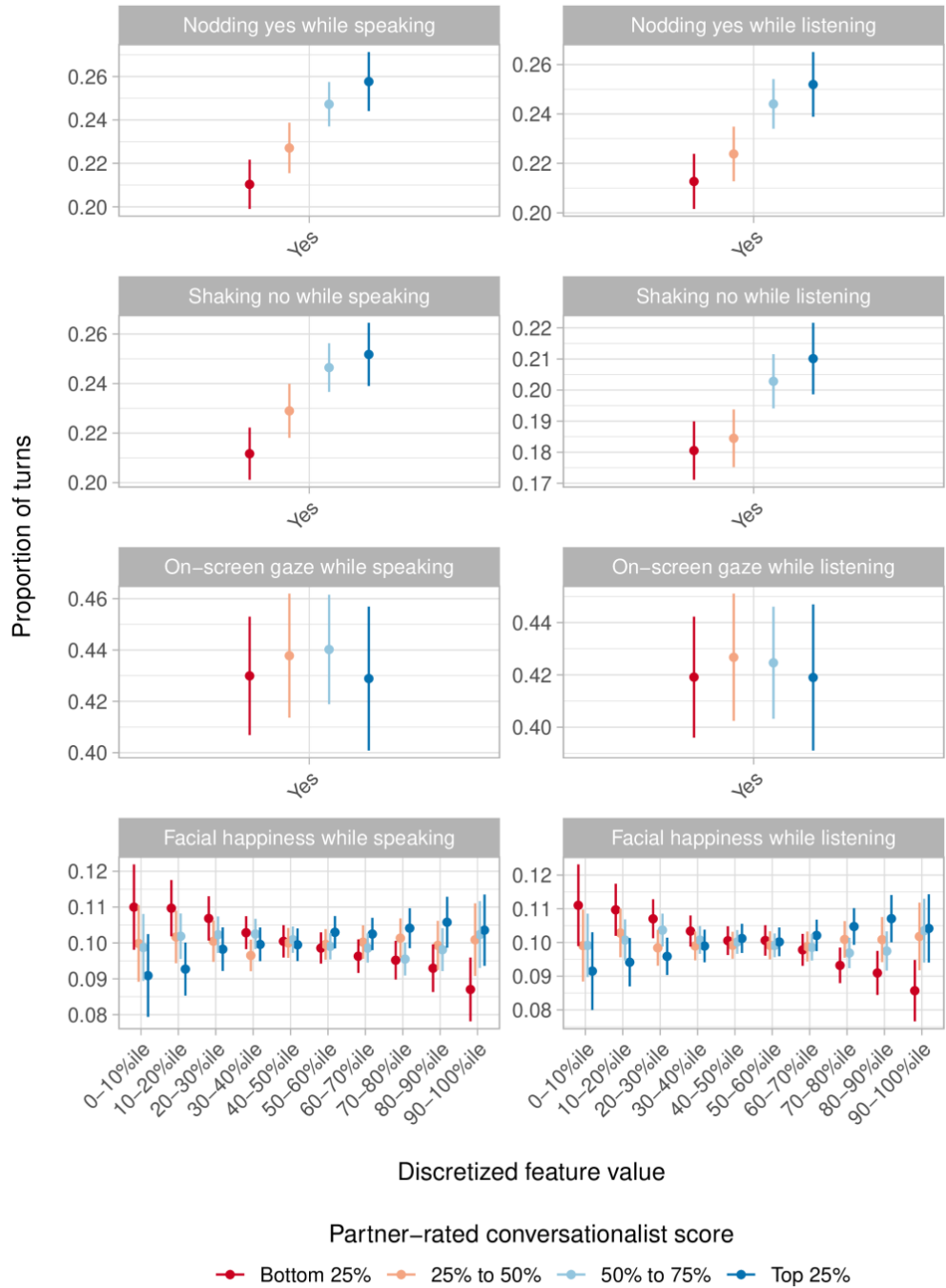

**Figure A.3. Behavior of good, middling, and bad conversationalists on visual-based features.** Each panel depicts the engagement patterns of good conversationalists (top 25% of partner-rated conversationalist score, depicted in blue) and bad conversationalists (bottom 25%, red) on a turn-level characteristic, expanding upon Figure 9 in the main text with additional panels. For completeness, the plot also depicts middling conversationalists who are above the median (50–75th percentile, light



blue) and below the median (25–50th percentile, light red). Horizontal axes denote categories of turn-level characteristics, defined in terms of feature deciles. The vertical position of each point indicates the average proportion of turns in a category for each group of conversationalists.

| | Feature | Diff. | 95% CI | $p_{adj}$ (mean) | $p_{adj}$ (distr.) |
|---|---|---|---|---|---|
| Transcript | Interval after prior turn | -0.0523 | [-0.0680, -0.0366] | <0.001 | <0.001 |
| Transcript | Duration | 0.0959 | [-0.1519, 0.3438] | 0.608 | <0.001 |
| Transcript | Words per second | 0.0996 | [0.0595, 0.1396] | <0.001 | <0.001 |
| Transcript | Backchannel rate | -0.0046 | [-0.0079, -0.0013] | 0.016 | <0.001 |
| Transcript | Cosine similarity to prior | -0.0053 | [-0.0081, -0.0025] | 0.001 | 0.001 |
| Transcript | Euclidean dist. to prior | 0.004 | [0.0014, 0.0067] | 0.009 | 0.001 |
| Auditory | Vocal intensity | 0.0131 | [0.0072, 0.0190] | <0.001 | <0.001 |
| Auditory | Pitch | 11.18 | [6.84, 15.53] | <0.001 | <0.001 |
| Auditory | Loudness | 0.1189 | [-0.4229, 0.6608] | 0.809 | 0.016 |
| Auditory | Pitch S.D. | -0.2609 | [-1.2030, 0.6812] | 0.744 | <0.001 |
| Auditory | Loudness S.D. | 0.0555 | [-0.3872, 0.4982] | 0.835 | 0.025 |
| Visual | Facial happiness (listening) | 0.0354 | [0.0150, 0.0557] | 0.003 | 0.045 |
| Visual | On-screen gaze (listening) | -0.0071 | [-0.0430, 0.0288] | 0.81 | 0.994 |
| Visual | Nodding yes (listening) | 0.0401 | [0.0215, 0.0588] | <0.001 | <0.001 |
| Visual | Shaking no (listening) | 0.0298 | [0.0134, 0.0461] | 0.002 | <0.001 |
| Visual | Facial happiness (speaking) | 0.0318 | [0.0118, 0.0518] | 0.006 | 0.255 |
| Visual | On-screen gaze (speaking) | -0.0059 | [-0.0417, 0.0300] | 0.821 | 0.981 |
| Visual | Nodding yes (speaking) | 0.0495 | [0.0302, 0.0687] | <0.001 | <0.001 |
| Visual | Shaking no (speaking) | 0.041 | [0.0230, 0.0591] | <0.001 | <0.001 |

**Table A.1. Statistical significance of differences in behavior between good and bad conversationalists (main results).** Each row assesses differences between good conversationalists (top 25% of partner-rated conversationalist score) and bad conversationalists (bottom 25%) on a turn-level conversational feature. Separate $p$ values are reported for tests of distributional equality and for tests of mean equality. The table reports only main analyses (all participants, no adjustment for turn duration), but multiple-testing adjustment accounts for robustness tests reported elsewhere (analyses restricted to male and female participants, as well as adjusting for turn duration).

*A.3.2. Gender-disaggregated Results*

In this section, we present additional findings from robustness tests that subset female and male respondents before comparing high- and low-skilled conversationalists.



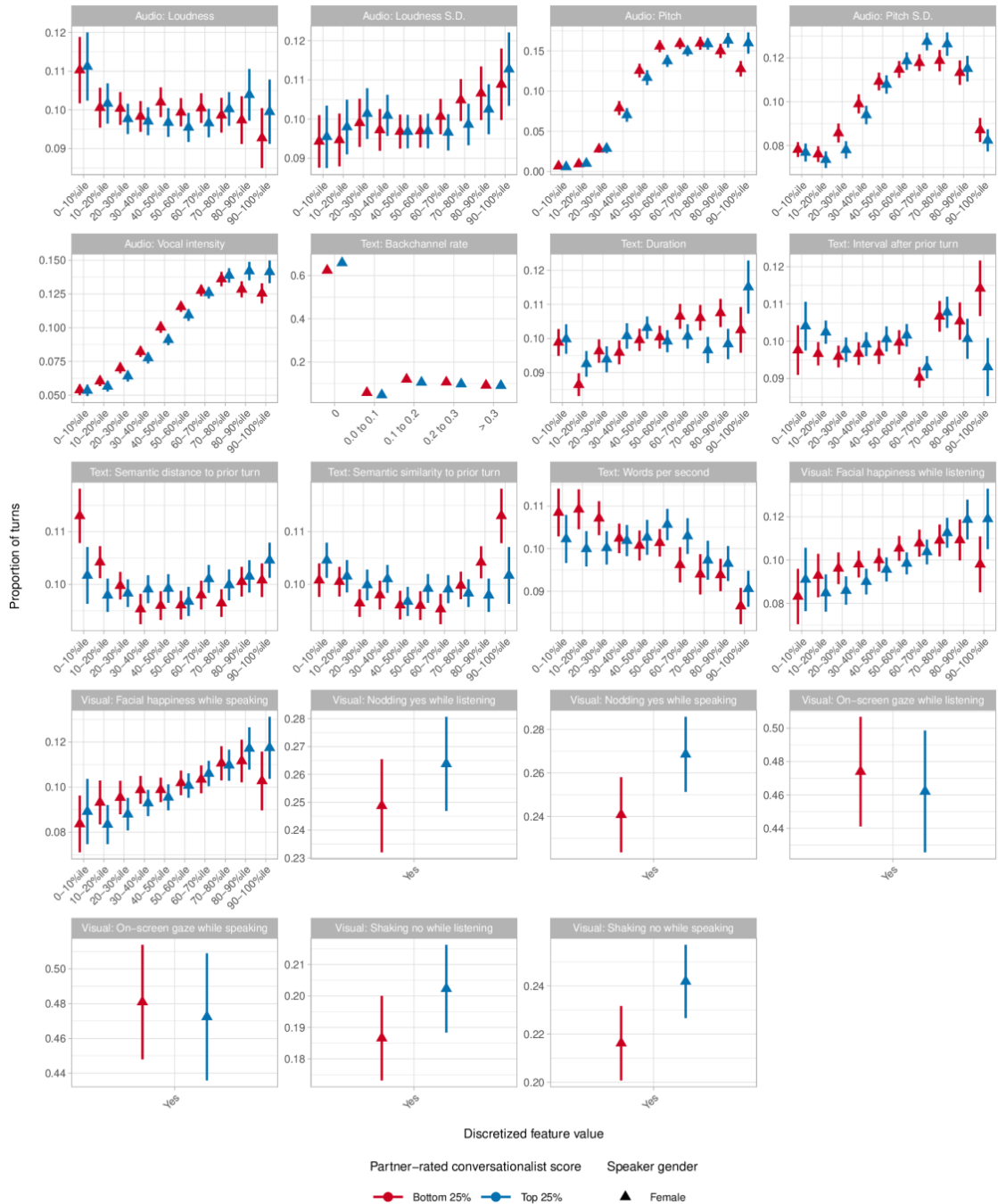

**Figure A.4. Behavior of good and bad conversationalists (female participants).** Each panel depicts the engagement patterns of good conversationalists (top 25% of partner-rated conversationalist score, depicted in blue) and bad conversationalists (bottom 25%, red) on a turn-level characteristic. Horizontal axes denote categories of turn-level characteristics, defined in terms of feature deciles. The vertical position of each point indicates the average proportion of turns in a category for good or bad conversationalists.



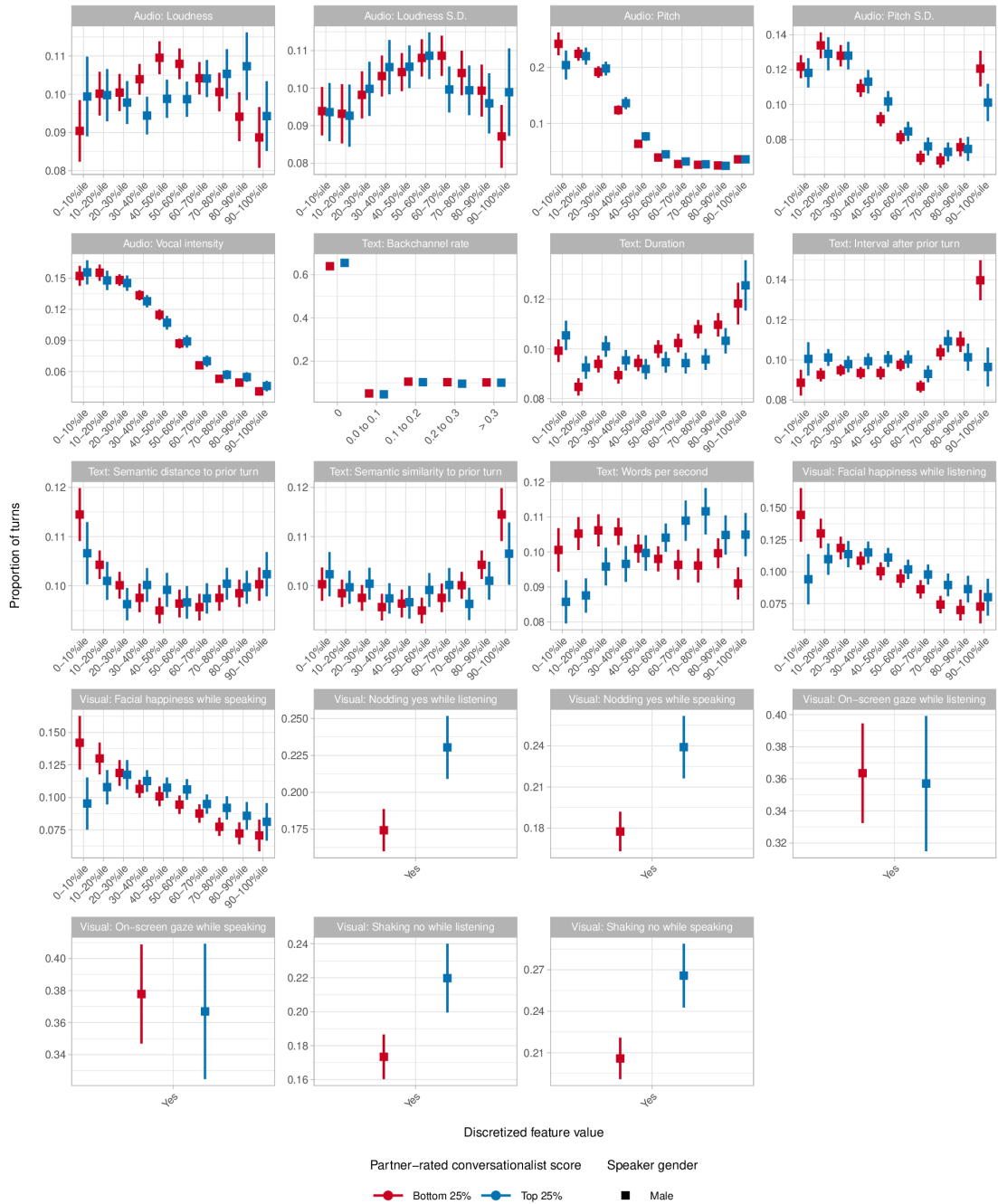

**Figure A.5. Behavior of good and bad conversationalists (male participants).** Each panel depicts the engagement patterns of good conversationalists (top 25% of partner-rated conversationalist score, depicted in blue) and bad conversationalists (bottom 25%, red) on a turn-level characteristic. Horizontal axes denote categories of turn-level characteristics, defined in terms of feature deciles. The vertical position of each point indicates the average proportion of turns in a category for good or bad conversationalists.

| Feature | | Diff. | 95% CI | $p_{adj}$ (mean) | $p_{adj}$ (distr.) |
|---|---|---|---|---|---|
| Transcript | Interval after prior turn | -0.034 | [-0.0550, -0.0139] | 0.003 | 0.060 |
| Transcript | Duration | 0.270 | [-0.0498, 0.5893] | 0.164 | <0.001 |



| | | | | |
|---|---|---|---|---|
| Transcript | Words per second | 0.060 [0.0091, 0.1107] | 0.044 | 0.212 |
| Transcript | Backchannel rate | -0.005 [-0.0097, -0.0004] | 0.065 | <0.001 |
| Transcript | Cosine similarity to prior | -0.006 [-0.0094, -0.0017] | 0.013 | 0.038 |
| Transcript | Euclidean dist. to prior | 0.004 [0.0003, 0.0078] | 0.065 | 0.038 |
| Auditory | Vocal intensity | 0.005 [0.0004, 0.0103] | 0.065 | 0.069 |
| Auditory | Pitch | 4.730 [1.42, 8.05] | 0.014 | 0.002 |
| Auditory | Loudness | 0.141 [-0.6036, 0.8858] | 0.810 | 0.691 |
| Auditory | Pitch S.D. | 0.200 [-0.7544, 1.1543] | 0.809 | 0.043 |
| Auditory | Loudness S.D. | -0.062 [-0.6728, 0.5495] | 0.858 | 0.735 |
| Visual | Facial happiness (listening) | 0.020 [-0.0067, 0.0475] | 0.223 | 0.195 |
| Visual | On-screen gaze (listening) | -0.006 [-0.0546, 0.0422] | 0.835 | 0.710 |
| Visual | Nodding yes (listening) | 0.019 [-0.0069, 0.0442] | 0.233 | 0.259 |
| Visual | Shaking no (listening) | 0.019 [-0.0024, 0.0405] | 0.145 | 0.155 |
| Visual | Facial happiness (speaking) | 0.015 [-0.0115, 0.0419] | 0.376 | 0.392 |
| Visual | On-screen gaze (speaking) | -0.006 [-0.0543, 0.0421] | 0.835 | 0.786 |
| Visual | Nodding yes (speaking) | 0.033 [0.0063, 0.0591] | 0.035 | 0.043 |
| Visual | Shaking no (speaking) | 0.030 [0.0067, 0.0536] | 0.029 | 0.038 |

**Table 2. Statistical significance of differences in behavior between good and bad conversationalists (female participants).** Each row assesses differences between good conversationalists (top 25% of partner-rated conversationalist score) and bad conversationalists (bottom 25%) on a turn-level conversational feature within a participant gender. Separate *p* values are reported for tests of distributional equality and for tests of mean equality. The table reports only analyses among female participants, but multiple-testing adjustment accounts for additional tests discussed elsewhere

| | Feature | Diff. | 95% CI | $p_{adj}$ (mean) | $p_{adj}$ (distr.) |
|---|---|---|---|---|---|
| Transcript | Interval after prior turn | -0.066 | [-0.0914, -0.0410] | <0.001 | <0.001 |
| Transcript | Duration | 0.070 | [-0.3307, 0.4710] | 0.818 | 0.015 |
| Transcript | Words per second | 0.157 | [0.0919, 0.2220] | <0.001 | <0.001 |
| Transcript | Backchannel rate | -0.003 | [-0.0081, 0.0021] | 0.361 | 0.192 |
| Transcript | Cosine similarity to prior | -0.004 | [-0.0079, -0.0000] | 0.091 | 0.324 |
| Transcript | Euclidean dist. to prior | 0.003 | [-0.0005, 0.0067] | 0.160 | 0.322 |
| Auditory | Vocal intensity | 0.002 | [-0.0053, 0.0095] | 0.744 | 0.193 |
| Auditory | Pitch | 2.849 | [-0.9402, 6.6389] | 0.223 | 0.255 |
| Auditory | Loudness | 0.017 | [-0.8124, 0.8456] | 0.969 | 0.021 |
| Auditory | Pitch S.D. | -1.279 | [-3.04, 0.48] | 0.233 | 0.030 |
| Auditory | Loudness S.D. | 0.165 | [-0.5010, 0.8310] | 0.777 | 0.141 |
| Visual | Facial happiness (listening) | 0.038 | [0.0079, 0.0687] | 0.032 | 0.053 |
| Visual | On-screen gaze (listening) | -0.022 | [-0.0746, 0.0302] | 0.565 | 0.855 |
| Visual | Nodding yes (listening) | 0.058 | [0.0293, 0.0864] | <0.001 | <0.001 |
| Visual | Shaking no (listening) | 0.046 | [0.0192, 0.0737] | 0.003 | <0.001 |
| Visual | Facial happiness (speaking) | 0.037 | [0.0068, 0.0665] | 0.035 | 0.060 |
| Visual | On-screen gaze (speaking) | -0.019 | [-0.0715, 0.0337] | 0.638 | 0.744 |
| Visual | Nodding yes (speaking) | 0.064 | [0.0345, 0.0942] | <0.001 | <0.001 |



| | | | | |
|---|---|---|---|---|
| Visual | Shaking no (speaking) | 0.061 [0.0312, 0.0910] | <0.001 | <0.001 |

**Table 3. Statistical significance of differences in behavior between good and bad conversationalists (male participants).** Each row assesses differences between good conversationalists (top 25% of partner-rated conversationalist score) and bad conversationalists (bottom 25%) on a turn-level conversational feature within a participant gender. Separate *p* values are reported for tests of distributional equality and for tests of mean equality. The table reports only analyses among male participants, but multiple-testing adjustment accounts for additional tests discussed elsewhere.

### *A.3.3. Duration-adjusted Results*

In this section, we present additional findings from robustness tests that controlled for turn duration in comparing high- and low-skilled conversationalists. To do so, we included demeaned turn duration as a linear predictor in the regressions described in Appendix A.2.1. This allowed for turn proportions in each category (lowest decile of a feature, second-lowest decile, etc.) to increase or decrease linearly as a function of duration. For example, the model allowed for a slight reduction in extremely loud speech (as measured by average decibels over the turn) for each additional second that the turn continued; this accounted for the possibility that, for instance, loud speech was difficult to sustain for long periods. At the same time, the model allowed for differently sized increases or decreases of other loudness categories (e.g., moderately quiet speech) for each additional second of turn duration. Figure A.6 depicts the predicted engagement patterns of good and bad conversationalists, holding turn duration fixed at the average value across the corpus. Table A.3 summarizes the statistical significance of duration-adjusted differences between high- and low-skilled conversationalists.



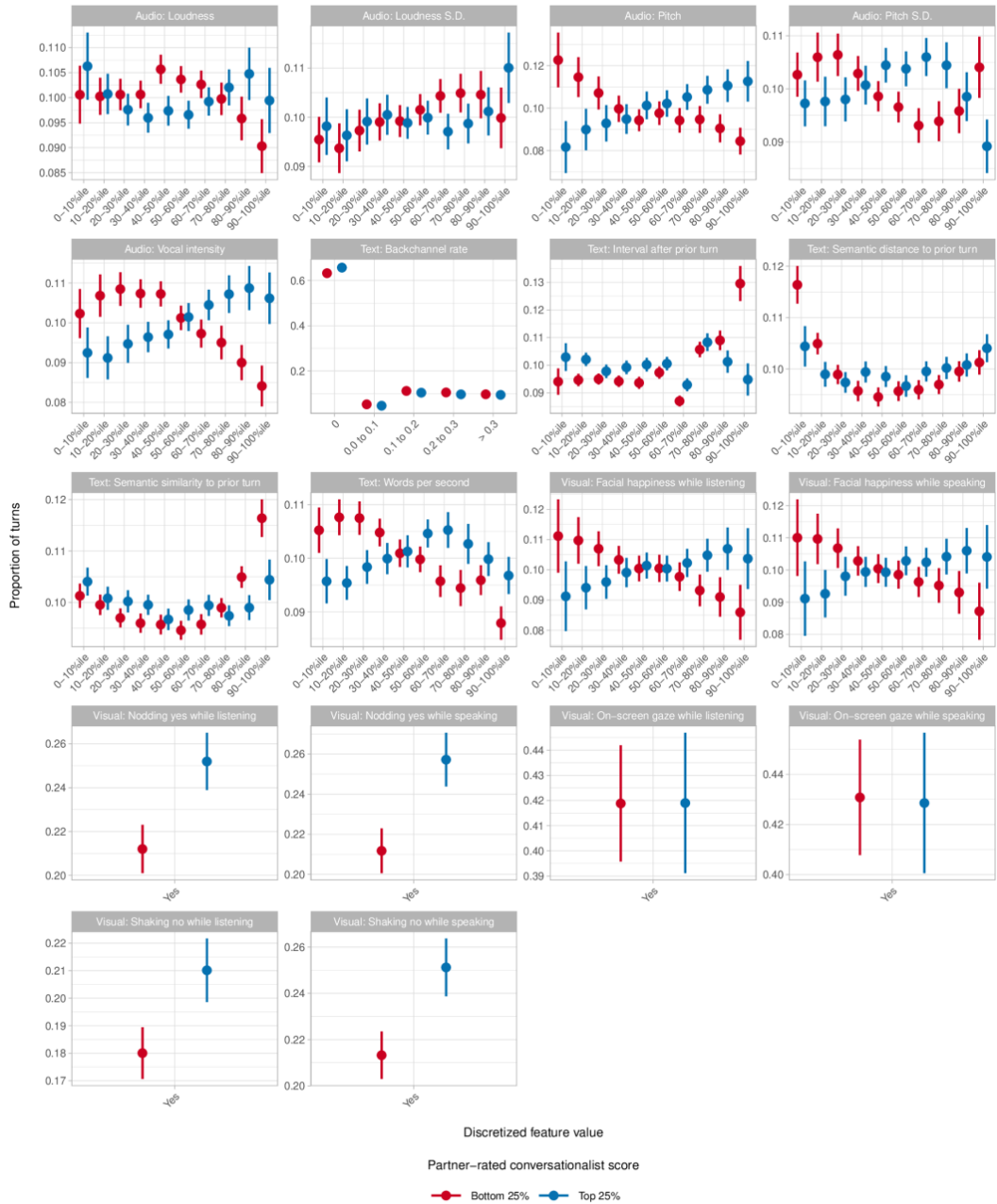

**Figure A.6. Behavior of good and bad conversationalists (duration-adjusted results).** Each panel depicts the engagement patterns of good conversationalists (top 25% of partner-rated conversationalist score, depicted in blue) and bad conversationalists (bottom 25%, red) on a turn-level characteristic. Horizontal axes denote categories of turn-level characteristics, defined in terms of feature deciles. The vertical position of each point indicates the average proportion of turns in a category for good or bad conversationalists.

*A.3.4. Robustness of Semantic Similarity Results*



Finally, we demonstrate that results on novelty and semantic similarity were not simply idiosyncratic artifacts of the particular embedding model we used to reduce turn transcripts to a quantitative representation. While the MPNet embedding model used in our main analyses was selected on the basis of achieving the highest average performance across a number of domains, RoBERTa embeddings (Liu et al., 2019) are a widely used alternative. In broad strokes, Figure A.7 replicates the overall pattern of our findings: bad conversationalists had higher average similarity to the previous turn, indicating more repetitive, less novel, responses. However, we find that these differences did not manifest in a consistent manner across embedding models. MPNet results suggested that bad conversationalists were differentiated by a large number of extremely high-similarity statements that were near duplicates of prior turns. In contrast, RoBERTa results indicated that bad conversationalists had fewer high-novelty (low-similarity) statements. While both approaches suggested that poor conversationalists' contributions were more mundane, the practical implications of the different models' results present substantially different interpretations. Resolving the apparent divergence between these approaches may constitute an important area for future work.



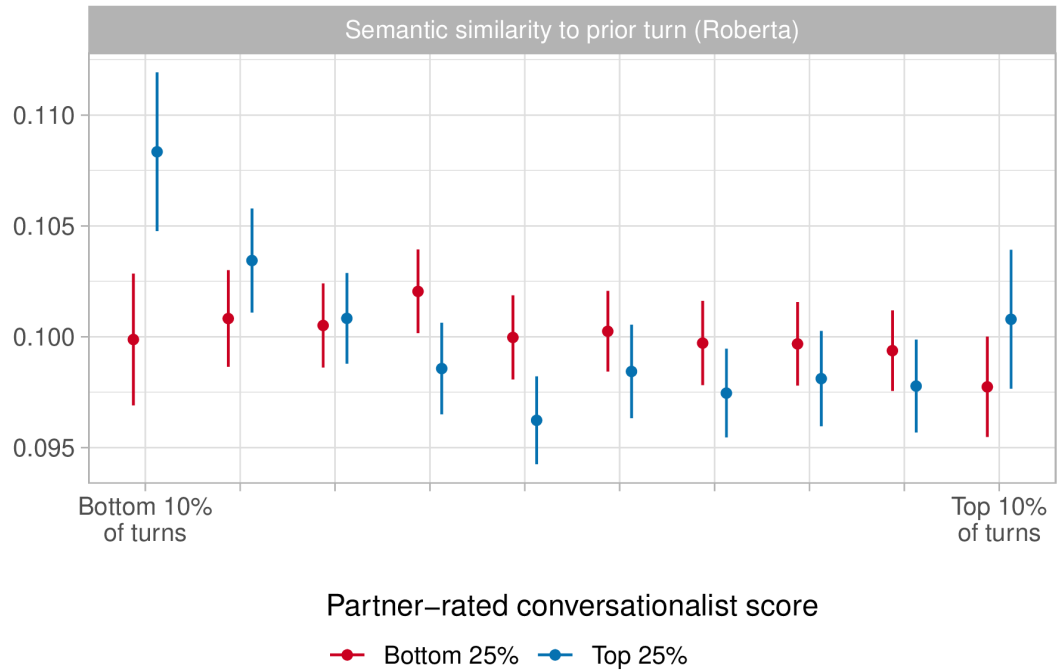

**Figure A.7**. Behavior of good and bad conversationalists on RoBERTa embedding similarity. Cosine similarity of current turn to partner's prior turn for good conversationalists (top 25% of partner-rated conversationalist score, depicted in blue) and bad conversationalists (bottom 25%, red). Horizontal axes denote categories of semantic similarity, defined in terms of deciles across the corpus. The vertical position of each point indicates the average proportion of turns in a category for good or bad conversationalists.

## Section 5: Quasi-experimental Analysis of Conversation Patterns by Partner Identity

In this section, we present results that examine how members of one identity group shift their conversational patterns when quasi-randomly assigned to partners of (1) their own group, or (2) partners of a differing group.

The same procedure assessing whether K groups of participants diverge in their conversational behavior was applied here; for a detailed discussion, see Supplement Section A.2. Here we first restrict analysis to participants from one demographic (e.g., older participants), then examine whether those participants engage in conversation differently when assigned to older, middling, or younger partners, i.e. among K=3 subgroups of older participants). Similarly,



to correct for multiple testing, the Benjamini-Hochberg procedure described in Supplement

Section A.2.3 was applied here to 340 tests about quasi-randomly assigned partner identity on

participant behavior (involving nine subgroups of participants and 17 contrasts between in-group

partners and various out-groups of partners, again repeated on 20 features).

All analyses compare *within* a group (e.g. subsetting to young participants), making

contrasts within that subset based on the group of the assigned partner (e.g., comparing those

assigned to old partners, as opposed to young partners). We emphasize that young and old

participants differ in many ways, such as their education level and political attitudes. Our

analysis does not seek to disentangle which specific attribute drives the difference in engagement

patterns—that is, it does not claim that effects are due to the age gap alone, holding all other

attributes fixed. Rather, it aims to approximate an ideal experiment in which a participant is

randomly assigned to converse with a partner from group A or B, where A and B differ on some

aspect of identity as well as the "bundle of sticks" (Sen and Wasow, 2016) that are associated

with or comprise that identity. Moreover, it does not attempt to identify psychological

mechanisms underlying the change in an individual's behavior, such as out-group animosity.

Finally, we note that as discussed in the main text, differences in one participant's behavior can

arise as a response to differing behavior by another participant. Throughout, reported *p* values

are adjusted for the multiplicity of features analyzed and partner-group comparisons made.

Partner assignment is based on an algorithm that greedily matches pairs of participants

that indicated their availability during the same time slot. Because the matching algorithm does

not incorporate demographic information, whether a participant is assigned to an in-group or out-

group partner is guaranteed to be ignorable conditional on availability. For purposes of analysis,

we assume that it is ignorable when aggregating over availability blocks as well. To assess the



plausibility of this design assumption, we conduct chi-squared tests to evaluate dependence in participant and partner identity—for example, whether older participants are more likely to be paired with other older participants, compared to a null model in which they are randomly assigned to partners of all ages. Chi-squared tests for dependence in age, gender, race, education, and political ideology pairings respectively produce $p$ values of 0.53, 0.33, 0.62, 0.91, and 0.43. These results suggest that availability is at most weakly related to membership in an identity group. Moreover, within identity groups, we assess that availability is unlikely to correlate strongly with baseline conversational patterns.

In what follows, Sections A.4.1–3 respectively present results on quasi-randomly assigned partner age, gender, and race/ethnicity. We do not detect significant differences in conversational patterns by partner education (distributed 37%, 40%, and 23% respectively below, at, and above the level of a bachelor's degree) or political ideology (65% liberal, 20% neutral, 15% conservative), though we caution that the statistical power of political-ideology results is limited by the relatively small proportion of conservative participants. These results are omitted to conserve space, but reported $p$ values include adjustment for all analyses that were conducted. To aid interpretation, Table A.4 in Section A.4.4 summarizes differences in average feature values with 95% confidence intervals; multiple-testing-adjusted $p$ values are reported for tests of differences in means.

### A.4.1. Quasi-random Partner Age Results

To analyze age, we divide participants approximately into tertiles representing the youngest (aged 19–28; N=1,204), middle (29–38; N=1,013), or oldest age groups (39–66; N=1,039). Note that tertiles are slightly imbalanced due to rounding in reported age. Figures A.8, A.9, and A.10 respectively present analyses that subset to young, middle, and oldest participants, examining the



distribution of their conversational features when paired with in- and out-group partners. For

compactness, we plot only results that are statistically significant at the 0.05 level.

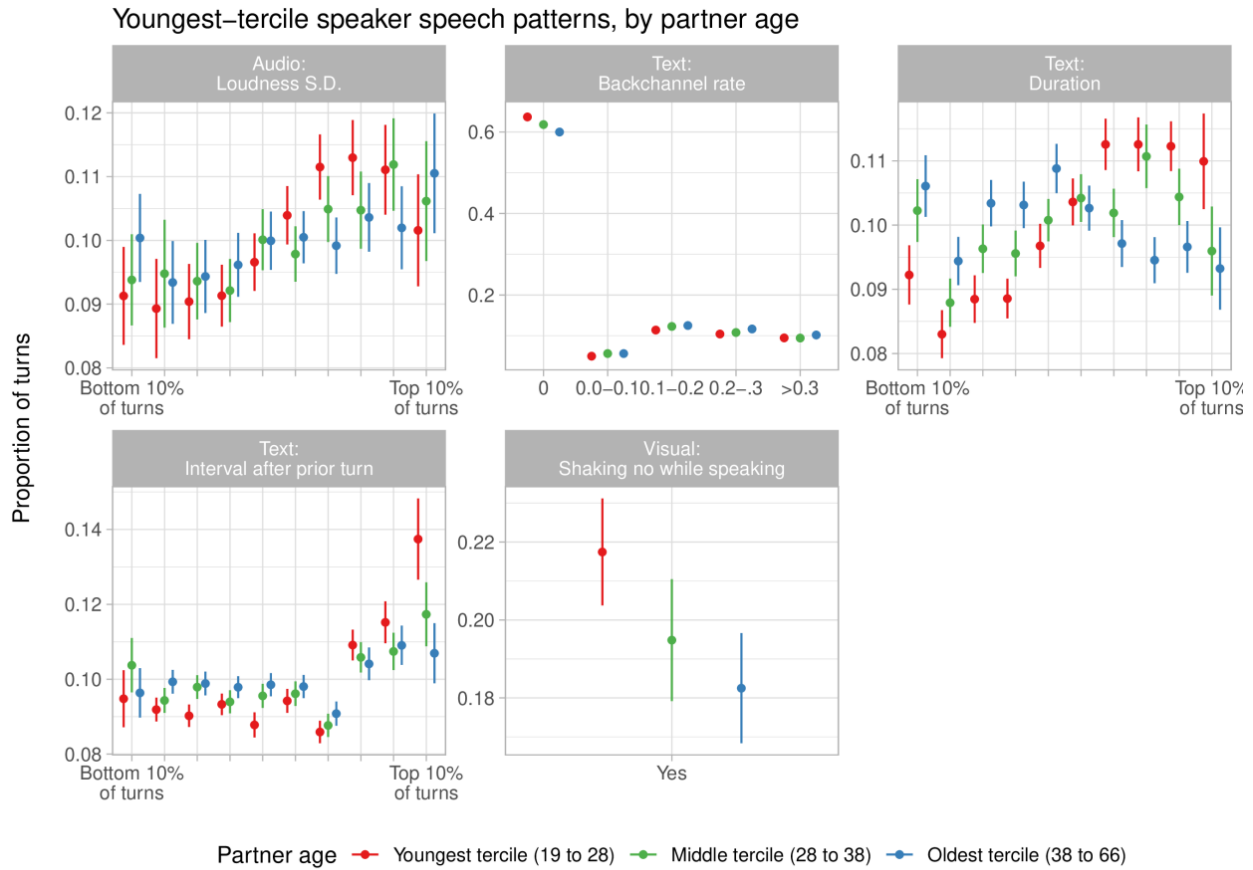

**Figure A.8. Behavior of youngest-tercile participants when assigned to young, middle, and old age-group partners.** Each panel depicts the engagement patterns of young participants assigned to young (red), middle (green), or old (blue) age-group partners on a turn-level characteristic. Horizontal axes denote categories of turn-level characteristics, defined in terms of feature deciles. The vertical position of each point indicates the average proportion of turns in a category. Distributions are presented only for features in which the null hypothesis of distributional equality is rejected at the 0.05 level after multiple-testing adjustment.



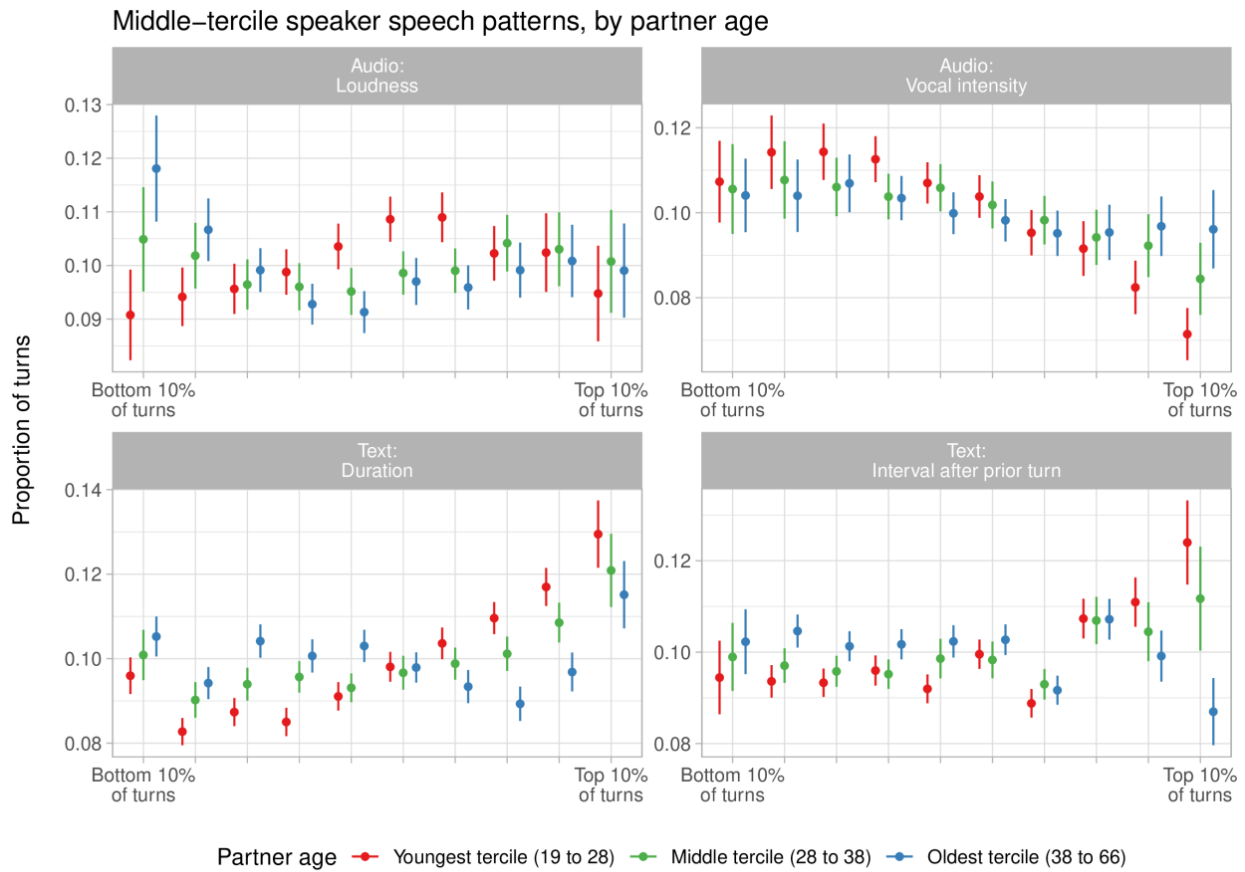

**Figure A.9. Behavior of middle age-group participants when assigned to young, middle, and old age-group partners.** Each panel depicts the engagement patterns of middle age-group participants assigned to young (red), middle (green), or old (blue) age-group partners on a turn-level characteristic. Horizontal axes denote categories of turn-level characteristics, defined in terms of feature deciles. The vertical position of each point indicates the average proportion of turns in a category. Distributions are presented only for features in which the null hypothesis of distributional equality is rejected at the 0.05 level after multiple-testing adjustment.



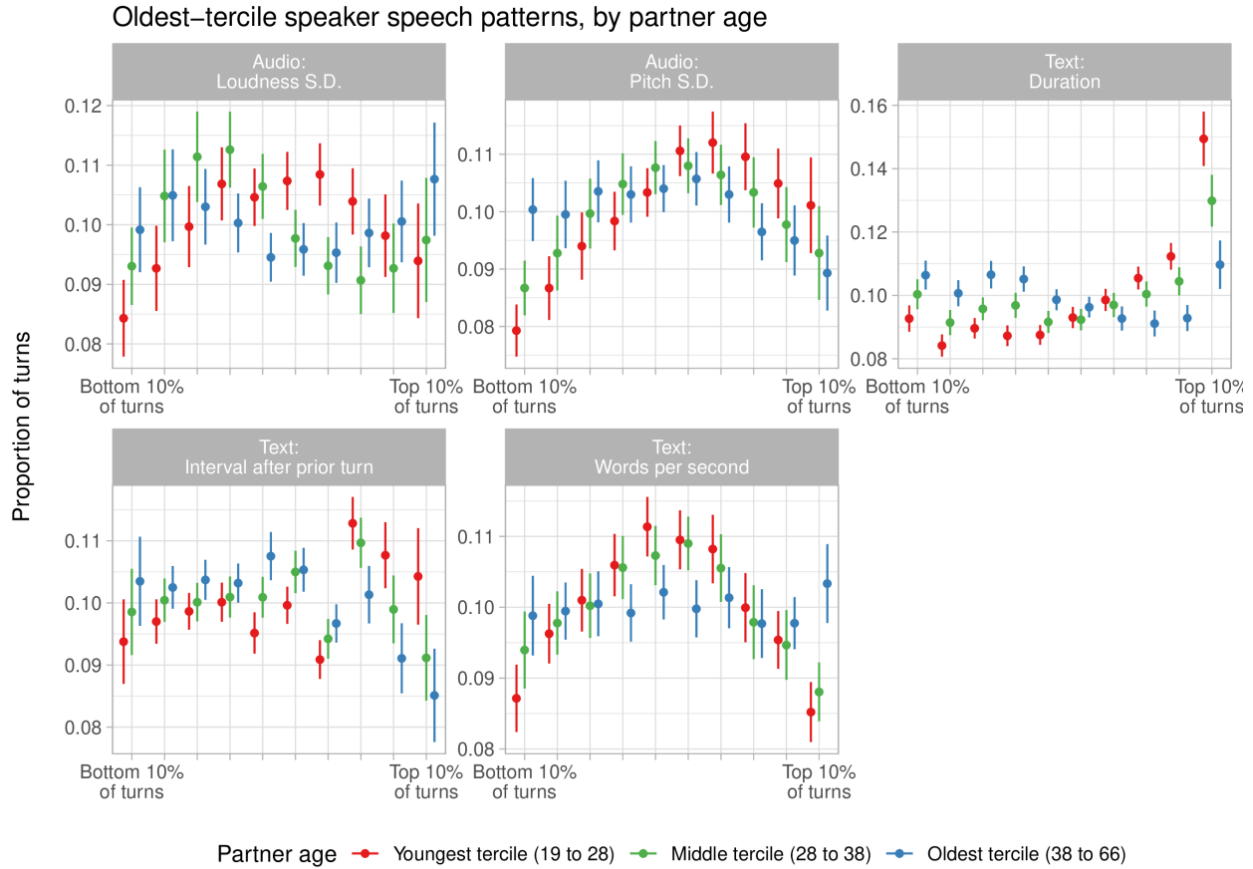

**Figure A.10. Behavior of oldest-tertile participants when assigned to young, middle, and old age-group partners.** Each panel depicts the engagement patterns of old participants assigned to young (red), middle (green), or old (blue) age-group partners on a turn-level characteristic. Horizontal axes denote categories of turn-level characteristics, defined in terms of feature deciles. The vertical position of each point indicates the average proportion of turns in a category. Distributions are presented only for features in which the null hypothesis of distributional equality is rejected at the 0.05 level after multiple-testing adjustment.

### *A.4.2. Quasi-random Partner Gender Results*

To analyze gender, we examine participants who self-describe as female (N=1,740) or male (N=1,463). Participants with other gender identities, as well as those who preferred not to answer, were not analyzed due to a lack of statistical power (N=109). Figures A.11 and A.12 respectively present analyses that subset to female and male participants, examining the distribution of their conversational features when paired with in- and out-group partners. For compactness, we plot only results that are statistically significant at the 0.05 level.



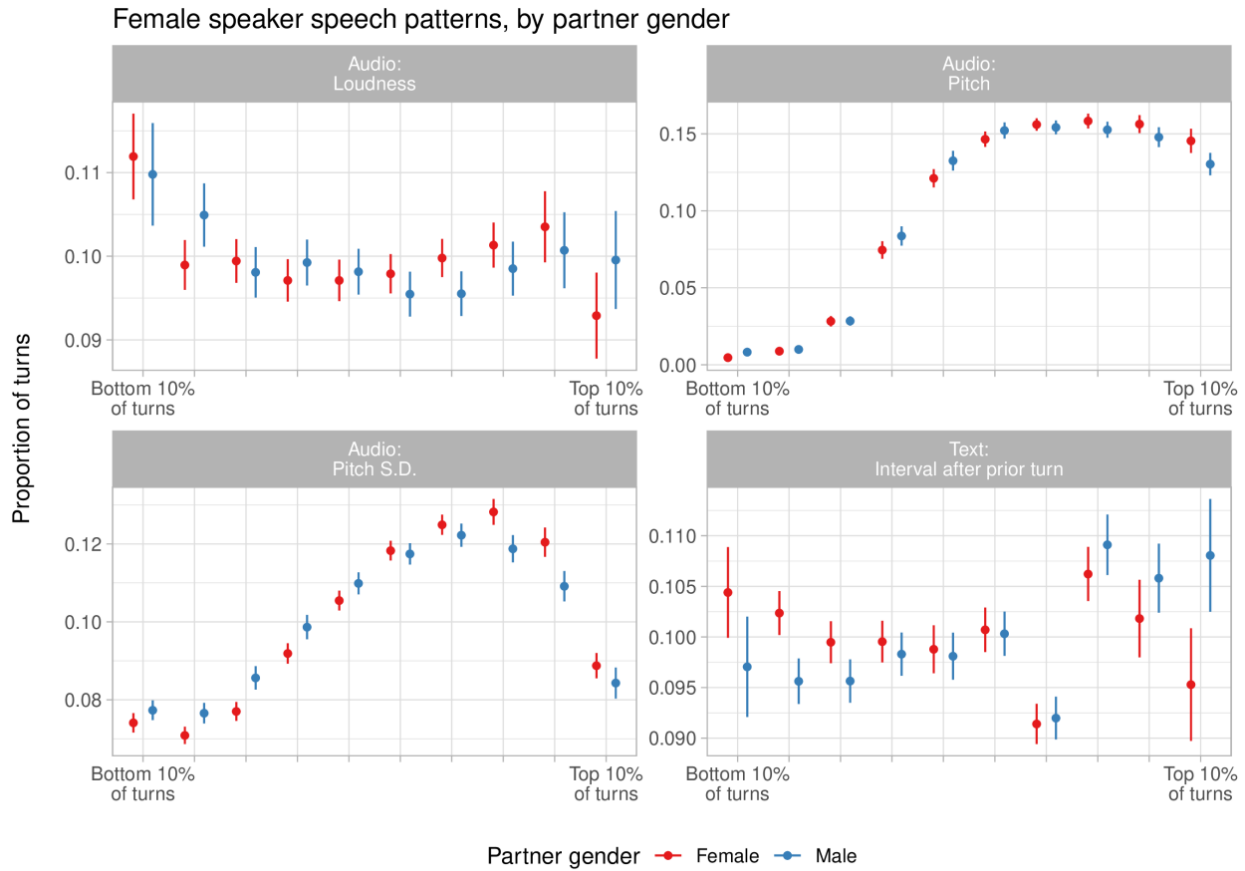

**Figure A.11. Behavior of female participants when assigned to female and male partners.**
Each panel depicts the engagement patterns of female participants assigned to female (red) or male (blue) partners on a turn-level characteristic. Horizontal axes denote categories of turn-level characteristics, defined in terms of feature deciles. The vertical position of each point indicates the average proportion of turns in a category. Distributions are presented only for features in which the null hypothesis of distributional equality is rejected at the 0.05 level after multiple-testing adjustment.

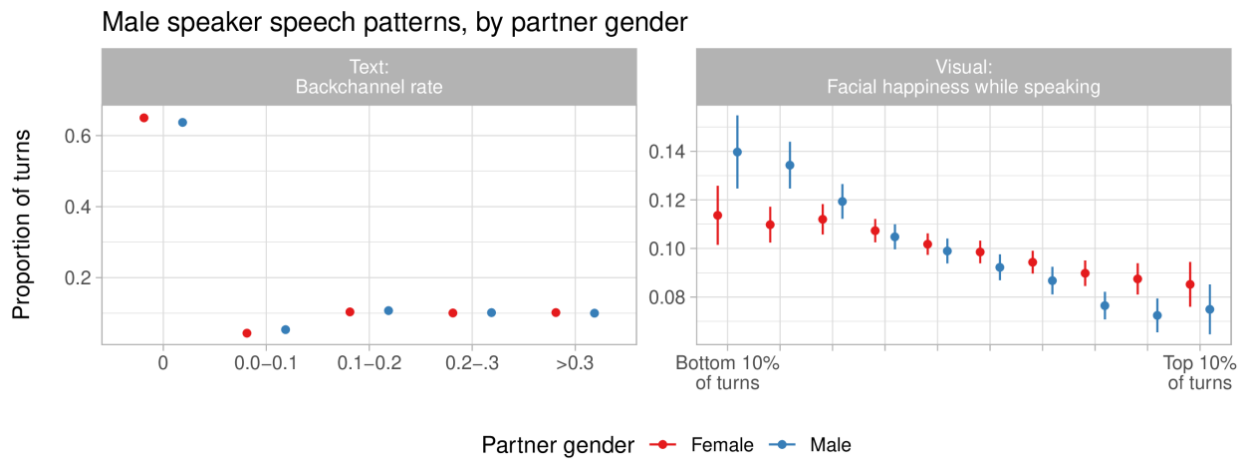



**Figure A.12. Behavior of male participants when assigned to female and male partners.**
Each panel depicts the engagement patterns of male participants assigned to female (red) or male (blue) partners on a turn-level characteristic. Horizontal axes denote categories of turn-level characteristics, defined in terms of feature deciles. The vertical position of each point indicates the average proportion of turns in a category. Distributions are presented only for features in which the null hypothesis of distributional equality is rejected at the 0.05 level after multiple-testing adjustment.

*A.4.3. Quasi-random Partner Race/Ethnicity Results*

To analyze race and ethnicity, we examine participants self-describing as Asian (N=485, 16%),

Black (N=248, 8%), Hispanic (N=220, 7%), or White (N=2,110, 69%). These proportions

roughly track the U.S. population (6% Asian, 13% Black, 19% Hispanic, and 60% White in 2021

Census data) but under-represent Black and Hispanic groups. Figure A.13 subsets to White

participants and examines the distribution of their conversational features when paired with in-

and out-group partners. Analyses of behavior by non-White groups is not feasible in this dataset

due to the sparsity of minority-minority pairings. For compactness, we plot only results that are

statistically significant at the 0.05 level

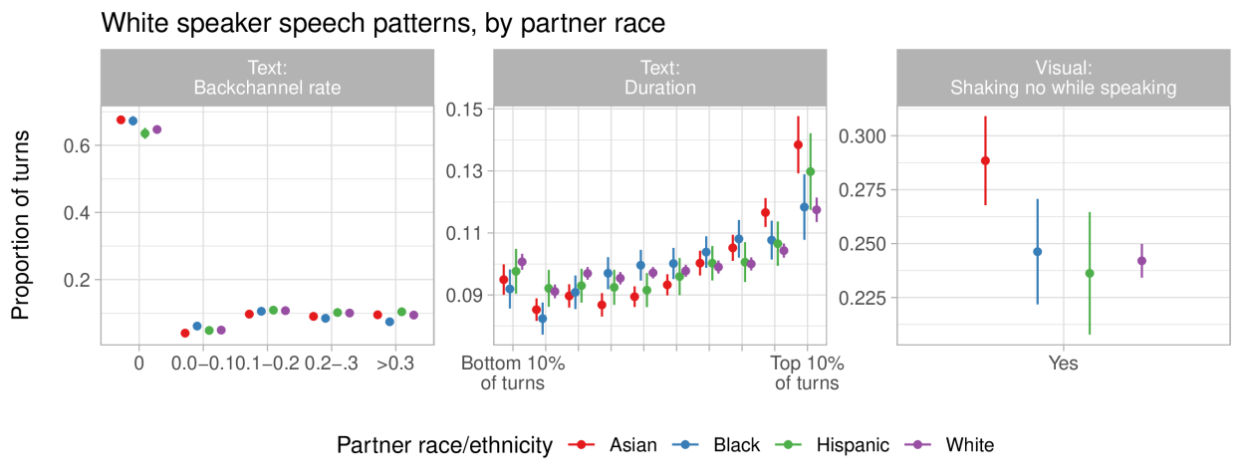

**Figure A.13. Behavior of White participants when assigned to Asian, Black, Hispanic, and White partners.** Each panel depicts the engagement patterns of White participants assigned to Asian (red), Black (blue), Hispanic (green), or White (purple) partners on a turn-level characteristic. Horizontal axes denote categories of turn-level characteristics, defined in terms of feature deciles. The vertical position of each point indicates the average proportion of turns in a category. Distributions are presented only for features in which the null hypothesis of distributional equality is rejected at the 0.05 level after multiple-testing adjustment.



| Participant Group | Partner Contrast | Feature | Diff. | 95% CI | $p_{adj}$ (mean) |
|---|---|---|---|---|---|
| | | *Gender-based analyses* | | | |
| F | M - F | Interval after prior turn | 0.0263 | [0.0111, 0.0414] | 0.018 |
| F | M - F | Pitch | -3.402 | [-5.6571, -1.1470] | 0.047 |
| F | M - F | Pitch S.D. | -1.213 | [-1.8446, -0.5815] | 0.007 |
| M | F - M | Facial happiness (speaking) | 0.033 | [0.0120, 0.0540] | 0.038 |
| M | F - M | Facial happiness (listening) | 0.034 | [0.0127, 0.0554] | 0.038 |
| | | *Age-based analyses* | | | |
| Y | M - Y | Duration | -0.5208 | [-0.8331, -0.2085] | 0.027 |
| Y | O - Y | Interval after prior turn | -0.0411 | [-0.0660, -0.0161] | 0.029 |
| Y | O - Y | Duration | -0.7693 | [-1.0665, -0.4722] | <0.001 |
| Y | O - Y | Backchannel rate | 0.0072 | [0.0026, 0.0119] | 0.038 |
| Y | O - Y | Shaking no (speaking) | -0.0338 | [-0.0553, -0.0123] | 0.038 |
| Y | O - Y | Nodding yes (listening) | 0.0365 | [0.0122, 0.0608] | 0.047 |
| M | O - M | Interval after prior turn | -0.0375 | [-0.0622, -0.0127] | 0.047 |
| O | M - O | Duration | 0.7401 | [0.3957, 1.0845] | 0.002 |
| O | Y - O | Interval after prior turn | 0.045 | [0.0234, 0.0666] | 0.002 |
| O | Y - O | Duration | 1.4652 | [1.1082, 1.8222] | <0.001 |
| O | Y - O | Shaking no (speaking) | 0.0381 | [0.0131, 0.0630] | 0.047 |
| O | Y - O | Pitch S.D. | 2.4611 | [1.2109, 3.7113] | 0.005 |
| | | *Race/ethnicity-based analyses* | | | |
| W | A - W | Duration | 0.7511 | [0.4451, 1.0572] | <0.001 |
| W | A - W | Cosine similarity to prior | 0.0059 | [0.0025, 0.0092] | 0.018 |
| W | A - W | Euclidean dist. to prior | -0.0054 | [-0.0085, -0.0023] | 0.018 |
| W | A - W | Shaking no (speaking) | 0.0434 | [0.0200, 0.0668] | 0.010 |
| W | B - W | Backchannel rate | -0.0128 | [-0.0170, -0.0085] | <0.001 |

**Table A.4. Statistical significance of differences in behavior of a participant group, contrasting members assigned to out-group partners versus in-group partners.** Each row reports the difference in the behavior of a group of participants toward out-group partners, as compared to in-group partners. The first column specifies the participant group for which conversational behavior is being analyzed. In gender analyses, abbreviations indicate Female and Male groups; in age analyses, Youngest, Middle, and Oldest tertile; in race/ethnicity analyses, Asian, Black, Hispanic, and White. The second column specifies the comparison of partner groups in an abbreviated "X - Y"; in each comparison, the first letter (here, "X") represents the out-group abbreviation and the second ("Y") always corresponds to the participant group being analyzed. Subsequent columns report differences in average conversational behavior toward out-group partners (compared to in-group partners), 95% confidence intervals, and multiple-testing adjusted $p$ values for the difference in expectation. To conserve space, only differences in expectation significant at the 95% level after multiple-testing adjustment are reported; note that multiple-testing adjustment accounts for 19 features and a total of 17 partner-group contrasts, totaling 323 analyses; these include additional education and ideology analyses for which no significant difference was found. Winsorized differences are reported for unbounded features.